\def\paperversion{2}
\ifnum\paperversion=1
\documentclass[opre,nonblindrev]{informs4}
\fi
\ifnum\paperversion=2
\documentclass[opre]{informs4}
\fi
\newcommand{\Var}{\mathbb{V}\mathrm{ar}}
\usepackage{booktabs}
\usepackage{anyfontsize}
\usepackage{multicol}
\usepackage{multirow}
\usepackage[numbers]{natbib}
\usepackage{floatflt}
 \def\bibfont{\small}%
\TheoremsNumberedThrough   
\ECRepeatTheorems
\theoremstyle{TH}%

\usepackage{wrapfig}
\newcommand{\RNum}[1]{\uppercase\expandafter{\romannumeral #1\relax}}
\newcommand{\cO}{\mathcal{O}}
\newcommand{\tO}{\widetilde{\mathcal{O}}}
\newcommand{\Proj}{\mathrm{Proj}}
\newcommand*{\QED}{%
\leavevmode\unskip\penalty9999 \hbox{}\nobreak\hfill
    \quad\hbox{$\square$}%
}
\newcommand*{\QEG}{%
\leavevmode\unskip\penalty9999 \hbox{}\nobreak\hfill
    \quad\hbox{$\clubsuit$}%
}

\theoremstyle{TH}%
\usepackage{booktabs}
\usepackage[font=small]{caption}
\usepackage[size=small]{subcaption}
\DeclareMathAlphabet{\mathsfit}{T1}{\sfdefault}{\mddefault}{\sldefault}
\SetMathAlphabet{\mathsfit}{bold}{T1}{\sfdefault}{\bfdefault}{\sldefault}
\DeclareMathAlphabet{\mathcal}{OMS}{cmsy}{m}{n}
\usepackage{enumitem}
\setlist[enumerate,1]{label=\normalfont{(\Roman*)},leftmargin=*}

\usepackage[margin=1in]{geometry}
\usepackage{caption}
\usepackage{mathtools}
\usepackage{algorithm}
\usepackage{algorithmic}

\usepackage{endnotes}

\let\footnote=\endnote
\RequirePackage{xcolor}
\definecolor{dkgreen}{rgb}{0,0.6,0}
\definecolor{gray}{rgb}{0.5,0.5,0.5}
\definecolor{mauve}{rgb}{0.58,0,0.82}

\newcommand{\nout}{n^{\mathrm{o}}}

\newcommand{\cP}{\mathcal{P}}

\newcommand{\prox}{\mathrm{Prox}}

\newcommand{\bD}{\mathbb{D}}

\newcommand{\cZ}{\mathcal{Z}}

\newcommand{\cW}{\mathcal{W}}
\newcommand{\SG}{\mathrm{SG}}
\newcommand{\RTMLMC}{\mathrm{RT}\text{-}\mathrm{MLMC}}
\newcommand{\esssup}{\mathrm{ess}\text{-}\mathrm{sup}}
\newcommand{\hP}{\widehat{\mathbb{P}}}

\newcommand{\trans}{^{\mathrm T}}

\newcommand{\diff}{\,\mathrm{d}}

\DeclarePairedDelimiterX{\inp}[2]{\langle}{\rangle}{#1, #2}

\newcommand{\bE}{\mathbb{E}}
\newcommand{\bP}{\mathbb{P}}
\newcommand{\bQ}{\mathbb{Q}}

\usepackage{hyperref}
\hypersetup{%
  colorlinks=true,%
  linkcolor={blue!66!black},%
  linkbordercolor=red,%
  citecolor={blue!66!black},
}

\TITLE{
Regularization for Adversarial Robust Learning
}
\ARTICLEAUTHORS{
\AUTHOR{Jie Wang}
\AFF{School of Industrial and Systems Engineering, Georgia Institute of Technology,
\EMAIL{jwang3163@gatech.edu}}
\AUTHOR{Rui Gao}
\AFF{Department of Information, Risk and Operations Management, The University of Texas at Austin,  \EMAIL{rui.gao@mccombs.utexas.edu}}
\AUTHOR{Yao Xie}
\AFF{School of Industrial and Systems Engineering, Georgia Institute of Technology, \EMAIL{yao.xie@isye.gatech.edu}}
}

\ABSTRACT{
Despite the growing prevalence of artificial neural networks in real-world applications, their vulnerability to adversarial attacks remains a significant concern, which motivates us to investigate the robustness of machine learning models.
While various heuristics aim to optimize the distributionally robust risk using the $\infty$-Wasserstein metric, such a notion of robustness frequently encounters computation intractability.
To tackle the computational challenge, we develop a novel approach to adversarial training that integrates $\phi$-divergence regularization into the distributionally robust risk function.
This regularization brings a notable improvement in computation compared with the original formulation.
We develop stochastic gradient methods with biased oracles to solve this problem efficiently, achieving the near-optimal sample complexity.
Moreover, we establish its regularization effects and demonstrate it is asymptotic equivalence to a regularized empirical risk minimization framework, by considering various scaling regimes of the regularization parameter and robustness level. 
These regimes yield gradient norm regularization, variance regularization, or a smoothed gradient norm regularization that interpolates between these extremes.
We numerically validate our proposed method in supervised learning, reinforcement learning, and contextual learning and showcase its state-of-the-art performance against various adversarial attacks.
}

\begin{document}
\maketitle


\section{Introduction}
Machine learning models are highly vulnerable to potential \emph{adversarial attack} on their input data, which intends to cause wrong outputs.
Even if the adversarial input is slightly different from the clean input drawn from the data distribution, these machine learning models can make a wrong decision.
\citet{goodfellow2014explaining} provided an example that, after adding a tiny adversarial noise to an image, a well-trained classification model may make a wrong prediction, even when such data perturbations are imperceptible to visual eyes.

Given that modern machine learning models have been applied in many safety-critical tasks, 
such as autonomous driving, medical diagnosis, security systems, \emph{etc}, improving the resilience of these models against adversarial attacks in such contexts is of great importance.
Neglecting to do so could be risky or unethical and may result in severe consequences.
For example, if we use machine learning models in self-driving cars, adversarial examples could allow an attacker to cause the car to take unwanted actions.

\emph{Adversarial training} is a process of training machine learning model to make it more robust to potential adversarial attacks.
To be precise, it aims to optimize the following robust optimization~(RO) formulation, called \emph{adversarial risk minimization}:
\begin{equation}
\min_{\theta\in\Theta}~\Big\{
\bE_{z\sim \hP}\big[
R_{\rho}(\theta;z)
\big]
\Big\},
\quad \text{where }R_{\rho}(\theta;z)\triangleq\sup_{z'\in\mathbb{B}_{\rho}(z)}~f_{\theta}(z').
\label{Eq:formula:adv}%
\end{equation}
Here $\hP$ represents the observed distribution on data, $\theta$ represents the machine learning model, $f_{\theta}(z)$ is a loss function, and the uncertainty set is defined as $\mathbb{B}_{\rho}(x)\triangleq\{z\in\cZ:~\|z-x\|\le \rho\}$ for some norm function $\|\cdot\|$ and some radius $\rho>0$.
In other words, this formulation seeks to train a machine learning model based on adversarial perturbations of data, where the adversarial perturbations can be found by considering all possible inputs around the data with radius $\rho$ and picking the one that yields the worst-case loss.
Unfortunately, problem~\eqref{Eq:formula:adv} is typically intractable to solve because the inner supremum objective function is in general nonconcave in $z$.
As pointed out by~\citep{sinha2017certifying}, solving the inner supremum problem in \eqref{Eq:formula:adv} with deep neural network loss functions is NP-hard.
Several heuristic algorithms~\citep{goodfellow2014explaining, kurakin2016adversarial, papernot2016limitations, carlini2017towards, madry2017towards, tramer2017ensemble} have been proposed to approximately find the optimal solution of \eqref{Eq:formula:adv}, but they lack of global convergence guarantees and it remains an open question whether they can accurately and efficiently find the adversarial perturbations of data.

In this paper, we propose a new approach for adversarial risk minimization by adding a $\phi$-divergence regularization.
Here is a brief overview.
By \citep[Lemma~EC.2]{gao2022wasserstein}, Problem~\eqref{Eq:formula:adv} can be viewed as the dual reformulation of the following DRO problem:%
\begin{equation}
\tag{$\infty$-WDRO}
\begin{aligned}
&\min_{\theta\in\Theta}~\left\{
\sup_{\mathbb{P}}~\Big\{\mathbb{E}_{z\sim \mathbb{P}}[f_{\theta}(z)]:~\cW_{\infty}(\mathbb{P},\hP)\le \rho
\Big\}
\right\},
\end{aligned}\label{Eq:formulation:wass}
\end{equation}%
where $\cW_{\infty}(\cdot,\cdot)$ is the $\infty$-Wasserstein metric defined as
\[
\cW_{\infty}(\bP,\bQ) = \inf_{\gamma}~\left\{
\text{ess.sup}_{\gamma}~\|\zeta_1-\zeta_2\|:~
\begin{array}{l}
\mbox{$\gamma$ is a joint distribution of $\zeta_1$ and $\zeta_2$}
\\
\mbox{with marginals $\bP$ and $\bQ$, respectively}
\end{array}
\right\}.
\]
Therefore, it is convenient to introduce the optimal transport mapping $\gamma$ to re-write problem~\eqref{Eq:formulation:wass} as 
\begin{equation}
\min_{\theta\in\Theta}~\left\{\sup_{
\substack{\bP,\gamma
}}~\left\{
\bE_{\bP}[f_{\theta}(z)]:~
\begin{array}{l}
\text{Proj}_{1\#}\gamma=\hP, \text{Proj}_{2\#}\gamma=\bP\\
\text{ess.sup}_{\gamma}\|\zeta_1-\zeta_2\|\le\rho
\end{array}
\right\}\right\}.
\label{Eq:formula:adv:transport}
\end{equation}
As long as the loss $f_{\theta}(z)$ is nonconcave in $z$, such as neural networks and other complex machine learning models, problem~\eqref{Eq:formula:adv:transport} is intractable for arbitrary radius $\rho>0$. Instead, we add \emph{$\phi$-divergence regularization} to the objective in \eqref{Eq:formula:adv:transport}, and focus on solving the following formulation:
\begin{equation}
\tag{Reg-$\infty$-WDRO}
\min_{\theta\in\Theta}~\left\{\sup_{
\substack{\bP,\gamma
}}~\left\{
\bE_{z\sim\bP}~[f_{\theta}(z)] - \eta \bD_{\phi}(\gamma,\gamma_0):~
\begin{array}{l}
\text{Proj}_{1\#}\gamma=\hP, \text{Proj}_{2\#}\gamma=\bP\\
\text{ess.sup}_{\gamma}\|\zeta_1-\zeta_2\|\le\rho
\end{array}
\right\}\right\},
\label{Eq:ent:adv:transport}
\end{equation}
where $\gamma_0$ is the reference measure satisfying $\diff\gamma_0(x,z)=\diff\hP(x)\diff\nu_x(z)$, with $\nu_x$ being the uniform probability measure on $\mathbb{B}_{\rho}(x)$, and $\bD_{\phi}(\gamma,\gamma_0)$ is the $\phi$-divergence~\citep{cover1999elements} between $\gamma$ and $\gamma_0$.
In the following, we summarize several notable features of our proposed formulation.
\subsubsection*{Strong Dual Reformulation.}
By the duality result in Theorem~\ref{Thm:strong:dual:general}, \eqref{Eq:ent:adv:transport} admits the strong dual reformulation:
\begin{subequations}
\label{Eq:CSO:formula}
\begin{align}
\min_{\theta\in\Theta}&\quad \bE_{z\sim\hP}[\psi_{\eta}(z)],\\
\mbox{where}&\quad \psi_{\eta}(z) =\inf_{\mu\in\mathbb{R}}~\bigg\{ 
\mu + \bE_{z'\sim\nu_x}\left[ 
(\eta\phi)^*\big(f_{\theta}(z') - \mu\big)
\right]
\bigg\}.\label{Eq:CSO:b}
\end{align}    
\end{subequations}
Compared with the original formulation~\eqref{Eq:formula:adv}, we replace the worst-case loss $R_{\rho}(\theta;z)$ defined in \eqref{Eq:formula:adv} by $\phi_{\eta}(z)$, which is a variant of \emph{optimized certainty equivalent}~(OCE) risk measure studied in \citep{ben1987penalty}.
Subsequently, it can be shown that $\phi_{\eta}(x)$ is a smooth approximation of the optimal value $R_{\rho}(\theta;x)$.
\subsubsection*{Worst-case Distribution Characterization.}
We characterize the worst-case distribution for problem~\eqref{Eq:ent:adv:transport} in Remark~\ref{Sec:discussions} and display its simplified expressions in Examples~\ref{Exp:indicator}-\ref{Exp:hinge}.
In contrast to the conventional formulation~\eqref{Eq:formulation:wass} that \emph{deterministically} transports each data from $\hP$ to its extreme perturbation, the worst-case distribution of our formulation transports each data $x$ towards the entire domain set $\mathbb{B}_{\rho}(x)$ through specific absolutely continuous distributions.
This observation indicates that our formulation~\eqref{Eq:ent:adv:transport} is well-suited for adversarial defense where the data distribution after adversarial attack manifests as absolutely continuous, such as through the addition of white noise to the data.
\subsubsection*{Efficient Stochastic Optimization Algorithm.}
We adopt the idea of stochastic approximation to solve our reformulation~\eqref{Eq:CSO:formula} by iteratively obtaining a stochastic gradient estimator and next performing projected gradient update.
To tackle the difficulty that one cannot obtain the unbiased gradient estimator, we introduce and analyze stochastic gradient methods with biased oracles inspired from \citep{hu2021biasvar}.
Our proposed algorithm achieves $\tO(\epsilon^{-2})$ sample complexity for finding $\epsilon$-optimal solution for convex $f_{\theta}(z)$ and general choices of $\phi$-divergence, 
and $\tO(\epsilon^{-4})$ sample complexity for finding $\epsilon$-stationary point for nonconvex $f_{\theta}(z)$ and KL-divergence.
These sample complexity results are near-optimal up to a near-constant factor.
\subsubsection*{Regularization Effects.}
We develop regularization effects for problem~\eqref{Eq:ent:adv:transport} in Section~\ref{Sec:reg}. Specifically, we show that it is asymptotically equivalent to regularized ERM formulations under three different scalings of the regularization value $\epsilon$ and radius $\rho$:
Let $\beta$ be an uniform distribution supported on the unit norm ball $\mathbb{B}_{1}(0)$, and $\|\cdot\|_*$ be the dual norm of $\|\cdot\|$.
When $\rho,\eta\to0$, it holds that $\eqref{Eq:ent:adv:transport}=\min\limits_{\theta\in\Theta}~\bE_{z\sim \hP}[f_{\theta}(z)] + \mathcal{E}(f_{\theta}; \rho,\eta)$, where
\[
\mathcal{E}(f; \rho,\eta)\simeq\left\{ 
\begin{aligned}
&\rho\cdot\mathbb{E}_{z\sim\hP}\left[ 
\inf_{\mu\in\mathbb{R}}\left\{ 
\mu + \frac{1}{C}\mathbb{E}_{b\sim \beta}\left[ 
\phi^*\Big( 
C\cdot(\nabla f(z)\trans b - \mu)
\Big)
\right]
\right\}
\right],
&\quad \text{if }\frac{\rho}{\eta}\to C,\\
& \rho\cdot\bE_{z\sim\hP}[\|\nabla f(z)\|_*],
 &\quad \text{if }\frac{\rho}{\eta}\to \infty,\\
&\frac{\rho^2}{2\eta\cdot \phi''(1)}\cdot \bE_{z\sim \hP}\Big[ 
\Var_{b\sim\beta}[\nabla f(z)\trans b]
\Big],
&\quad \text{if }\frac{\rho}{\eta}\to 0.
\end{aligned}
\right.
\]


In other words, when $\rho/\eta\to\infty$, it corresponds to the gradient norm regularized ERM formulation; 
when $\rho/\eta\to0$, it corresponds to a special gradient variance regularized ERM formulation;
when $\rho/\eta\to C$, it corresponds to a regularized formulation that interpolates between these extreme cases. 
\subsubsection*{Generalization Error Analysis.}
We investigate the generalization properties of our proposed adversarial training framework.
In particular, the optimal value in \eqref{Eq:ent:adv:transport} is the confidence upper bound of its population version up to a negligible residual error.
Next, we present the specific generalization error bound for linear and neural network function classes.

\subsubsection*{Numerical Applications.}
Finally, we provide numerical experiments in Section~\ref{Sec:numerical} on supervised learning, reinforcement learning, and contextual learning.
Numerical results demonstrate the state-of-the-art performance attained by our regularized adversarial learning framework against various adversarial attacks.





\subsection*{Related Work}

\subsubsection*{Adversarial Learning.}
Ever since the seminal work \citep{goodfellow2014explaining} illustrated the vulnerability of neural networks to adversarial perturbations, the research on adversarial attack and defense has progressively gained much attention in the literature.
The NP-hardness of solving the adversarial training problem~\eqref{Eq:formula:adv} with ReLU neural network structure has been proved in \citet{sinha2017certifying}, indicating that one should resort to efficient approximation algorithms with satisfactory solution quality. 
Numerous approaches for adversarial defense have been put forth~\citep{goodfellow2014explaining, kurakin2016adversarial, papernot2016limitations, carlini2017towards, madry2017towards, tramer2017ensemble}, aiming to develop heuristic algorithms to optimize the formulation~\eqref{Eq:formula:adv} relying on the local linearization~(i.e., first-order Taylor expansion) of the loss $f_{\theta}$.
Unfortunately, the Taylor expansion may not guarantee an accurate estimate of the original objective in~\eqref{Eq:formula:adv}, especially when the robustness level $\rho$ is moderate or large.
Henceforth, these algorithms often fail to find the worst-case perturbations of the adversarial training.

\subsubsection*{Distributionally Robust Optimization.}
Our study is substantially related to the DRO framework.
In literature, the modeling of distributional uncertainty sets~(also called ambiguity sets) for DRO can be categorized into two approaches.
The first considers finite-dimensional parameterizations of the ambiguity sets by taking into account the support, shape, and moment information~\citep{bertsimas2006persistence, scarf1958min, Delage10, Goh10, Zymler13, wiesemann2014distributionally, Chen19, popescu2005semidefinite, van2015generalized}.
The second approach, which has received great attention recently, constructs ambiguity sets using non-parametric statistical discrepancy, including $f$-divergence~\citep{hu2013kullback, Ben13,wang2016likelihood,bayraksan2015data, Duchi21}, Wasserstein distance and its entropic-regularized variant~\citep{pflug2007ambiguity,wozabal2012framework,Mohajerin18,zhao2018data,blanchet2019quantifying, gao2016distributionally,chen2018data, xie2019distributionally, wang2021sinkhorn, azizian2023regularization, wang2024non}, and maximum mean discrepancy~\citep{staib2019distributionally, Kernelzhu}.

There are many results on the computational traceability of DRO.
\citet{sinha2017certifying} showed that replacing $\infty$-Wasserstein distance with $2$-Wasserstein distance in \eqref{Eq:formulation:wass} yields more tractable formulations.
Unfortunately, their proposed algorithm necessitates a sufficiently small robustness level such that the involved subproblem becomes strongly convex, which is not well-suited for adversarial training in scenarios with large perturbations.
\citet{wang2021sinkhorn} added entropic regularization regarding the $p$-WDRO formulation to develop more efficient algorithms.
We highlight that their result cannot be applied to the entropic regularization for $\infty$-WDRO setup because the associated transport cost function is not finite-valued.




\subsubsection*{Stochastic Gradient Methods with Biased Gradient Oracles.}
Stochastic biased gradient methods have received great attention in both theory and applications.
References~\citep{hu2017analysis, hu2016bandit, chen2018stochastic, ajalloeian2020convergence} construct gradient estimators with small biases at each iteration and analyze the iteration complexity of their proposed algorithms, ignoring the cost of querying biased gradient oracles.
\citet{hu2021biasvar, yjhc2024} proposed efficient gradient estimators using multi-level Monte-Carlo (MLMC) simulation and provided a comprehensive analysis of the total complexity of their algorithms by considering both iteration and per-iteration costs.
This kind of algorithm is especially useful when constructing unbiased estimators can be prohibitively expansive or even infeasible for many emerging machine learning and data science applications, such as $\phi$-divergence/Sinkhorn DRO~\citep{levy2020large, wang2021sinkhorn, zhang2024large}, meta learning~\citep{ji2022theoretical, hu2024contextual}, and contextual learning~\citep{diao2020distribution}.
We show that our formulation~\eqref{Eq:ent:adv:transport} can also be solved using this type of approach.


\noindent
\textbf{Notations.}
Denote by $\mathrm{Proj}_{1\#}\gamma, \mathrm{Proj}_{2\#}\gamma$ the first and the second marginal distributions of $\gamma$, respectively.
For a measurable set $\cZ$, denote by $\mathcal{P}(\cZ)$ the set of probability measures on $\cZ$.
Denote by $\mathrm{supp}\, \bP$ the support of probability distribution $\bP$.
Given a measure $\mu$ and a measurable variable $f:~\cZ\to\mathbb{R}$, we write $\mathbb{E}_{z\sim\mu}[f]$ for $\int f(z)\diff\mu(z)$.
Given a subset $E$ in Euclidean space, let $\mathrm{vol}(E)$ denote its volume.
Let $\theta^*\in\argmin\limits_{\theta\in\Theta}~F(\theta)$.
We say a given random vector $\theta$ is a $\delta$-optimal solution if $\bE[F(\theta) - F(\theta^*)]\le\delta$.
In addition, we say $\theta$ is a $\delta$-stationary point if for some step size $\gamma>0$, it holds that 
$\bE\left\|
\frac{1}{\gamma}
\left[
\theta - \Proj_{\Theta}\big(\theta -\delta\nabla F(\theta)\big)\right]
\right\|^2_2\le \delta^2.$
For a given probability measure $\mu$ in $\mathbb{R}^d$, denote by $f_{\#}\mu$ the pushforward measure of $\mu$ by $f:~\mathbb{R}^d\to\mathbb{R}$.

\section{Phi-Divergence Regularized Adversarial Robust Training}\label{Sec:phi:divergence}
In this section, we discuss the regularized formulation of the adversarial robust training problem~\eqref{Eq:formula:adv:transport}.
Define the reference measure $\nu_z$ as the uniform probability measure supported on $\mathbb{B}_{\rho}(z)\subseteq \cZ$, i.e., 
\begin{equation}
\frac{\diff\nu_z(\omega)}{\diff\omega} = \frac{\textbf{1}\{\omega\in \mathbb{B}_{\rho}(z)\}}{\mathrm{vol}(\mathbb{B}_{\rho}(z))}\triangleq \mathsf{V}_{\rho}^{-1}\textbf{1}\{\omega\in \mathbb{B}_{\rho}(z)\},\label{Eq:def:nu:z}
\end{equation}
where we denote $\mathsf{V}_{\rho}=\mathrm{vol}(\mathbb{B}_{\rho}(z))$, since the volume of $\mathbb{B}_{\rho}(z)$ is independent of the choice of $z$.
Next, we take the reference measure $\gamma_0$, a transport mapping from $\cZ$ to $\cZ$, as
\[
\diff\gamma_0(z,z') = \diff\hP(z)\diff \nu_z(z'),\quad \forall z,z'\in \cZ.
\]
Such a reference measure transports the probability mass of $\hP$ at $z$ to its norm ball $\mathbb{B}_{\rho}(z)$ uniformly.
With such a choice of $\gamma_0$, each probability mass of $\hP$ is allowed to move around its neighborhood (the norm ball of radius $\rho$) according to certain continuous probability density values, which takes account into a flexible type of adversarial attack.
With these notations, we add the following $\phi$-divergence regularization on the formulation~\eqref{Eq:formula:adv:transport}.
Notably, it ensures the worst-case distribution is absolutely continuous.
\begin{definition}[$\phi$-divergence Regularization]
Let $\phi:~\mathbb{R}\to\mathbb{R}_+\cup\{\infty\}$ be a convex lower semi-continuous function such that $\phi(1)=0, \phi(x)=\infty$ if $x<0$.
Given an optimal transport mapping $\gamma\in\cP(\cZ^2)$, define the $\phi$-divergence regularization
\begin{equation}
\bD_{\phi}(\gamma, \gamma_0) = \bE_{(z,z')\sim\gamma_0}
\left[\phi\left( 
\frac{\diff \gamma(z,z')}{\diff\gamma_0(z,z')}
\right)\right].
\tag*{\QEG}
\end{equation}
\end{definition}
For simplicity, we focus solely on the inner maximization and omit the dependence of the parameter $\theta$ on the loss $f_{\theta}(z)$.
Now, the regularized formulation of \eqref{Eq:formula:adv:transport} becomes
\begin{equation}
\tag{Primal-$\phi$-Reg}
\sup_{
\substack{\bP,\gamma
}}~\left\{
\bE_{z\sim\bP}~[f(z)] - \eta 
\bD_{\phi}(\gamma, \gamma_0):\quad
\begin{array}{l}
\text{Proj}_{1\#}\gamma=\hP, \text{Proj}_{2\#}\gamma=\bP\\
\text{ess.sup}_{\gamma}\|\zeta_1 - \zeta_2\|\le\rho
\end{array}
\right\}.
\label{Eq:primal:WDRO:phi:reg}
\end{equation}
By convention, we say for an optimal solution $(\bP_*, \gamma^*)$ to \eqref{Eq:primal:WDRO:phi:reg}, if exists, the distribution $\bP_*$ is its \emph{worst-case} distribution, and $\gamma^*$ is its \emph{worst-case} transport mapping.
Define the dual formulation of \eqref{Eq:primal:WDRO:phi:reg} as
\begin{equation}
\tag{Dual-$\phi$-Reg} 
\label{Eq:dual:WDRO:phi:reg}
\bE_{z\sim\hP}~\left[ 
\inf_{\mu\in\mathbb{R}}~\bigg\{ 
\mu + \bE_{z'\sim\nu_z}\left[ 
(\eta\phi)^*\big(f(z') - \mu\big)
\right]
\bigg\}
\right].
\end{equation}
The following summarizes the main result in this section, which shows the strong duality result, and reveals how to compute the worst-case distribution of \eqref{Eq:primal:WDRO:phi:reg} from its dual.
The proof of Theorem~\ref{Thm:strong:dual:general} is provided in Appendix~\ref{Appendix:Sec:phi:divergence}.
\begin{theorem}[Strong Duality]\label{Thm:strong:dual:general}
Assume that $\cZ$ is a measurable space, $f:~\cZ\to\mathbb{R}\cup\{\infty\}$ is a measurable function, and for every joint distribution $\gamma\in\cP(\cZ\times\cZ)$ with $\mathrm{Proj}_{1\#}\gamma=\hP$, it has a regular conditional distribution $\gamma_z$ given the value of the first marginal equals $z$. 
Then for any $\eta>0$, it holds that 
\begin{enumerate}
    \item 
$\eqref{Eq:primal:WDRO:phi:reg} = \eqref{Eq:dual:WDRO:phi:reg}$;
    \item
Additionally assume that for $\hP$-almost surely every $z$, there exists a primal-dual pair $(\mu_z^*, \zeta_z^*)$ such that 
\begin{equation}
\zeta_z^*\in\cZ_+^*,\quad 
\bE_{\nu_z}[\zeta_z^*]=1, \quad 
\zeta_z^*(\omega)=(\eta\phi)^{*}{}'\big[f(\omega) - \mu_z^*\big],
\label{Eq:condition:zeta:mu}
\end{equation}
then there exists a worst-case distribution $\bP_*$ having the density
\begin{equation*}
\frac{\diff\bP_*(\omega)}{\diff\omega} = \mathsf{V}_{\rho}^{-1}\cdot\bE_{z\sim \hP}\Big[ 
\mathrm{\bf1}\{\omega\in \mathbb{B}_{\rho}(z)\}\cdot \zeta_z^*(\omega)
\Big].
\end{equation*}
\end{enumerate}
\end{theorem}
Theorem~\ref{Thm:strong:dual:general} requires $\gamma$ having a regular conditional distribution $\gamma_z$ given the value of its first marginal equals $z$.
It simply means that for any given $z\in\mathrm{supp}\, \hP$, $\gamma_z$ is a well-defined transition probability kernel, which always holds for Polish probability space.
We refer to \citep[Chapter~5]{kallenberg1997foundations} for a detailed discussion of the regular conditional distribution.

\subsection{Discussions}\label{Sec:discussions}
In the following examples, we show that for some common choices of the function $\phi$, Condition~\eqref{Eq:condition:zeta:mu} can be further simplified such that one can obtain more analytical expressions of the worst-case distribution $\mathbb{P}_*$.

\begin{example}[Indicator Regularization]\label{Exp:indicator}
For $\alpha\in(0,1]$, consider the indicator function $\phi$ such that $\phi(x)=0$ for $x\in[0,\alpha^{-1}]$ and otherwise $\phi(x)=\infty$.
Let $\mu_z^*$ be the left-side $(1-\alpha)$-quantile of $f_{\#}\nu_z$, which is also called the value-at-risk and denoted as $\mathrm{V@R}_{\alpha,\nu_z}(f)$, and define $\zeta_z^*(\omega)=\alpha^{-1}\cdot\textbf{1}\{f(\omega)\ge \mu_z^*\}.$
One can verify that $(\mu_z^*, \zeta_z^*)$ is a primal-dual optimal solution to \eqref{Eq:condition:zeta:mu}, and therefore 
the worst-case distribution $\bP_*$ has the density 
\[
\frac{\diff\bP_*(\omega)}{\diff\omega} = (\alpha\mathsf{V}_{\rho})^{-1}\cdot\bE_{z\sim \hP}\Big[ 
\textbf{1}\Big\{\big(\omega\in \mathbb{B}_{\rho}(z)\big)\bigwedge
\big(f(\omega)\ge \mathrm{V@R}_{\alpha,\nu_z}(f)\big)
\Big\}
\Big].
\]
Define the average risk-at-risk~(AVaR) functional 
$\mathrm{AV@R}_{\alpha,\mathbb{P}}(f)=\inf\limits_{\mu}~\big\{\mu + \alpha^{-1}\bE_{z\sim \bP}[f(z) - \mu]_+\big\},$ then 
\begin{equation}
\eqref{Eq:dual:WDRO:phi:reg} = \bE_{z\sim\hP}\Big[ 
\mathrm{AV@R}_{\alpha,\nu_z}(f)
\Big].
\tag*{\QEG}
\end{equation}
\end{example}

\begin{example}[Entropic Regularization]\label{Example:ent}
Consider $\phi(x) = x\log x - x + 1, x\ge0$.
In this case, it can be verified that the primal-dual pair to Condition~\eqref{Eq:condition:zeta:mu} is unique and has closed-form expression:
\[
\mu_z^* = \eta\log
\bE_{\omega\sim \nu_z}
\Big[ 
\exp\Big( 
\frac{f(\omega)}{\eta}
\Big)
\Big],\quad 
\zeta_z^*(\omega)=\alpha_z\cdot \exp\Big( 
\frac{f(\omega)}{\eta}
\Big).
\]
where $\alpha_z:=\left(\bE_{\omega\sim \nu_z}[e^{f(\omega)/\eta}]\right)^{-1}$ is a normalizing constant.
Consequently, the worst-case distribution $\bP_*$ satisfies
\[
\frac{\diff\bP_*(\omega)}{\diff \omega} = \mathsf{V}_{\rho}^{-1}\cdot\bE_{z\sim\hP}\Big[ 
\alpha_z\cdot \exp\Big( 
\frac{f(\omega)}{\eta}
\Big)\cdot\textbf{1}\{\omega\in \mathbb{B}_{\rho}(z)\}
\Big],
\]
and Problem~\eqref{Eq:dual:WDRO:phi:reg} simplifies into an expectation of logarithm of another conditional expectation, which corresponds to the objective of conditional stochastic optimization~(CSO)~\citep{Yifan20, hu2020sample}:
\[
\eqref{Eq:dual:WDRO:phi:reg}=
\bE_{z\sim\hP}\left[
\psi_{\text{Entr}}(z;\eta)
\right],\quad 
\text{where }\quad\psi_{\text{Entr}}(z;\eta)=\eta\log
\bE_{z'\sim \nu_z}
\Big[ 
\exp\left( 
\frac{f(z')}{\eta}
\right)
\Big].
\]
Compared with the original formulation~\eqref{Eq:formula:adv}, 
the entropic regularization framework replaces the worst-case loss $\sup_{z'\in \mathbb{B}_{\rho}(z)}f(z')$ by $\psi_{\text{Entr}}(z;\eta)$.
Based on the well-known Laplace's method~(also called the log-sum-exp approximation)~\citep{butler2007saddlepoint}, this framework provides a smooth approximation of the optimal value in \eqref{Eq:formula:adv}.\QEG
\end{example}

\begin{example}[Quadratic Regularization]\label{Example:quad}
Consider $\phi(x) = \frac{1}{2}(x^2-1), x\ge0$.
By Condition~\eqref{Eq:condition:zeta:mu}, one can verify that $\mu_z^*$ is a solution to the scalar equation $\bE_{\omega\sim\nu_z}[f(\omega) - \mu_z]_+=\eta$ and $\zeta_z^*(\omega)=
\eta^{-1}\big(f(\omega) - \mu_z^*\big)_+$.
Hence, the worst-case distribution $\bP_*$ has the density 
\[
\frac{\diff\bP_*(\omega)}{\diff\omega} = (\eta\mathsf{V}_{\rho})^{-1}\cdot\bE_{z\sim \hP}\Big[ 
\textbf{1}\big\{\omega\in \mathbb{B}_{\rho}(z)
\big\}\cdot \big(f(\omega) - \mu_z^*\big)_+
\Big].
\]
Additionally, 
\begin{equation}
\tag*{\QEG}
\eqref{Eq:dual:WDRO:phi:reg} = \bE_{z\sim\hP}~\left[ 
\inf_{\mu\in\mathbb{R}}~\bigg\{ 
\frac{1}{2\eta}\bE_{z'\sim\nu_x}[f(z') - \mu]_+^2 + \frac{\eta}{2} + \mu
\bigg\}
\right].
\end{equation}
Compared to entropic regularization, this method requires solving a one-dimensional minimization problem before determining the worst-case distribution density or evaluating the dual reformulation, which can be accomplished using a bisection search algorithm. Consequently, the computational cost may be higher. However, this regularization is more stable, especially for small values of $\eta$, whereas using small $\eta$ in entropic regularization can lead to numerical errors due to the log-sum-exp operator.
Besides, quadratic regularization implicitly promotes sparsity on the support of the worst-case distribution, as the density corresponding to support point $\omega$ equals zero when $f(\omega)<\mu_z^*$ for $\hP$-almost sure $z$.
\end{example}

\begin{example}[Absolute Value Regularization]\label{Example:abs}
    Consider $\phi(x) = |x-1|, x\ge0$.
Assume 
\begin{equation}\label{Eq:f:nu_z:infty}
\|f\|_{\nu_z, \infty}:= \text{ess-sup}_{\nu_z}f=\max_{z'\in\mathbb{B}_{\rho}(z)}f(z')<\infty
\end{equation}
for $\hP$-almost surely $z$.
One can verify that 
\[
\mu_z^*= - \eta + \|f\|_{\nu_z, \infty},\quad 
\zeta_z^*(\omega) = \textbf{1}\Big\{ 
f(\omega) + 2\eta - \|f\|_{\nu_z, \infty}\ge 0
\Big\}.
\]
Hence, the worst-case distribution $\bP_*$ has the density
\[
\frac{\diff\bP_*(\omega)}{\diff \omega} = \mathsf{V}_{\rho}^{-1}\cdot\bE_{z\sim \hP}\Big[ 
\textbf{1}\Big\{\omega\in \mathbb{B}_{\rho}(z)
\bigwedge
f(\omega) + 2\eta - \|f\|_{\nu_z, \infty}\ge 0
\Big\}
\Big].
\]
In this case,
\begin{equation}
\tag*{\QEG}
\eqref{Eq:dual:WDRO:phi:reg}=
\bE_{z\sim\hP}\Big[
\|f\|_{\nu_z, \infty} - 2\eta + \bE_{z'\sim\nu_z}\Big[f(z') - \|f\|_{\nu_z, \infty} + 2\eta\Big]_+
\Big].
\end{equation}
\end{example}

\begin{example}[Hinge Loss Regularization]\label{Exp:hinge}
Consider $\phi(x) = (x-1)_+, x\ge0$.
Under the same assumption as in Example~\ref{Example:abs}, one can verify 
\[
\mu_z^*=- \eta + \|f\|_{\nu_z, \infty} ,\quad \zeta_z^*(\omega)=\textbf{1}\{f(\omega) - \mu_z^*\ge 0\}.
\]
Hence, the worst-case distribution $\bP_*$ has the density
\[
\frac{\diff\bP_*(\omega)}{\diff \omega} = \mathsf{V}_{\rho}^{-1}\cdot\bE_{z\sim \hP}\Big[ 
\textbf{1}\Big\{\omega\in \mathbb{B}_{\rho}(z)
\bigwedge
f(\omega) + \eta - \|f\|_{\nu_z, \infty}\ge 0
\Big\}
\Big].
\]
In this case,
\begin{equation}
\tag*{\QEG}
\eqref{Eq:dual:WDRO:phi:reg}=
\bE_{z\sim\hP}\Big[
\|f\|_{\nu_z, \infty} - \eta + \bE_{z'\sim \nu_z}\Big[f(z') - \|f\|_{\nu_z, \infty}+ \eta\Big]_+
\Big].
\end{equation}
\end{example}

\begin{remark}[Connections with Bayesian DRO]
Our formulation is closely related to the dual formulation of Bayesian DRO~\citep[Eq.~(2.10)]{Shapiro2023} with two major differences: 
(i) we treat the reference measure $\nu_z$ as an uniform distribution supported on $\mathbb{B}_{\rho}(z)$, while the authors therein consider a more general conditional distribution;
(ii) we fix the regularization value $\eta$, while the authors therein treat it as a Lagrangian multiplier associated with the hard constraint $\bE_{z'\sim\nu_z}\left[ 
\phi\left( 
\frac{\diff\gamma_z(z')}{\diff\nu_z(z')}
\right)
\right]\le \eta'$ for some constant $\eta'>0$.
\QEG
\end{remark}

Following the discussion in Example~\ref{Example:ent}, we are curious to study under which condition will the regularized formulation serve as the smooth approximation of the classical adversarial training formulation as the regularization value vanishes, called the \emph{consistency property}.
Proposition~\ref{Pro:consistency} gives its sufficient condition.
Its proof is provided in Appendix~\ref{Appendix:Sec:phi:divergence}.
\begin{assumption}\label{Assumption:divergence}
Assume either one of the following conditions hold:
\begin{enumerate}
    \item\label{Assumption:divergence:I}
$\lim_{t\to\infty}\frac{\phi(t)}{t}<\infty$;
    \item\label{Assumption:divergence:II}
$\lim_{t\to\infty}\frac{\phi(t)}{t}=\infty$, $\text{dom}(\phi) = \mathbb{R}_+$.
\end{enumerate}
\end{assumption}
\begin{proposition}[Consistency of Regularized Formulation]\label{Pro:consistency}
Suppose Assumption~\ref{Assumption:divergence} holds, and for any $\eta>0$ and $\hP$-almost sure $z$, the minimizer to the inner infimum problem in \eqref{Eq:dual:WDRO:phi:reg} exists and is finite.
Then, as $\eta\to0$, the optimal value of \eqref{Eq:primal:WDRO:phi:reg} converges to $\bE_{z\sim\hP}\left[ 
\max_{z'\in \mathbb{B}_{\rho}(z)}~f(z')
\right].$
\end{proposition}
Assumption~\ref{Assumption:divergence} is widely utilized in the $\phi$-divergence DRO literature~\citep{love2015phi}.
Within this context, Assumption~\ref{Assumption:divergence}\ref{Assumption:divergence:I} is referred as \emph{popping property} and the first condition of \ref{Assumption:divergence}\ref{Assumption:divergence:II} is referred as \emph{non-popping property}.
Under the non-popping property, we further assume that the domain of $\phi$ is $\mathbb{R}_+$. This assumption is crucial: the indicator function in Example~\ref{Exp:indicator} does not satisfy this assumption, and consequently, the consistency property in this case does not hold.
However, as demonstrated in Proposition~\ref{Pro:consistency}, the divergence choices in Examples~\ref{Example:ent}-\ref{Exp:hinge} do satisfy this consistency property.
As we will demonstrate in the next subsection, by taking the regularization value $\eta\to0$, the worst-case distributions from Examples~\ref{Example:ent}-\ref{Exp:hinge} indeed concentrate around the worse-case perturbation area.

\subsection{Visualization of Worst-case Distribution}\label{Sec:visualization:worst:case}

\begin{figure}[!ht]
  \begin{center}
    \includegraphics[width=0.48\textwidth]{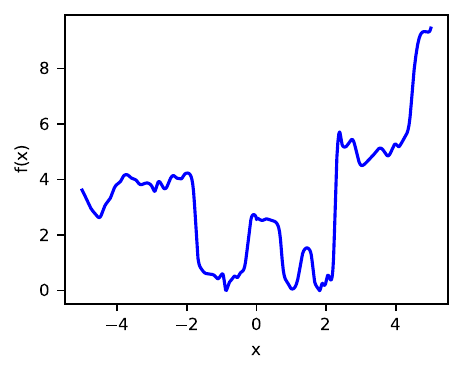}
  \end{center}
  \caption{Landscape of the $1$-dimensional objective $f(\cdot)$}
  \label{Fig:landscape}
\end{figure}

In this subsection, we display the worst-case distributions studied in Examples~\ref{Exp:indicator}-\ref{Exp:hinge} using a toy example.
We obtain the densities of these distributions by discretizing the continuous distribution $\nu_z$ using $10^4$ grid points.
The loss $f(\cdot)$ is constructed using a three-layer neural network, whose detailed configuration is provided in Appendix~\ref{Appendix:imp:detail} and landscape is displayed in Figure~\ref{Fig:landscape}. 
We take $\hP=\delta_0$, and the domain of adversarial attack as $\mathbb{B}_{\rho}(z) = [-5,5]$.
The inner maximization problem corresponding to the un-regularized formulation \eqref{Eq:formula:adv} amounts to solve the optimization problem $\max_{z\in[-5,5]}~f(z)$.
From the plot of Figure~\ref{Fig:landscape}, we can see that solving the un-regularized adversarial learning problem is highly non-trivial because the inner maximization problem contains many local maxima, and it is difficult to find the global maxima.

\begin{figure}[!ht]
    \centering
    \includegraphics[width=1\textwidth]{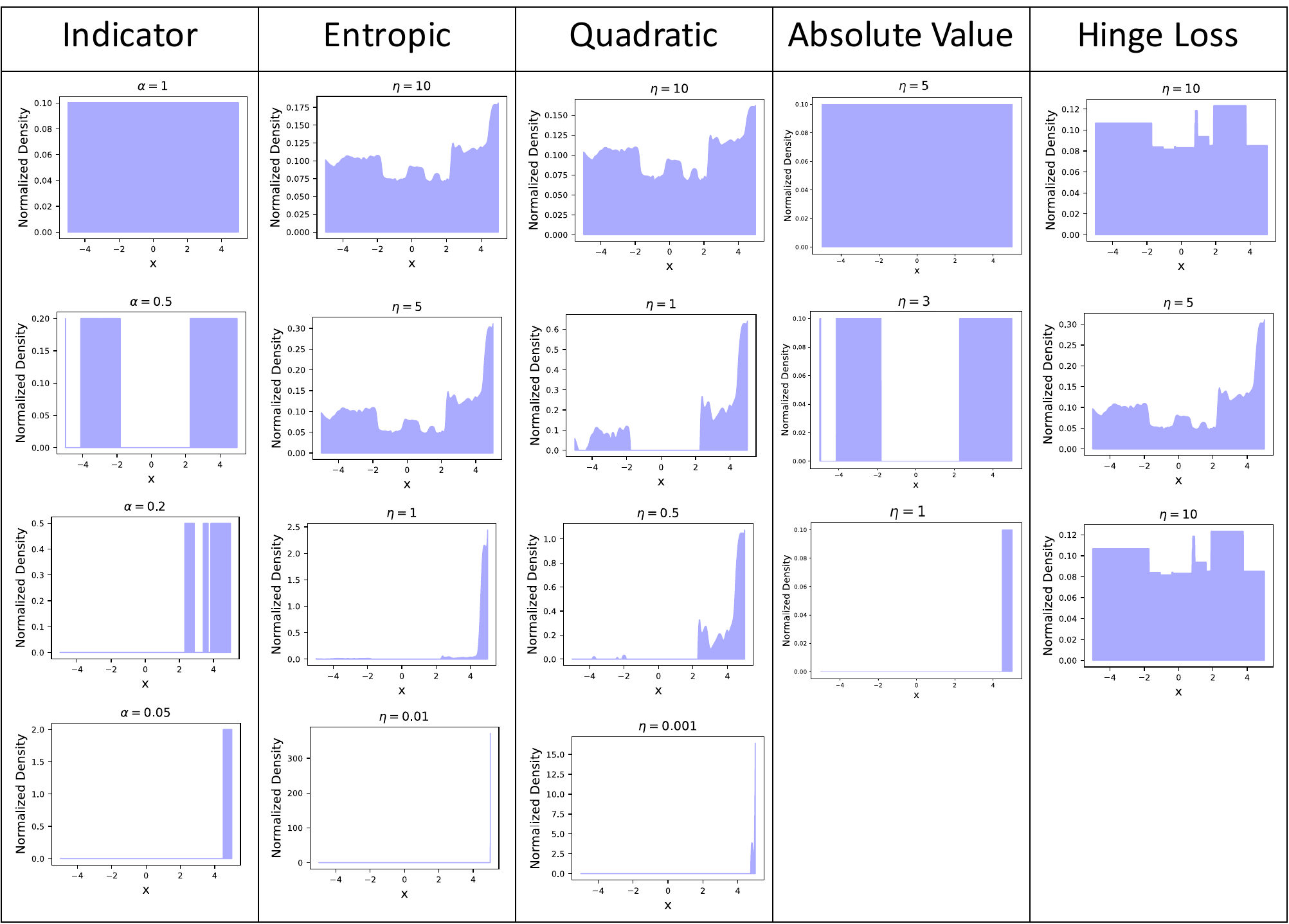}
    \caption{Worse-case distributions for different kinds of regularizations and different choices of parameters (including risk level $\alpha$ and regularization level $\eta$).}
    \label{fig:table:new}
\end{figure}

The worse-case distributions are provided in Figure~\ref{fig:table:new}, in which different columns correspond to different regularizations (i.e., indicator, entropic, quadratic, absolute value, or hinge loss), and different rows correspond to different choices of parameters: for indicator regularization, we tune the risk level $\alpha\in(0,1]$, whereas for other regularizations, we tune the regularization level $\eta$.
Our findings are summarized as follows.
\begin{enumerate}
    \item 
For indicator regularization, the worst-case distribution does not vary w.r.t. the choice of regularization value $\eta$ but the risk level $\alpha$.
From the plot, we can see that the worst-case distribution for $\alpha=1$ becomes the uniform distribution around the domain, whereas as $\alpha$ decreases, it tends to center around the worst-case perturbation area, i.e., the area close to $x=5$.
    \item
For entropic regularization, we find for large $\eta$, the worst-case distribution tends to be uniform over its support, whereas, for small $\eta$, the worst-case distribution tends to center around the worst-case perturbation area.
Compared to the plot for indicator regularization, the worst-case distribution here demonstrates greater flexibility by allowing unequal weight values across different support points.
   \item
For quadratic regularization, we obtain similar observations as in entropic regularization.
Besides, the maximum density value does not increase to infinity at such a quick rate, which demonstrates that the quadratic regularized formulation is more numerically stable to solve.
One should also notice that even for small $\eta$, the support of the worst-case distribution from entropic regularization is still the whole domain $[-5,5]$, whereas most density values can be extremely small.
In contrast, the support of that from quadratic regularization only takes a tiny proportion of the whole domain.
   \item
For absolute value and hinge loss regularizations, unlike entropic or quadratic regularization, the worst-case distributions here are constructed using histograms with equal weights assigned to different support points. This indicates that these two choices lack the flexibility needed to represent a meaningful worst-case distribution.
\end{enumerate}

Based on the discussions above, we recommend using entropic or quadratic regularization for adversarial robust learning in practice. Since these two regularizations are special cases of the Cressie-Read family of $\phi$-divergences~\citep{cressie1984multinomial}, exploring other types of $\phi$-divergences as regularizations opens an interesting avenue for further research.

\endproof

\section{Optimization Algorithm}\label{Sec:opt}
We now develop stochastic gradient-type methods to solve the proposed formulation~\eqref{Eq:ent:adv:transport}.
We first re-write it as
\begin{subequations}\label{Eq:ent:adv:trans:max:summary}
\begin{align}
\min_{\theta\in\Theta}&~
\Big\{F(\theta)=
\bE_{z\sim\hP}~\left[ 
R(\theta;z)
\right]\Big\}\label{Eq:ent:adv:trans:max:upper}
\\ 
\mbox{where}&\quad R(\theta;z)=\sup_{\gamma\in\cP(\cZ)}~
\left\{
\bE_{z'\sim\gamma}[f_{\theta}(z)] 
-
\eta\bE_{z'\sim\nu_z}
\left[ 
\phi\left( 
\frac{\diff\gamma(z')}{\diff\nu_z(z')}
\right)
\right]
\right\}, \quad \forall z.\label{Eq:ent:adv:trans:max:lower}
\end{align}
\end{subequations}
The formulation above is difficult to solve because each $z\in \mathrm{supp}\, \hP$ corresponds to a lower-level subproblem~\eqref{Eq:ent:adv:trans:max:lower}.
Consequently, Problem~\eqref{Eq:ent:adv:trans:max:summary} necessitates solving a large number of these subproblems, given that the size of $\mathrm{supp}\,\hP$ is typically large or even uncountably infinite.
In contrast, we will provide an efficient optimization algorithm whose sample complexity is \emph{near-optimal} and \emph{independent} of the size of $\mathrm{supp}\,\hP$.

Throughout this section, we assume the divergence function $\phi$ is {strongly convex} with modulus $\kappa$, which is a standard condition studied in literature.
Thus, for each $z$, the maximizer~(denoted as $\bar{\gamma}_z$) to the lower level problem \eqref{Eq:ent:adv:trans:max:lower} is unique and guaranteed to exist.
Therefore, we update $\theta$ in the outer minimization problem according to the projected stochastic gradient descent~(SGD)~\footnote{The projection $\Proj_{\Theta}(\cdot)$ can be replaced by the generalized projection mapping defined by the proximal operator. This modified algorithm is called stochastic mirror descent, which incorporates the geometry of the constraint set $\Theta$ and results in the same order of complexity bound but with (potentially) lower constant.} outlined in Algorithm~\ref{alg:Eq:ent:adv:trans:max:summary}.
\begin{algorithm}[!ht]
\caption{
Projected SGD for solving \eqref{Eq:ent:adv:trans:max:summary}
}
\label{alg:Eq:ent:adv:trans:max:summary} 
\begin{algorithmic}[1] %
\REQUIRE
{
Maximum iteration $T$, initial guess $\theta_1$, constant stepsize $\tau$
}
\FOR{$t=1,\ldots,T-1$}
\STATE{Obtain a stochastic estimator of the (sub-)gradient $\nabla F(\theta_t)$.}
\STATE{Update $
\theta_{t+1} = \Proj_{\Theta}\Big(\theta_t - \tau V(\theta_t)\Big).
$}
\ENDFOR
\\
\textbf{Output} iteration points $\{\theta_t\}_{t=1}^T$.
\end{algorithmic}
\end{algorithm}

The Step~2 of Algorithm~\ref{alg:Eq:ent:adv:trans:max:summary} requires the construction of the gradient of the objective at the upper level.
According to the Danskin's theorem, it holds that
\[
\nabla F(\theta)=\bE_{z\sim \hP}[\nabla R(\theta;z)] = 
\bE_{z\sim \hP}\mathbb{E}_{z'\sim \bar{\gamma}_z}[\nabla f_{\theta}(z')].
\]
In the following, we discuss how to construct the stochastic (sub-)gradient estimator $V(\cdot)$ in Step~2 of Algorithm~\ref{alg:Eq:ent:adv:trans:max:summary}.
More specifically, how to construct the estimator of $\nabla R(\theta;z)$.

\subsection{Gradient Estimators}
Since $\nabla R(\theta;z)$ is challenging to estimate, we first construct an approximation objective of $R(\theta;z)$ whose gradient is easier to estimate.
Denote the collection of random sampling parameters $\zeta^{\ell}:=(z, \{z_i'\}_{i\in[2^\ell]})$, where $z\sim \hP$ and $\{z_i'\}_{i\in[2^\ell]}$ are $2^\ell$ i.i.d. samples generated from distribution $\nu_z$.
Then, we define the approximation function 
\begin{equation}
F^{\ell}(\theta)=\bE_{\zeta^{\ell}}~\Big[ 
\widehat{R}\big(\theta;\{z_i'\}_{i\in[2^\ell]}\big)
\Big],
\label{Eq:F:ell}
\end{equation}
where for fixed decision $\theta$, sample $z$, and sample set $\{z_i'\}$ consisting of $m$ samples, define
\begin{equation}\label{Eq:hR:z:z}
\widehat{R}\big(\theta;\{z_i'\}\big)=\max_{\gamma\in\Delta^{m}}~\left\{ 
\sum_{i\in[m]}\gamma_if_{\theta}(z_i') - \frac{\eta}{m}\sum_{i\in[m]}\phi(m\gamma_i)
\right\}.
\end{equation}
The function $\widehat{R}\big(\theta;\{z_i'\}\big)$ can be viewed as the optimal value of the $\phi$-divergence DRO with discrete empirical distribution supported on $\{z_i'\}$. 
The high-level idea of the approximation function $F^{\ell}(\theta)$ is to replace the lower-level problem \eqref{Eq:ent:adv:trans:max:lower}, the $\phi$-divergence DRO with continuous reference distribution, using another $\phi$-divergence DRO with its empirical reference distribution.
As the number of samples of the empirical reference distribution goes to infinity, one can expect that $F^{\ell}(\theta)$ approximates the original objective $F(\theta)$ with negligible error.

It is easy to generate gradient estimators for the approximation function $F^{\ell}(\theta)$.
For fixed random sampling parameter $\zeta^{\ell}$, assume there exists an oracle that returns $\widetilde{\gamma}_{n_1:n_2}$ as \emph{near-optimal} probability mass values for $\widehat{R}\big(\theta;\{z_i'\}_{i\in[n_1:n_2]}\big)$, and define the gradient
\begin{equation}
\nabla\widetilde{R}\big(\theta;\{z_i'\}_{i\in[n_1:n_2]}\big)
=
\sum_{i\in[n_1:n_2]}(\widetilde{\gamma}_{n_1:n_2})_i\nabla_{\theta}f_{\theta}(z_i').\label{Eq:tilde:R}
\end{equation}
Due to the near-optiamlity of $\widetilde{\gamma}_{n_1:n_2}$, it holds that $\nabla\widetilde{R}\big(\theta;\{z_i'\}_{i\in[n_1:n_2]}\big)\approx \nabla\widehat{R}\big(\theta;\{z_i'\}_{i\in[n_1:n_2]}\big)$.
Next, we define
\begin{align}
g^{\ell}(\theta, \zeta^\ell)&=\nabla\widetilde{R}\big(\theta;\{z_i'\}_{i\in[1:2^\ell]}\big)\label{Eq:g:ell}\\
G^{\ell}(\theta, \zeta^\ell)&=\nabla\widetilde{R}\big(\theta;\{z_i'\}_{i\in[1:2^\ell]}\big)
-
\frac{1}{2}
\Big[ 
\nabla\widetilde{R}\big(\theta;\{z_i'\}_{i\in[1:2^{\ell-1}]}\big)
+
\nabla\widetilde{R}\big(\theta;\{z_i'\}_{i\in[2^{\ell-1}+1:2^\ell]}\big)
\Big].
\label{Eq:G:ell}
\end{align}
Now, we list two choices of gradient estimators at a point $\theta$:\\
\begin{subequations}
\noindent{\bf Stochastic Gradient~(SG) Estimator. }
For fixed level $L$, generate $\nout_L$ i.i.d. copies of $\zeta^L$, denoted as $\{\zeta_i^L\}$.
Then construct
\begin{equation}
V^{\text{SG}}(\theta)=\frac{1}{\nout_L}\sum_{i=1}^{\nout_L}g^{L}(\theta, \zeta_i^L).
\label{Eq:SG:estimator}
\end{equation}
\noindent{\bf Randomized Truncation MLMC~(RT-MLMC) Estimator. }
For fixed level $L$, generate $\nout_L$ i.i.d. random levels following the \emph{truncated geometric distribution} $\mathbb{P}(\widehat{L}=\ell)=\frac{2^{-\ell}}{2-2^{-L}}, \ell=0,\ldots,L$, denoted as $\widehat{L}_1,\ldots,\widehat{L}_{\nout_L}$.
Then construct
\begin{equation}
V^{\text{RT-MLMC}}(\theta)=\frac{1}{\nout_L}\sum_{i=1}^{\nout_L}
\mathbb{P}(\widehat{L}=\widehat{L}_i)^{-1}\cdot G^{\widehat{L}_i}(\theta, \zeta_i^{\widehat{L}_i}).\label{Eq:RTMLMC:estimator}
\end{equation}
\end{subequations}
The SG estimator is a conventional approach to estimate $\nabla F^{\ell}(\theta)$.
Instead, the RT-MLMC estimator has the following attractive features:
\begin{enumerate}
    \item 
It constitutes a gradient estimator of the approximation function $F^L(\theta)$ with a small bias.
More specifically, RT-MLMC and SG estimators have the same bias:
\begin{align*}
\bE[V^{\text{RT-MLMC}}(\theta)]
&=
\bE_{\widehat{L}_1}\left[ 
\frac{1}{\mathbb{P}(\widehat{L}=\widehat{L}_1)}
\bE_{\zeta^{\widehat{L}_1}}[G^{\widehat{L}_1}(\theta, \zeta^{\widehat{L}_1})]
\right]
=
\sum_{\ell=0}^L\mathbb{P}(\widehat{L}=\ell)\cdot
\left[ 
\frac{1}{\mathbb{P}(\widehat{L}=\ell)}\bE_{\zeta^{\ell}}[G^{\ell}(\theta, \zeta^{\ell})]
\right]\\
&=
\sum_{\ell=0}^L\bE_{\zeta^{\ell}}[G^{\ell}(\theta, \zeta^{\ell})]
=
\bE_{z\sim\hP}
\bE_{\{z_i'\}_{i\in[2^\ell]}\sim \nu_z}~\Big[ 
\nabla\widetilde{R}\big(\theta;\{z_i'\}_{i\in[2^\ell]}\big)
\Big]=\bE[V^{\text{SG}}(\theta)],
\end{align*}
and the bias vanishes quickly as $L\to\infty$.
\item
Since $\nabla\widetilde{R}\big(\theta;\{z_i'\}_{i\in[1:2^\ell]}\big),\nabla\widetilde{R}\big(\theta;\{z_i'\}_{i\in[1:2^{\ell-1}]}\big)$, and $\nabla\widetilde{R}\big(\theta;\{z_i'\}_{i\in[2^{\ell-1}+1:2^\ell]}\big)$ are generated using the same random sampling parameters $\zeta^{\ell}$, they are highly correlated, which indicates the stochastic estimator $G^{\ell}(\theta,\zeta^{\ell})$ defined in \eqref{Eq:G:ell} has small second-order moment and variance thanks to the control variate effect~\citep{rubinstein1985efficiency}, making it a suitable recipe for gradient simulation.
\item
The construction of SG estimator requires generating $\nout_L\cdot 2^L=\cO(2^L)$ samples, while the (expected) number of samples for RT-MLMC is $\nout_L\cdot \frac{L}{2 -2^{-L}}=\cO(L)$.
As a result, the computation of RT-MLMC estimator is remarkably smaller than that of SG estimator.
\end{enumerate}

\subsection{Solving penalized \texorpdfstring{$\phi$}{abc}-divergence DRO with finite support}\label{Sec:oracle}
In the last subsection, it is assumed that one has the oracle for solving the generic penalized $\phi$-divergence DRO with $m$ support points $\{f_1,\ldots,f_m\}$:
\begin{equation}
\mathcal{R} = \max_{\gamma\in\Delta^m}~\left\{ 
\sum_{i\in[m]}\gamma_if_i - \frac{\eta}{m}\sum_{i\in[m]}\phi(m\gamma_i)
\right\}.
\label{Eq:expression:R}
\end{equation}
The formulation~\eqref{Eq:hR:z:z} is a special case of this problem by taking $f_i=f_{\theta}(z_i'), \forall i$.
In the following, we provide an algorithm that returns returns the optimal solution to \eqref{Eq:expression:R} up to precision $\epsilon$.
We write its Lagrangian reformulation as 
\[
\min_{\mu}\max_{\gamma\in\mathbb{R}_+^m}~\Big\{ 
\mathcal{L}(\lambda, \gamma) = \sum_{i\in[m]}\gamma_if_i - \frac{\eta}{m}\sum_{i\in[m]}\phi(m\gamma_i) + \mu\Big( 
1 - \sum_{i\in[m]}\gamma_i
\Big)
\Big\}.
\]
Based on this minimax formulation, we present an efficient algorithm that finds a near-optimal primal-dual solution to \eqref{Eq:expression:R} in Algorithm~\ref{alg:Eq:expression:R}, whose complexity analysis is presented in Proposition~\ref{Pro:complexity:alg:Eq:expression:R}.
The complexity is quantified as the number of times to query samples $f_1,\ldots,f_m$.
Its proof is provided in Appendix~\ref{Appendix:Sec:opt}.
\begin{algorithm}[htb]
\caption{
Bisection search for solving \eqref{Eq:expression:R}
}
\label{alg:Eq:expression:R} 
\begin{algorithmic}[1] %
\REQUIRE
{
Interval $[\underline{\mu}, \overline{\mu}]$, maximum iteration $T$, constant $K=\lim_{s\to0+}\phi'(s)$.
}
\FOR{$t=1,\ldots,T$}
\STATE{Update $\mu = \frac{1}{2}(\underline{\mu} + \overline{\mu})$}
\STATE{Obtain index set $\mathcal{N}=\Big\{i\in[m]:~f_i\le \mu + \eta K\Big\}$.}
\STATE{Compute $h(\mu):=\frac{1}{m}\sum_{i\in[m]\setminus\mathcal{N}}(\phi')^{-1}\Big( 
\frac{f_i - \mu}{\eta}
\Big)-1$.}
\STATE{Update $\overline{\mu}=\mu$ if $h(\mu)\le 0$ and otherwise $\underline{\mu}=\mu$.}
\ENDFOR
\STATE{Obtain $\gamma^*$ such that $\gamma_i=0$ if $i\in\mathcal{N}$ and otherwise $\gamma_i=\frac{1}{m}(\phi')^{-1}(\frac{f_i - \mu}{\eta})$.}
\\
\textbf{Output} the estimated primal-dual optimal solution $(\gamma, \mu)$ and estimated optimal value $h(\mu)$.
\end{algorithmic}
\end{algorithm}

\begin{proposition}[Performance Guarantees of Algorithm~\ref{alg:Eq:expression:R}]\label{Pro:complexity:alg:Eq:expression:R}
Fix the precision $\epsilon>0$.
Suppose 
we choose hyper-parameters in Algorithm~\ref{alg:Eq:expression:R} as
\begin{align*}
T=
\frac{1}{2}\log_2\left(
\frac{\varrho^2}{2\kappa\eta}\cdot\frac{1}{\epsilon}
\right)
=
\cO(\log\frac{1}{\epsilon}),\quad
\underline{\mu}&=\underline{f},\qquad 
\overline{\mu}=\left\{ 
\begin{aligned}
\overline{f} - \eta(\underline{f} - \overline{f}),&\quad \text{if $\phi'(s)\to-\infty$ as $s\to0+$,}\\
\overline{f} - \eta K,&\quad \text{if $\phi'(s)\to K>-\infty$ as $s\to0+$,}
\end{aligned}
\right.
\end{align*}
where $\varrho = \overline{\mu} - \underline{\mu}$, $\underline{f}=\min_{i\in[m]}f_i$, and $\overline{f}=\max_{i\in[m]}f_i$.
As a consequence, Algorithm~\ref{alg:Eq:expression:R} finds a primal-dual solution to \eqref{Eq:expression:R} such that 
\begin{enumerate}
    \item%
the estimated objective value $\widetilde{\mathcal{R}}$ satisfies $|\widetilde{\mathcal{R}} - \mathcal{R}|\le \epsilon$;
    \item\label{Pro:complexity:alg:Eq:expression:R:III}
the difference between the estimated primal solution $\gamma$ and the optimal primal solution $\gamma^*$ is bounded: $\|\gamma - \gamma^*\|_{\infty}\le \frac{1}{m}\sqrt{\frac{2\epsilon}{\kappa\eta}}$;
    \item
the distance between the estimated dual solution $\mu$ and the set of optimal dual solutions $\mathcal{S}^*$ is bounded: $\textsf{D}(\mu, \mathcal{S}^*)\le \varrho\cdot 2^{-T}=(2\eta\kappa\epsilon)^{1/2}$;
    \item
its worst-case computational cost is $\cO(mT)=\cO(m\log\frac{1}{\epsilon})$.
\end{enumerate}
\end{proposition}

Most literature (such as \citep{namkoong2016stochastic, ghosh2018efficient, hu2021biasvar}) provided algorithms for solving hard-constrained $\phi$-divergence DRO problem, but they did not study how to extend their framework for penalized $\phi$-divergence DRO in \eqref{Eq:expression:R}.
One exception is that \citet{levy2020large} mentioned this problem can be solved using bisection search as in Algorithm~\ref{alg:Eq:expression:R} but did not provide detailed parameter configurations.



\begin{remark}[Near-Optimality of Algorithm~\ref{alg:Eq:expression:R}]
For some special choices of the divergence function $\phi$, the optimal solution of problem~\eqref{Eq:hR:z:z} can be obtained with closed-form solution, such as the entropy function $\phi(s) = s\log s - s + 1$.
However, Algorithm~\ref{alg:Eq:expression:R} is applicable to solving problem~\eqref{Eq:hR:z:z} for general choices of the divergence function.
When considering $\phi(s) = s\log s - s + 1$, the optimal solution to \eqref{Eq:expression:R} becomes
\[
\gamma^*_i = \frac{e^{f_i/\eta}}{\sum_{i\in[m]}e^{f_i/\eta}},\quad\forall i\in[m].
\]
Computing this optimal solution requires computational cost at least $\Omega(m)$.
Compared with the complexity in Proposition~\ref{Pro:complexity:alg:Eq:expression:R}, Algorithm~\ref{alg:Eq:expression:R} is a near-optimal choice because it matches the lower bound up to negligible constant $\cO(\log\frac{1}{\epsilon})$.
\QEG
\end{remark}

\subsection{Complexity Analysis}\label{Sec:complexity}
In this subsection, we provide the convergence analysis of our projected SGD algorithm using SG and RT-MLMC gradient estimators.
Throughout this subsection, the computational cost is quantified as the number of times to generate samples $z\sim \hP$ or samples $z'\sim\nu_z$ for any $z\in\mathrm{supp}\,\hP$.
We consider the following assumptions regarding the loss function.
\begin{assumption}[Loss Assumptions]\label{Assumption:throughout:loss:updated}
\begin{enumerate}
    \item\label{Assumption:throughout:loss:cvx}(Convexity): The loss $f_{\theta}(z)$ is convex in $\theta$.
    \item\label{Assumption:throughout:loss:lip}(Lipschitz Continuity): For fixed $z$ and $\theta_1,\theta_2$, it holds that $|f_{\theta_1}(z) - f_{\theta_2}(z)|\le L_f\|\theta_1 - \theta_2\|_2$.
    \item\label{Assumption:throughout:loss:bound}(Boundedness): for any $z$ and $\theta$, it holds that $0\le f_{\theta}(z)\le B$.
    \item\label{Assumption:throughout:loss:smooth}(Lipschitz Smoothness): The loss function $f_{\theta}(z)$ is continuously differentiable and for fixed $z$ and $\theta_1,\theta_2$, it holds that 
   $\|\nabla f_{\theta_1}(z) - \nabla f_{\theta_2}(z)\|_2\le S_f\|\theta_1-\theta_2\|_2$.
\end{enumerate}
\end{assumption}

\subsubsection{Nonsmooth Convex Loss}
To analyze the convergence rate, we rely on the following technical assumptions.
\begin{assumption}\label{Assumption:idf:chi2}
\begin{enumerate}
    \item\label{Assumption:chi2}
For any data points $\{f_1,\ldots,f_m\}$, the optimal probability vector $\gamma^*$ in the data-driven penalized $\phi$-divergence DRO problem~\eqref{Eq:expression:R} satisfies $\mathcal{D}_{\mathcal{X}^2}(\gamma^*, \frac{1}{m}\mathbf{1}_m)\le C$.
    \item\label{Assumption:idf}
For each $\theta\in\Theta$ and $z\in \mathrm{supp}\,\hP$, the inverse cdf of the random variable $(f_{\theta})_{\#}\nu_z$ is $G_{\mathrm{idf}}$-Lipschitz.
\end{enumerate}
\end{assumption}
Assumption~\ref{Assumption:idf:chi2} is relatively mild and has originally been proposed in \citep[Assumption~A1]{levy2020large} to investigate the complexity of solving standard $\phi$-divergence DRO.
Assumption~\ref{Assumption:idf:chi2}\ref{Assumption:chi2} holds by selecting proper divergence function $\phi$, such as quadratic or entropy function in Examples~\ref{Example:quad} and \ref{Example:ent}.
As long as the probability density of $f_{\theta}(Z)$ is lower bounded by $\Upsilon$ within its support, Assumption~\ref{Assumption:idf:chi2}\ref{Assumption:idf} holds with $G_{\mathrm{idf}}=\Upsilon^{-1}$.
Now, we derive statistics of SG and RT-MLMC estimators.
\begin{proposition}[Bias/Second-order-Moment/Cost of SG and RT-MLMC Estimators]\label{Pro:bias:var:RTMLMC}
Fix the precision $\epsilon>0$.
Suppose Assumption 
\ref{Assumption:idf:chi2}\ref{Assumption:chi2} holds, and during the construction of SG/RT-MLMC estimators, one query Algorithm~\ref{alg:Eq:expression:R} with optimality gap controlled by $\epsilon$.
Then it holds that
\begin{enumerate}
    \item 
(Bias): 
Suppose, in addition, Assumption~\ref{Assumption:idf:chi2}\ref{Assumption:idf} holds, then
$\bE[V^{\SG}(\theta)]=\bE[V^{\RTMLMC}(\theta)]=\nabla\widetilde{F}(\theta)$, where $|\widetilde{F}(\theta) - F(\theta)|\le \epsilon + G_{\mathrm{idf}}\cdot 2^{-L}$.
    \item
(Second-order Moment): 
\begin{align*}
\bE\|V^{\SG}(\theta)\|^2_2&\le 2L_f^2\left[1 + (2\epsilon)/(\kappa\eta)\right],\quad
\bE\|V^{\RTMLMC}(\theta)\|^2_2\le \frac{96L_f^2}{\kappa\eta}\cdot (2^L\epsilon) +6(L+1)L_f^2C.
\end{align*}
    \item
(Variance): 
\begin{align*}
\Var[V^{\SG}(\theta)]&\le \frac{2L_f^2\left[1 + (2\epsilon)/(\kappa\eta)\right]}{\nout_L},~~
\Var[V^{\RTMLMC}(\theta)]\le \frac{1}{\nout_L}\left[
\frac{96L_f^2}{\kappa\eta}\cdot (2^L\epsilon)  +6(L+1)L_f^2C
\right].
\end{align*}
    \item
(Cost): Generating a single SG estimator requires cost $\cO(\nout_L\cdot 2^L\log\frac{1}{\epsilon})$, whereas generating a single RT-MLMC estimator requires expected cost $\cO(\nout_L\cdot L\log\frac{1}{\epsilon})$.
\end{enumerate}
\end{proposition}
Let the estimated solution returned by the projected SGD algorithm be $\widetilde{\theta}_{1:T}=\frac{1}{T}\sum_{t=1}^T\theta_t$.
Based on Proposition~\ref{Pro:bias:var:RTMLMC}, we derive complexity bounds for solving \eqref{Eq:ent:adv:trans:max:summary} when the loss function is convex and Lipschitz continuous.
We formalize our results in the following theorem.
Its proof is provided in Appendix~\ref{Appendix:Sec:opt}.

\begin{theorem}[Complexity for Nonsmooth Convex Loss]\label{Thm:nonsmooth:cvx}
Suppose Assumptions 
\ref{Assumption:throughout:loss:updated}\ref{Assumption:throughout:loss:cvx}, \ref{Assumption:throughout:loss:updated}\ref{Assumption:throughout:loss:lip}, 
\ref{Assumption:idf:chi2} hold, 
and $\delta>0$ is a sufficiently small precision level.
During the construction of SG/RT-MLMC estimators, let the optimality gap of querying Algorithm~\ref{alg:Eq:expression:R} controlled by $\epsilon=\frac{\delta}{8}$, and specify the hyper-parameters of SGD algorithm with SG or RT-MLMC estimators as in Table~\ref{tab:summary:cvx}.
As a result, 
\begin{enumerate}
    \item(SG Estimator)
The SGD algorithm with SG estimator finds a $\delta$-optimal solution to \eqref{Eq:ent:adv:trans:max:summary} with computational cost $\mathcal{O}(T\cdot \nout_L 2^L\log\frac{1}{\epsilon})=\cO(\delta^{-3}\log\frac{1}{\delta})$.
\item(RT-MLMC Estimator)
The SGD algorithm with RT-MLMC estimator finds a $\delta$-optimal solution to \eqref{Eq:ent:adv:trans:max:summary} with computational cost $\mathcal{O}(T\cdot \nout_L L\log\frac{1}{\epsilon})=\cO(\delta^{-2}(\log\frac{1}{\delta})^4)$.
\end{enumerate}
\begin{table}[ht!]
	\caption{
	Hyper-parameters used in the projected SGD algorithm with SG/RT-MLMC gradient estimators for nonsmooth convex loss.
	}
	\label{tab:summary:cvx}
	\begin{center}%
{
			\begin{tabular}{
>{\centering\arraybackslash}m{2cm}|>{\centering\arraybackslash}m{3cm}|>{\centering\arraybackslash}m{3cm}|>{\centering\arraybackslash}m{3cm}|>{\centering\arraybackslash}m{3cm}
   }
			\toprule%
\textbf{Method} & Batch Size $\nout_L$ & Max Level $L$ & Max Iteration $T$ & Step Size $\gamma$\\[8pt]
\hline
\textbf{SG} & $1$ & $\log\frac{8G_{\text{idf}}}{\delta}$ & $\cO(1/\delta^2)$ & $\cO(\delta)$
\\[8pt]
\hline
\textbf{RT-MLMC} & $1$ & $\log\frac{8G_{\text{idf}}}{\delta}$ & $\cO((\log1/\delta)^2/\delta^2)$ & $\cO((\log1/\delta)^{-2}\delta)$
\\[8pt]
   \bottomrule
\end{tabular}
}
	\end{center}
\end{table}
\end{theorem}


\subsubsection{
Smooth Nonconvex Loss
}
When the loss $f_{\theta}(z)$ is nonconvex in $\theta$, we focus on finding the near-stationary point of \eqref{Eq:ent:adv:trans:max:summary} instead.
The key in this part is to build the bias between our gradient estimator in \eqref{Eq:SG:estimator} or \eqref{Eq:RTMLMC:estimator}  and the true gradient of the objective.
Unfortunately, such a result for general choice of $\phi$-divergence regularization is hard to show.
In this part, we only investigate the convergence behavior of entropic regularization~(see Example~\ref{Example:ent}).
In such a case, we have the closed-form expression regarding the optimal solution of the lower-level problem~\eqref{Eq:ent:adv:trans:max:lower}, and therefore, Problem~\eqref{Eq:ent:adv:trans:max:summary} can be reformulated as
\begin{equation}\label{Eq:ent:adv:trans:max:summary:transfer}
\min_{\theta\in\Theta}~
\Big\{F(\theta)=
\bE_{z\sim\hP}~\left[ 
\eta\log\bE_{z'\sim\nu_z}~\left[ 
\exp\left( 
\frac{f_{\theta}(z')}{\eta}
\right)
\right]
\right]\Big\}.
\end{equation}
Similarly, the approximation function $F^{\ell}$ defined in \eqref{Eq:F:ell} becomes
\begin{equation}\label{Eq:stat:robust:formula:approximation}
F^{\ell}(\theta) = \bE_{\zeta^{\ell}}\left[ \eta\log\left(
\frac{1}{2^{\ell}}\sum_{i\in[2^{\ell}]}\exp\left( 
\frac{f_{\theta}(z_i')}{\eta}
\right)
\right)
\right].
\end{equation}
In this case, we do not use \eqref{Eq:g:ell} or \eqref{Eq:G:ell} but adopt the following way to construct the random vectors $g^{\ell}(\theta,\zeta^{\ell})$ and $G^{\ell}(\theta,\zeta^{\ell})$:
define
\[
U_{n_1:n_2}(\theta,\zeta^{\ell}) = \eta\log\left(
\frac{1}{n_2-n_1+1}\sum_{j\in[n_1:n_2]}\exp\left( 
\frac{f_{\theta}(z_j')}{\eta}
\right)
\right).
\]
and construct 
\begin{align}
g^{\ell}(\theta,\zeta^{\ell})&=\nabla_{\theta}U_{1:2^{\ell}}(\theta,\zeta^{\ell}),
\\
   G^{\ell}(\theta,\zeta^{\ell})&=\nabla_{\theta}\left[ 
U_{1:2^{\ell}}(\theta,\zeta^{\ell}) - \frac{1}{2}U_{1:2^{\ell-1}}(\theta,\zeta^{\ell})
- \frac{1}{2}U_{2^{\ell-1}+1:2^{\ell}}(\theta,\zeta^{\ell})
\right].\label{Eq:expression:G}
\end{align}
The following theorem presents the complexity of obtaining a $\delta$-stationary point for our projected SGD algorithm using either SG or RT-MLMC estimator.
Its proof is provided in Appendix~\ref{Appendix:Sec:opt}.
\begin{theorem}[Complexity for Smooth Nonconvex Loss]\label{Theorem:complexity:BSMD}
Under Assumptions~\ref{Assumption:throughout:loss:updated}\ref{Assumption:throughout:loss:lip}, \ref{Assumption:throughout:loss:updated}\ref{Assumption:throughout:loss:bound}, and
\ref{Assumption:throughout:loss:updated}\ref{Assumption:throughout:loss:smooth},
with properly chosen hyper-parameters of the RT-MLMC estimator as in Table~\ref{tab:summary:tests}, the following results hold:
\begin{enumerate}
    \item\label{Thm:S:noncvx}
(Smooth Nonconvex Optimization)
The computation cost of RT-MLMC scheme for finding $\epsilon$-stationary point is of $\tO(\epsilon^{-4})$ with memory cost $\tO(\epsilon^{-2})$.
    \item
(Unconstrainted Smooth Nonconvex Optimization)
Additionally assume the constraint set $\Theta=\mathbb{R}^{d_{\theta}}$, then the memory cost of RT-MLMC improves to $\tO(1)$.
\end{enumerate}
\begin{table}[ht!]
	\caption{
 Hyper-parameters used in the projected SGD algorithm with SG/RT-MLMC gradient estimators for smooth nonconvex loss.
	}
	\label{tab:summary:tests}
	\small
	\begin{center}%
{
\setlength\extrarowheight{4pt}
\renewcommand{\arraystretch}{0.7}
			\begin{tabular}{c|m{5cm}|m{5cm}}
			\toprule%
\textbf{Scenarios} & Hyper-parameters & Comp./Memo.\\[4pt]
\hline
\multirow{2}{*}{
$
\begin{array}{c}
\textbf{Smooth Nonconvex}\\
\textbf{Optimization}
\end{array}$} & 
$L = \cO(\log\frac{1}{\epsilon^2}), \quad T=\tO(\epsilon^{-2})$ & 
$\mbox{Comp.}=\cO(T(\nout_LL))=\tO(\epsilon^{-4})$
\\[4pt]
&  $\nout_L=\tO(\epsilon^{-2}),\quad \gamma=O(1)$ & 
$\mbox{Memo.}=\cO(\nout_LL)=\tO(\epsilon^{-2})$
\\[4pt]
\hline
\multirow{2}{*}{
$
\begin{array}{c}
\textbf{Unconstrainted Smooth}\\
\textbf{Nonconvex Optimization}
\end{array}$} & 
$L = \cO(\log\frac{1}{\epsilon^2}), \quad T=\tO(\epsilon^{-4})$ & 
$\mbox{Comp.}=\cO(T(\nout_LL))=\tO(\epsilon^{-4})$
\\[4pt]
&  $\nout_L=\cO(1),\quad \gamma=\tO(\epsilon^2)$ & 
$\mbox{Memo.}=\cO(\nout_LL)=\tO(1)$
\\[4pt]
   \bottomrule
\end{tabular}
}
	\end{center}
\end{table}
\end{theorem}

\begin{remark}[Comparision with Sinkhorn DRO]
Sinkhorn DRO~\citep{wang2021sinkhorn} introduces entropic regularization to the ambiguity set constructed using the $p$-Wasserstein distance, resulting in a dual reformulation that closely resembles \eqref{Eq:ent:adv:trans:max:summary:transfer}, with the key modification of replacing the uniform distribution $\nu_z$ with certain kernel probability distributions. The primary distinction lies in the fact that the authors of the original work provide only the SG or RT-MLMC estimator for nonsmooth convex loss, whereas we extend their analysis to the smooth nonconvex loss setting. A crucial aspect of the convergence analysis is that for the constrained smooth nonconvex case, as highlighted in \citep{ghadimi2016mini}, a large mini-batch size $\nout_L$ is required at each iteration to estimate the gradient with sufficiently small variance to ensure convergence. In contrast, for the unconstrained case, a mini-batch size $\nout_L=\cO(1)$ is sufficient.\QEG
\end{remark}
\begin{remark}[Comparison with $\infty$-WDRO]
When solving~\eqref{Eq:formulation:wass}, the involved subproblems are finding the global optimal value of the supremum $\sup_{z\in\mathbb{B}_{\rho}(x)}~f(z)$ for $x\in\mathrm{supp}\, \hP$, which are computationally challenging in general.
Various heuristics~\citep{kurakin2016adversarial, goodfellow2014explaining, madry2017towards} have been proposed to approximately solve it by replacing $f(z)$ with its linear approximation $f(x)+\nabla f(x)\trans z$.
It is worth noting that such an approximation is not accurate, especially when the radius $\rho$ of domain set $\mathbb{B}_{\rho}(x)$ is moderately large, which corresponds to large adversarial perturbation scenarios.
For example, for the loss $f(z)$ depicted in Figure~\ref{Fig:landscape}, its linear approximation around $x=0$ will yield a wrong global maximum estimate.
In contrast, we proposed stochastic gradient methods to solve the regularized formulation~\eqref{Eq:ent:adv:transport} with provable convergence guarantees, which avoids solving such a hard maximization subproblem.
Numerical comparisons in Section~\ref{Sec:supervised} also suggest that our method outperforms those heuristics when adversarial perturbations are moderately large.\QEG
\end{remark}


\section{Regularization Effects of Regularized Adversarial Robust Learning}
\label{Sec:reg}
In this section, we provide an interpretation on how our proposed formulation~\eqref{Eq:ent:adv:transport} works by showing its close connection to the regularized ERM problem:
\[
\min_{\theta\in\Theta}~\bE_{z\sim\hP}[f_{\theta}(z)] + \mathcal{R}(f_{\theta}; \rho,\eta)
\]
for certain regularization $\mathcal{R}(f_{\theta}; \rho,\eta)$.
As we focus on small-perturbation attacks, it is assumed that $\rho, \eta\to0$. 
Also, we omit the dependence of $f_{\theta}$ on $\theta$ for simplicity.
Subsequently, we derive the regularization effects of \eqref{Eq:ent:adv:transport} by considering different scaling of $\rho$ and $\eta$.
To begin with, we define regularizer $\mathcal{E}$ as the difference between regularized robust loss in \eqref{Eq:primal:WDRO:phi:reg} and non-robust loss:
\begin{equation}
\mathcal{E}_{\hP}(f;\rho,\eta) = \mbox{\normalfont Optval}\eqref{Eq:primal:WDRO:phi:reg} - \mathbb{E}_{\hP}[f].\label{Eq:regularizer:residual}
\end{equation}
Besides, we define the following regularizations.
Let $\beta$ be the uniform probability distribution supported on $\mathbb{B}_1(0)$, and define
\begin{subequations}
\begin{align}
\mathcal{R}_1(f; \rho,\eta)&=
\rho\cdot\mathbb{E}_{z\sim\hP}\left[ 
\inf_{\mu\in\mathbb{R}}\left\{ 
\mu + \frac{1}{C}\mathbb{E}_{b\sim \beta}\left[ 
\phi^*\Big( 
C\cdot(\nabla f(z)\trans b - \mu)
\Big)
\right]
\right\}
\right],
\label{Eq:R:1}\\
\mathcal{R}_2(f; \rho,\eta)&=
\rho\cdot\bE_{z\sim\hP}\Big[\|\nabla f(z)\|_*\Big],
\label{Eq:R:2}
\\
\mathcal{R}_3(f; \rho,\eta)&=
\frac{\rho^2}{2\eta\cdot \phi''(1)}\cdot \bE_{z\sim \hP}\Big[ 
\Var_{b\sim\beta}[\nabla f(z)\trans b]
\Big],
\label{Eq:R:3}
\end{align}
\end{subequations}
where $C>0$ is some constant to be specified.
These three regularizations correspond to the asymptotic approximations of the regularizer under three different scaling regions of $\rho$ and $\eta$.

We impose the following smoothness assumption on the loss $f$, which is a standard technique assumption when investigating the regularization effects of Wasserstein DRO~\citep{gao2022wasserstein}.
\begin{assumption}[Smooth Loss]\label{Assumption:f:smooth}
The loss $f(\cdot)$ is smooth with respect to the norm $\|\cdot\|$ such that
$
\|\nabla f(x) - \nabla f(x')\|_*\le S(x)\cdot \|x-x'\|,\quad\forall x,x'.\label{Eq:condition:smooth:norm}
$
\end{assumption}
We now present our main result in this section in Theorem~\ref{Theorem:ref:eff}, whose proof is provided in Appendix~\ref{Appendix:Sec:reg}.
\begin{theorem}[Regularization Effects]\label{Theorem:ref:eff}
Suppose Assumption~\ref{Assumption:f:smooth} holds, and $\rho\to0,\eta\to0$, we have the following results.
\begin{enumerate}
    \item\label{Proposition:3:smooth}(OCE Regularization) When $\rho/\eta\to C\in(0,\infty)$, it holds that 
\[
\Big| 
\mathcal{E}_{\hP}(f;\rho,\eta) -
\mathcal{R}_1(f; \rho,\eta)
\Big|=o(\rho).
\]
    \item\label{Prop:rho:Reg}(Variation Regularization) When $\rho/\eta\to\infty$ and suppose additionally Assumption~\ref{Assumption:divergence} holds, then
\[
\Big| 
\mathcal{E}_{\hP}(f;\rho,\eta) -
\mathcal{R}_2(f; \rho,\eta)
\Big|=o(\rho).
\]
    \item\label{Proposition:2:smooth}(Variance Regularization) When $\rho/\eta\to 0$ and suppose additionally that $\phi(t)$ is two times continuously differentiable in a neighborhood of $t=1$ with $\phi''(1)>0$, then 
\[
\Big| 
\mathcal{E}_{\hP}(f;\rho,\eta) -
\mathcal{R}_3(f; \rho,\eta)
\Big|=o(\rho).
\]
\end{enumerate}
\end{theorem}

The proof idea is to consider the surrogate of $\mathcal{E}_{\hP}(f;\rho,\eta)$, by replacing $f$ with its first-order Taylor expansion, which leads to 
\begin{equation}
\widetilde{\mathcal{E}}_{\hP}(f;\rho,\eta) =\rho\cdot \mathbb{E}_{z\sim\hP}\left[ 
\inf_{\mu\in\mathbb{R}}\left\{ 
\mu + \frac{1}{\rho/\eta}\mathbb{E}_{b\sim \beta}\left[ 
\phi^*\Big( 
\frac{\rho}{\eta}\cdot(\nabla f(z)\trans b - \mu)
\Big)
\right]
\right\}
\right].\label{Eq:surrogate}
\end{equation}
Based on Assumption~\ref{Assumption:f:smooth}, it can be shown that $\mathcal{E}_{\hP}(f;\rho,\eta) = \widetilde{\mathcal{E}}_{\hP}(f;\rho,\eta) + O(\rho^2)$.
Thus, it suffices to derive approximations of $\widetilde{\mathcal{E}}$ under different scaling regimes of $\rho/\eta$.
Below, we provide interpretations on this main result for each scaling regime.

\subsubsection*{Case 1: $\rho/\eta\to C\in(0,\infty)$.}
In this case, the perturbation budget $\rho$ and regularization level $\eta$ decay in the same order. It is noteworthy that the drived regularization $\mathcal{R}_1(f; \rho,\eta)$ in \eqref{Eq:R:1} has close connection to the optimized certainty equivalent risk~(OCE) measure studied in \citep{ben1987penalty}: Define the OCE of random variable $X$ with parameter $\eta$ as 
\[
\mathfrak{S}_{\eta}(X) = \eta\cdot \inf_{\mu\in\mathbb{R}}~\Big\{ 
\mu + \bE\Big[\phi^*\Big(\frac{X}{\eta} - \mu\Big)\Big]
\Big\} = \inf_{\mu\in\mathbb{R}}~\Big\{ 
\mu + \bE\Big[\eta\phi^*\Big(
\frac{X-\mu}{\eta}
\Big)\Big]
\Big\},
\]
then $\mathcal{R}_1(f; \rho,\eta)=
\rho\cdot\bE_{z\sim\hP}\Big[\mathfrak{S}_{1/C}
\Big( 
(\nabla f(z))_{\#}\beta
\Big)\Big]$, where $(\nabla f(z))_{\#}\beta$ is the pushforward probability measure of $\beta$ by the linear projection map $\inp{\cdot}{\nabla f(z)}$.
Namely, the regularization term $\mathcal{R}_1(f; \rho,\eta)$ represents the averaged value of OCE across the projection of the loss gradient $\nabla f(z)$ with random projection directions.

Interestingly, it can be shown that the regularization $\mathcal{R}_1(f; \rho,\eta)$ converges to $\mathcal{R}_2(f; \rho,\eta)$ as the constant $C\to\infty$, and converges to $\mathcal{R}_3(f; \rho,\eta)$ as $C\to0$.
When $\rho/\eta\to C$ for some $C>0$, the corresponding regularized ERM is an interpolation between the regularized ERM formulations corresponding to other two extreme cases.

\subsubsection*{Case 2: $\rho/\eta\to\infty$.}
In this case, the convergence rate of the regularization level $\eta$ is faster than that of the perturbation budget $\rho$. 
We showed that \eqref{Eq:regularizer:residual} is asymptotically equivalent to the gradient norm regularization in \eqref{Eq:R:2}.
Recall that \citet{gao2022wasserstein} showed the standard $\infty$-Wasserstein DRO can be approximated using the same regularization term, and the authors therein call it the \emph{variation regularization}.
Therefore, our finding in this case matches our intuition since the regularization level $\eta$ has little impact on the regularization effect.

\subsubsection*{Case 3: $\rho/\eta\to 0$.}
Finally, we consider the case where the convergence rate of $\rho$ is faster than that of the regularization $\eta$. We showed that \eqref{Eq:regularizer:residual} is asymptotically equivalent to the variance regularization \eqref{Eq:R:3} in terms of the projected gradient of the loss $f$.
Note that the regularization effect for this case requires the assumption that the divergence function $\phi(t)$ should be two times continuously differentiable and locally strongly convex around $t=1$, which is a common condition in the study of general $\phi$-divergence DRO~\cite{duchi2019variance, blanchet2023statistical, lam2016robust, duchi2019variance}.

Recall that \cite{duchi2019variance, blanchet2023statistical} showed the $\phi$-divergence DRO with a sufficiently small size of ambiguity set can be well-approximated by the ERM with variance regularization in terms of the loss.
By taking the first-order Taylor expansion regarding the loss, these regularizations relate to each other.
An intuitive explanation is that the impact of regularization level dominates in this case, which corresponds to the regime where the worst-case transport mapping \eqref{Eq:primal:WDRO:phi:reg} is sufficiently close to the reference mapping.
This indeed corresponds to the case studied in the aforementioned reference.

\section{Generalization Error Bound}
\label{Sec:generation}

In this section, we investigate the generalization properties of our proposed adversarial learning framework in \eqref{Eq:ent:adv:transport}.
To simplify our analysis, we focus on the \emph{multi-class} classification setup, i.e., the loss function $f_{\theta}(z)$ is defined as 
$f_{\theta}(z) = \ell\big(g_{\theta}(x), y\big)$, where the data point $z=(x,y)$ represents the feature-label pair, $g_{\theta}(x)$ is the predictor function parameterized by $\theta$, and $\ell:~\mathbb{R}^K\times\{1,\ldots,K\}\to[0,1]$ denotes a $K$-class classification loss function such as the cross-entropy loss.
Throughout this section, we take the norm function that appears in the $\infty$-Wasserstein metric as
\begin{equation*}
\|z - z'\| = \|x-x'\|_{\infty} + \infty\cdot \textbf{1}\{y\ne y'\},
\end{equation*}
where $z=(x,y)$ and $z'=(x',y')$ are two different data points.
 Thus, we take into account only the distribution shift of the feature vector and omit the label distribution shift.
Let $S=\{z_i\}_{i=1}^n$ denote the set of $n$ i.i.d. sample points generated from $\bP_{\text{true}}$, and $\bP_n$ be the empirical distribution supported on $S$.
For fixed parameter $\theta$, let $\widehat{R}_{\text{adv}}(\theta)$ be the objective value of \eqref{Eq:ent:adv:transport} with $\hP=\bP_n$, and ${R}_{\text{adv}}(\theta)$ be its population version, i.e.,
\begin{equation*}
\begin{aligned}
{R}_{\text{adv}}(\theta) & = \sup_{
\substack{\bP,\gamma
}}~\left\{
\bE_{(x,y)\sim\bP}~[\ell\big(g_{\theta}(x), y\big)] - \eta \bD_{\phi}(\gamma,\gamma_0):~
\begin{array}{l}
\text{Proj}_{1\#}\gamma=\bP_{\text{true}}, \text{Proj}_{2\#}\gamma=\bP\\
\text{ess.sup}_{\gamma}\|\zeta_1-\zeta_2\|\le\rho
\end{array}
\right\}\\
&=\bE_{(x,y)\sim \bP_{\text{true}}}\left[ 
\inf_{\mu\in\mathbb{R}}~\bigg\{ 
\mu + \bE_{b\sim\beta}\left[ 
(\eta\phi)^*\left( 
\ell\big(g_{\theta}(x+b), y\big) - \mu
\right)
\right]
\bigg\}
\right],
\end{aligned}
\label{Eq:R:adv:theta}
\end{equation*}
where the last equality is based on the strong duality result in Theorem~\ref{Thm:strong:dual:general}, and $b\sim \beta$ is a random vector uniformly distributed on $\mathbb{B}_{\rho}(0)$, the $\|\cdot\|_{\infty}$-ball of radius $\rho$ centered at the origin.
One of the most important research questions in learning theory is to provide the gap between the \emph{empirical regularized adversarial risk} $\widehat{R}_{\text{adv}}(\theta)$ and the \emph{population regularized adversarial risk} ${R}_{\text{adv}}(\theta)$~(see, e.g., \citep{yin2019rademacher, attias2019improved, awasthi2020adversarial}).
We answer this question leveraging the covering number argument.


Let us begin with some technical preparation. 
Let $\epsilon>0$ and $(\mathcal{V}, \|\cdot\|)$ be a normed space.
We say $\mathcal{C}\subseteq \mathcal{V}$ is an $\epsilon$-cover of $\mathcal{V}$ if for any $V\in\mathcal{V}$, there exists $V'\in\mathcal{C}$ such that $\|V - V'\|\le \epsilon$. The least cardinality of $\mathcal{C}$ is called the $\epsilon$-covering number, denoted as $\mathcal{N}(\mathcal{V}, \epsilon, \|\cdot\|)$.
For any $x,x'\in\mathcal{V}^n$, we take the norm $\|x - x'\| = \max_{i\in[n]}\|x_i - x'_i\|$.
Define the regularized adversarial function class
\[
\mathcal{G}_{\text{adv}} = \left\{ 
(x,y)\mapsto 
\inf_{\mu\in\mathbb{R}}~\bigg\{ 
\mu + \bE_{b\sim\beta}\left[ 
(\eta\phi)^*\left( 
\ell\big(g_{\theta}(x+b), y\big) - \mu
\right)
\right]
\bigg\}:\quad 
\theta\in\Theta
\right\}.
\]
For dataset $S = \{z_i\}_{i=1}^n = \{(x_i,y_i)\}_{i=1}^n$, define 
\[
\mathcal{G}_{\text{adv}\mid_{S}} = \Big\{ 
(g(x_1,y_1),\ldots, g(x_n,y_n)):~g\in \mathcal{G}_{\text{adv}}
\Big\}.
\]
An intermediate consequence of the covering number argument is the following.
\begin{proposition}[{\citep{yin2019rademacher, shalev2014understanding}}]\label{Proposition:cover}
Suppose the range of the loss function $(x,y)\mapsto \ell(g_{\theta}(x),y)$ is $[0,1]$.
With probability at least $1-\delta$ with respect to $\hP=\bP_n$, it holds that for all $\theta\in\Theta$,
\begin{equation}
{R}_{\text{adv}}(\theta)\le \widehat{R}_{\text{adv}}(\theta) + \inf_{\alpha>0}~\left( 
8\alpha + \frac{24}{\sqrt{n}}\int_{\alpha}^{1}\sqrt{\log\mathcal{N}(\mathcal{G}_{\text{adv}\mid_{S}}, \epsilon, |\cdot|)}\diff\epsilon
\right) + 3\sqrt{\frac{\log(2/\delta)}{n}}.\label{Eq:Radv:eRadv}
\end{equation}
\end{proposition}
The remaining challenge is to provide the upper bound on the covering number $\mathcal{N}(\mathcal{G}_{\text{adv}\mid_{S}}, \epsilon, |\cdot|)$.
Define the new function class of interest as
\[
\mathcal{G} = \Big\{ 
(x,y)\mapsto \ell(g_{\theta}(x + \cdot),y)\in\mathbb{R}^{\mathbb{B}_{\rho}(0)}:~\theta\in\Theta
\Big\},\qquad 
\mathcal{G}_{\mid_{S}} = \Big\{ 
(g(x_1,y_1),\ldots, g(x_n,y_n)):~g\in \mathcal{G}
\Big\}.
\]
and the associated norm $\|g\|_{\infty} = \sup_{b\in\mathbb{B}_{\rho}(0)}|g(b)|, \forall g\in\mathcal{G}$.
The following proposition controls the covering number of $\mathcal{G}_{\text{adv}\mid_{S}}$ using that of $\mathcal{G}_{\mid S}$.

\begin{proposition}\label{Lemma:covering:Gadv}
Assume that $\phi$ is strictly convex.
Then, it holds that $\mathcal{N}(\mathcal{G}_{\text{adv}\mid_{S}}, \epsilon, |\cdot|)\le\mathcal{N}(\mathcal{G}_{\mid_{S}}, \epsilon, \|\cdot\|_{\infty}) $.
\end{proposition}
Proposition~\ref{Lemma:covering:Gadv} gives an estimate of $\mathcal{N}(\mathcal{G}_{\text{adv}\mid_{S}}, \epsilon, |\cdot|)$ by taking $\mathcal{G}_{\mid_{S}}$ that involves the perturbation set $\mathbb{B}_{\rho}(0)$ into account.
The advantage is that the covering number $\mathcal{N}(\mathcal{G}_{\mid_{S}}, \epsilon, \|\cdot\|_{\infty})$ can be easily computed. 
The following presents several applications of Propositions~\ref{Proposition:cover} and \ref{Lemma:covering:Gadv}.

\begin{example}[Linear Function Class]
Let us model the predictor $g_{\theta}(x)$ using the linear function class
\begin{equation*}
\label{Eq:linear:class}
g_{\theta}(x) = Wx,\quad \theta\in\Theta:=\Big\{W\in\mathbb{R}^{K\times d},\quad \|W\|_{1,\infty}\le \Lambda_1, \|W\|_{2,2}\le \Lambda\Big\},
\end{equation*}
where the feature vector $x\in\mathbb{R}^d$ is assumed to be bounded: $\sup_x\|x\|_2\le\Psi$.
Next, we consider the ramp loss $\ell:~\mathbb{R}^K\times\{1,\ldots,K\}\to\mathbb{R}_+$
\begin{equation}
\ell(t, y)=\left\{
\begin{aligned}
&1, &\quad\text{if }M(t,y)\le0,\\
&1-\frac{1}{\varrho}M(t,y),&\quad\text{if }0<M(t,y)<\varrho,\\
&0,&\quad\text{if }M(t,y)\ge\varrho,
\end{aligned}
\right.\label{Eq:rmap}
\end{equation}
where $M(t,y) = t_y - \max_{y'\ne y}t_{y'}$.
By \citep[Lemmas~4.4 and 5.2]{mustafa2022generalization}, it holds that $\log\mathcal{N}(\mathcal{G}_{\mid_{S}}, \epsilon, \|\cdot\|_{\infty}) = \widetilde{O}(
\frac{\Lambda^2d(\Psi + \sqrt{d}\rho)^2}{\epsilon^2\varrho^2}
)$, where $\widetilde{O}(\cdot)$ hides constant logarithmically dependent on related parameters.
As a consequence, the generalization bound~\eqref{Eq:Radv:eRadv} further simplifies to
\begin{equation*}
{R}_{\text{adv}}(\theta)\le \widehat{R}_{\text{adv}}(\theta)  + 3\sqrt{\frac{\log(2/\delta)}{n}} + \frac{8}{n} + \widetilde{O}\left( 
\frac{\Lambda\sqrt{d}(\Psi + \sqrt{d}\rho)}{\varrho\sqrt{n}}
\right).\label{Eq:adv:Lambda:linear}
\end{equation*}
\end{example}

\begin{example}[Neural Network Function Class]
We next consider the predictor $g_{\theta}(x)$ belongs to the $L$-layer and $m$-width neural network function class with $1$-Lipschitz nonlinear activation $\sigma$:
\[
f(x) = W_L\cdot \sigma\Big( 
W_{L-1}\cdots \sigma(W_1x)
\Big),
\]
where the feature vector $x\in\mathbb{R}^d$ satisfies $\sup_x\|x\|_2\le\Psi$, and the model parameter
\[
\begin{aligned}
\theta\in\Theta:=\Big\{ 
&(W_1,\ldots,W_L):~\|W^l\|_2\le a_l, \|W^l\|_{\text{sp}}\le s_l, l=1,\ldots,L-1\\
&\qquad\qquad\qquad\qquad\qquad\qquad\|W^L\|_2\le a_L, \|W^L\|_{2,\infty}\le s_L, \|W^1\|_{1,\infty}\le s_1'
\Big\}.
\end{aligned}
\]
When considering the ramp loss in \eqref{Eq:rmap}, according to \citep[Lemma~5.14]{mustafa2022generalization},
\[
\log\mathcal{N}(\mathcal{G}_{\mid_{S}}, \epsilon, \|\cdot\|_{\infty}) = \widetilde{O}\left(
\frac{L^2{d}(\Psi + \sqrt{d}\rho)^2}{\varrho^2\epsilon}\cdot 
\prod_{l\in[L]}s_l^2\cdot\sum_{l\in[L]}\frac{a_l^2}{s_l^2}
\right).
\]
Then, the generalization bound~\eqref{Eq:Radv:eRadv} becomes
\[
{R}_{\text{adv}}(\theta)\le \widehat{R}_{\text{adv}}(\theta)  + 3\sqrt{\frac{\log(2/\delta)}{n}} + \frac{8}{n} + \widetilde{O}\left( 
\frac{L\sqrt{d}(\Psi + \sqrt{d}\rho)}{\varrho\sqrt{n}}\cdot 
\prod_{l\in[L]}s_l\cdot \sqrt{\sum_{l\in[L]}\frac{a_l^2}{s_l^2}}
\right).
\]
\end{example}

\begin{remark}
Compared to the generalization analysis of the unregularized adversarial robust learning formulation in \eqref{Eq:formula:adv} (see, e.g., \citep{khim2018adversarial, yin2019rademacher, awasthi2020adversarial, xiao2022adversarial}), our error bound for the regularized case is similar. It aligns with the state-of-the-art error bound for the unregularized case. The novelty in our theoretical analysis lies in upper bounding the covering number of $\mathcal{G}_{\text{adv}\mid S}$ in Proposition~\ref{Proposition:cover} by that of $\mathcal{G}_{\mid S}$.
Based on the lower bound of Rademacher complexity for neural network function classes~\citep[Theorem~3.4]{bartlett2017spectrally}, our generalization bound for the linear function class matches the lower bound in terms of the parameters $\Psi$, $\Lambda$, and $n$, but introduces an additional $O(d)$ term. For neural network function classes, our bound introduces an additional $O(Ld\cdot \sqrt{\sum_{l\in[L]}\frac{a_l^2}{s_l^2}})$ term. While these additional terms are relatively small, developing new proof techniques to further tighten the generalization analysis would be desirable.
\end{remark}


\section{Numerical Study}\label{Sec:numerical}
In this section, we examine the numerical performance of our proposed algorithm on three applications: supervised learning, reinforcement learning, and contextual learning.
We compare our method with the following baselines: (i) empirical risk minimization~(ERM), (ii) fast-gradient method~(FGM)~\citep{goodfellow2014explaining}, (iii) and its iterated variant~(IFGM)~\citep{kurakin2016adversarial}.
Those baseline methods are heuristic approaches to approximately solving the $\infty$-WDRO model.


\subsection{Supervised Learning}\label{Sec:supervised}
We validate our method on three real-world datasets: MNIST~\citep{lecun-mnisthandwrittendigit-2010}, Fashion-MNIST~\citep{xiao2017fashion}, and Kuzushiji-MNIST~\citep{clanuwat2018deep}.
The experiment setup largely follows from \citet{sinha2017certifying}.
Specifically, we build the classifer using a neural network with $8\times8$, $6\times6$, and $5\times 5$ convolutional filter layers and ELU activations, and followed by a connected layer and softmax output.
After the training process with those listed methods, we then add various perturbations to the testing datasets, such as the $\ell_2$-norm and $\ell_\infty$-norm adversarial projected gradient method~(PGM) attacks~\citep{madry2017towards}, and white noises uniformly distributed in a $\ell_2$ or $\ell_\infty$ norm ball.
We use the mis-classification rate on testing dataset to quantify the performance for the obtained classifers.
For fair comparison, we take the same level of robustness parameter $\rho=\texttt{0.45}$ for all approaches, and specify the number of epochs (i.e., the number of times the data points are passed through the model) as $\texttt{30}$ and use Adam optimizer with stepsize $\gamma=\texttt{1e-3}$.
For PGM attack and FGM/IFGM defense, we specify the stepsize for the attack step as $\texttt{0.1}$.
The number of iterations for the attack step of PGM attack and IFGM defense is set to be $\texttt{15}$.
Since the scaling of $\eta$ that satisfies $\rho/\eta\to C$ for some constant $C$ corresponds to an interpolation of gradient norm and gradient variance regularized ERM training as suggested by Section~\ref{Sec:reg}, we specify the regularization value $\eta=2\cdot\rho$ in this experiment.
Since the loss is highly nonconvex, we examine our regularized adversarial training using entropic regularization only, with RT-MLMC gradient estimator and the maximum level $L=\texttt{7}$.

\begin{figure}[!ht]
    \centering
    \includegraphics[height=0.26\textwidth]{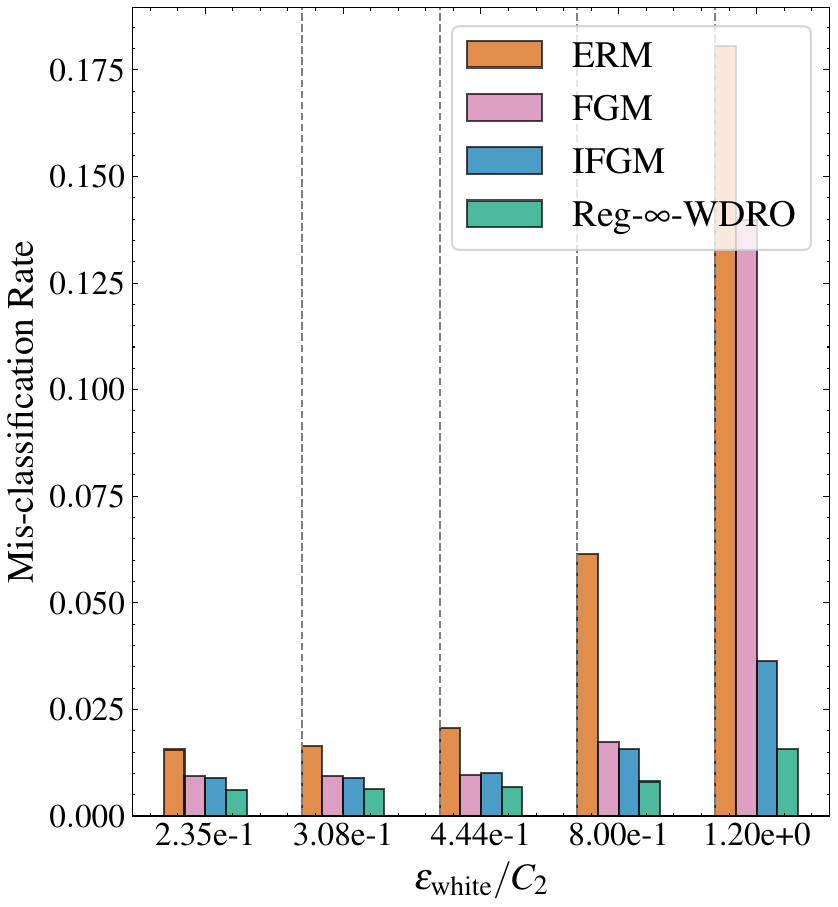}
     \includegraphics[height=0.26\textwidth]{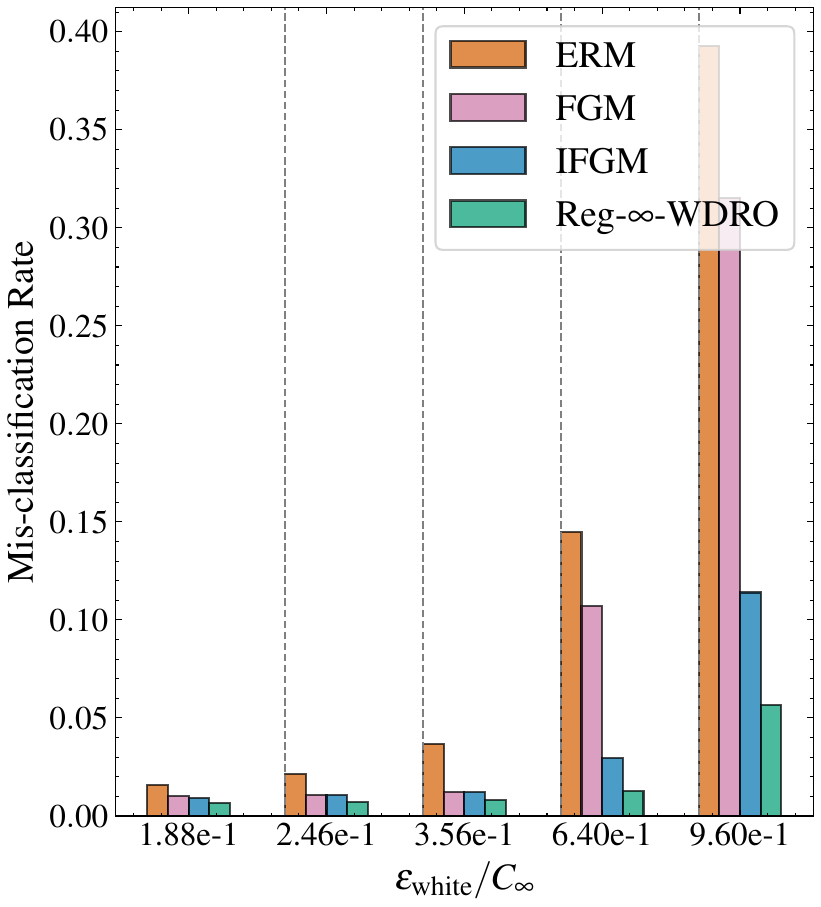}
     \includegraphics[height=0.26\textwidth]{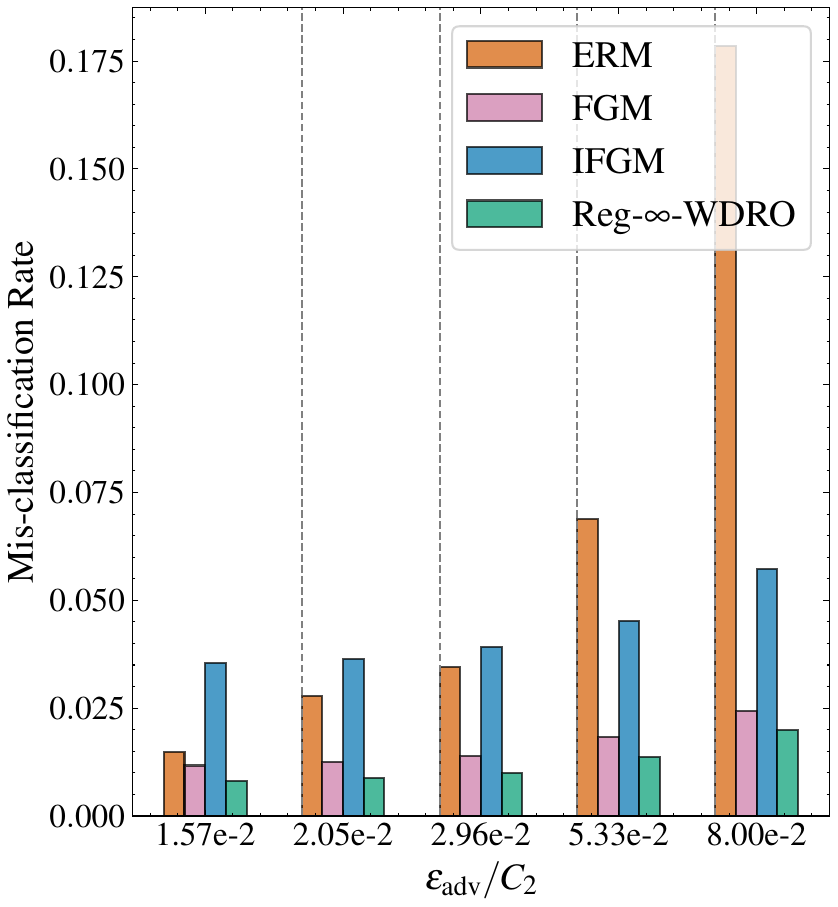}
      \includegraphics[height=0.26\textwidth]{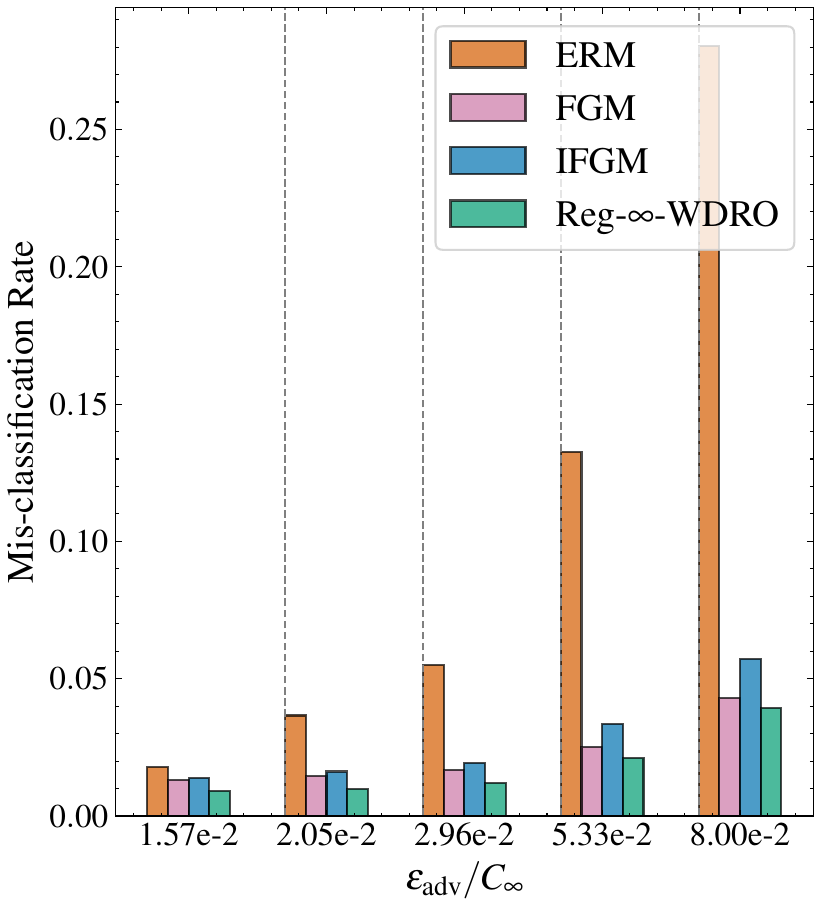}

      \includegraphics[height=0.26\textwidth]{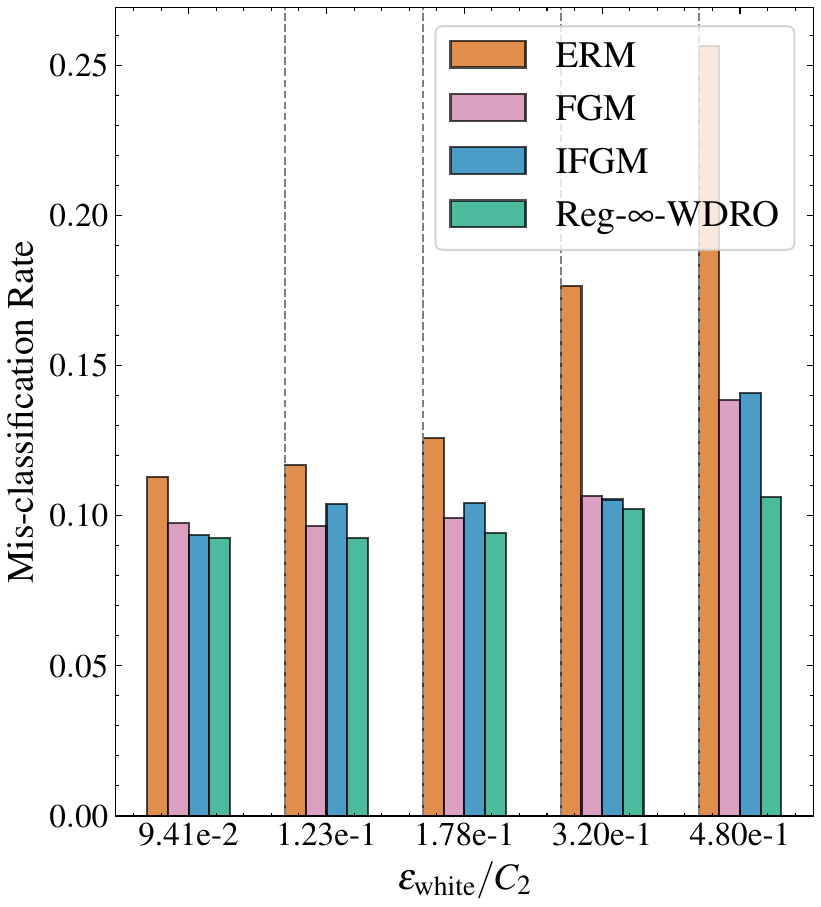}
     \includegraphics[height=0.26\textwidth]{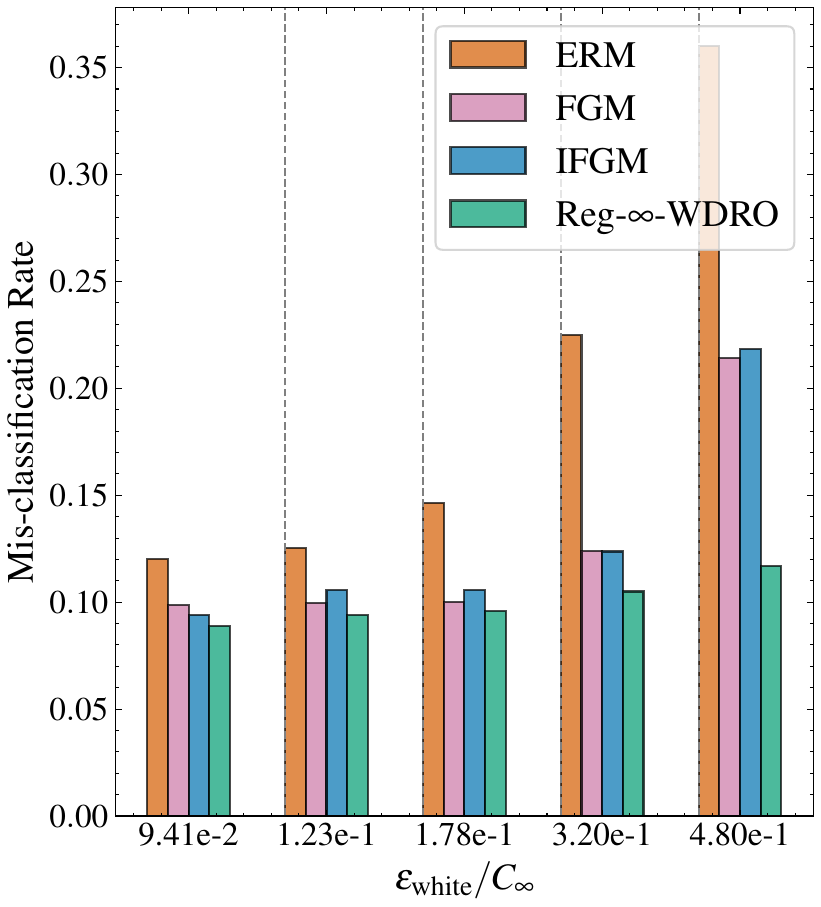}
     \includegraphics[height=0.26\textwidth]{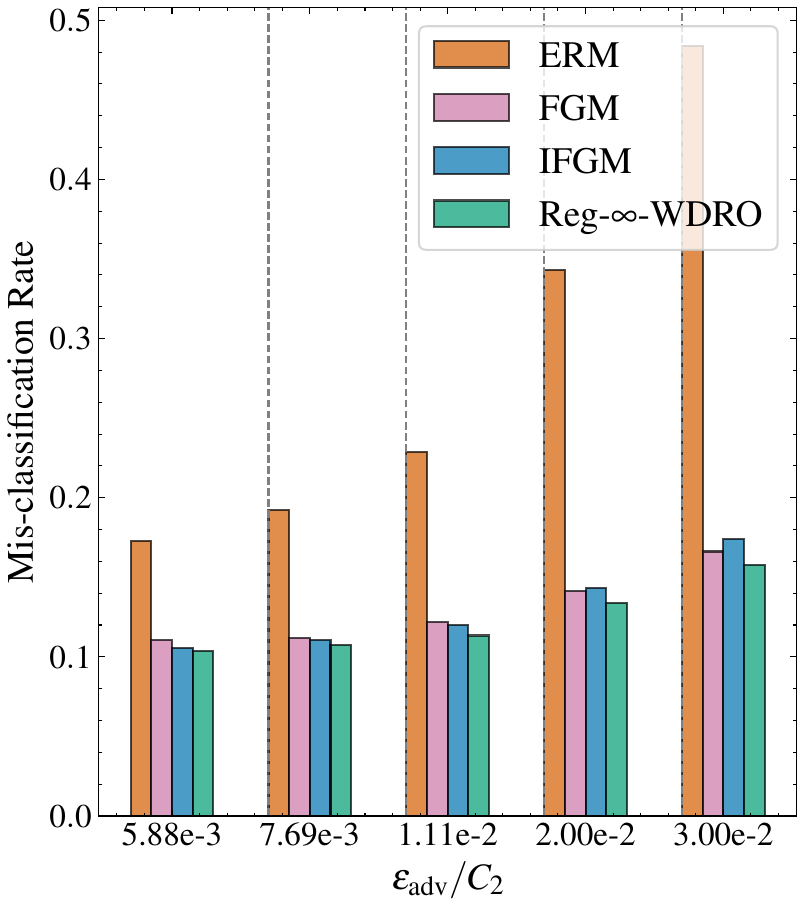}
      \includegraphics[height=0.26\textwidth]{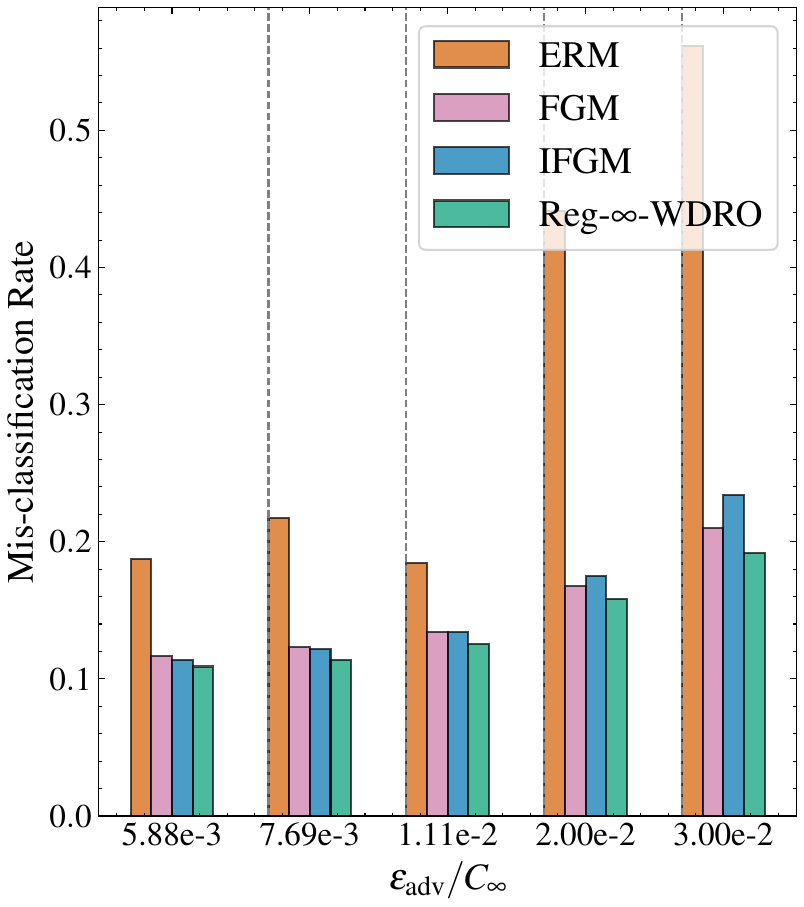}

   \includegraphics[height=0.26\textwidth]{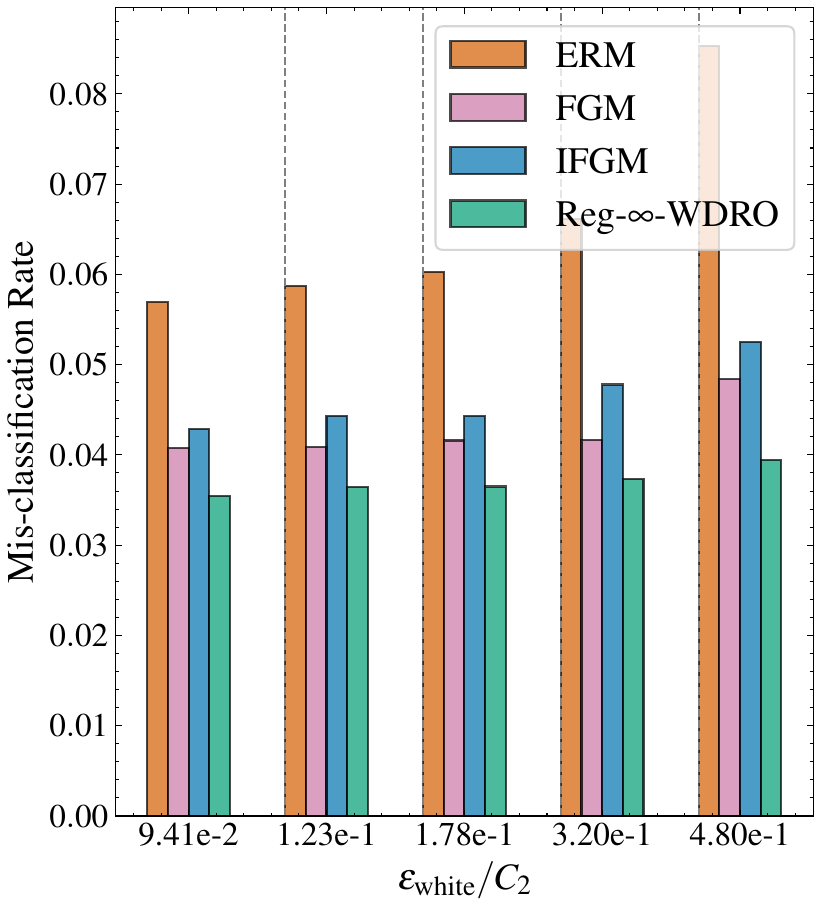}
     \includegraphics[height=0.26\textwidth]{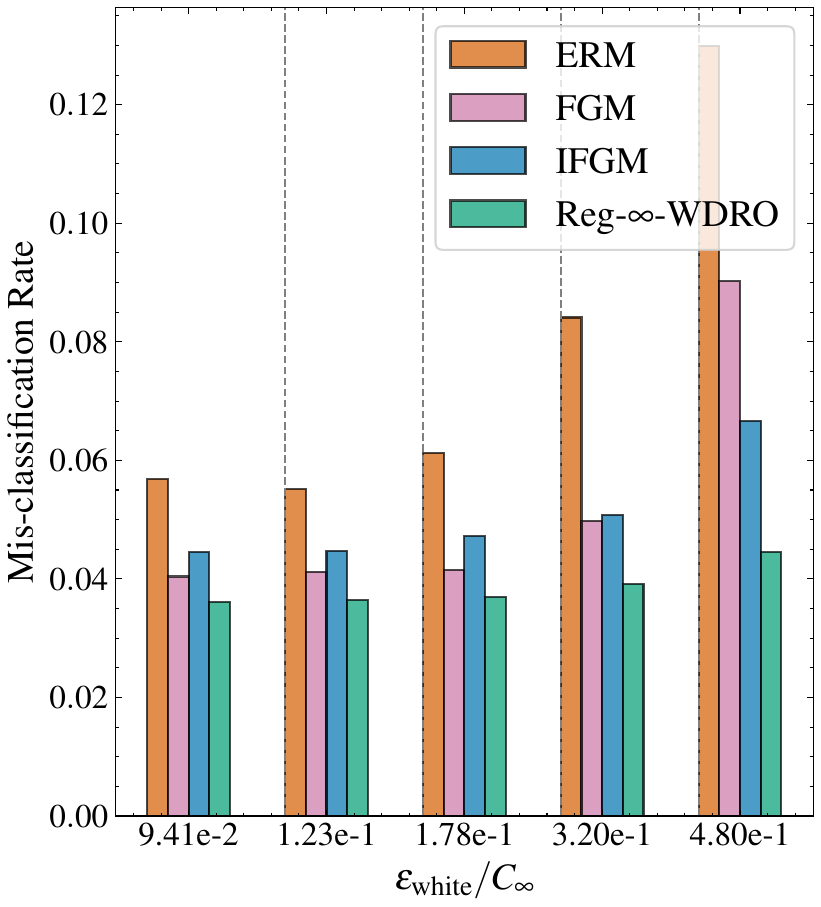}
     \includegraphics[height=0.26\textwidth]{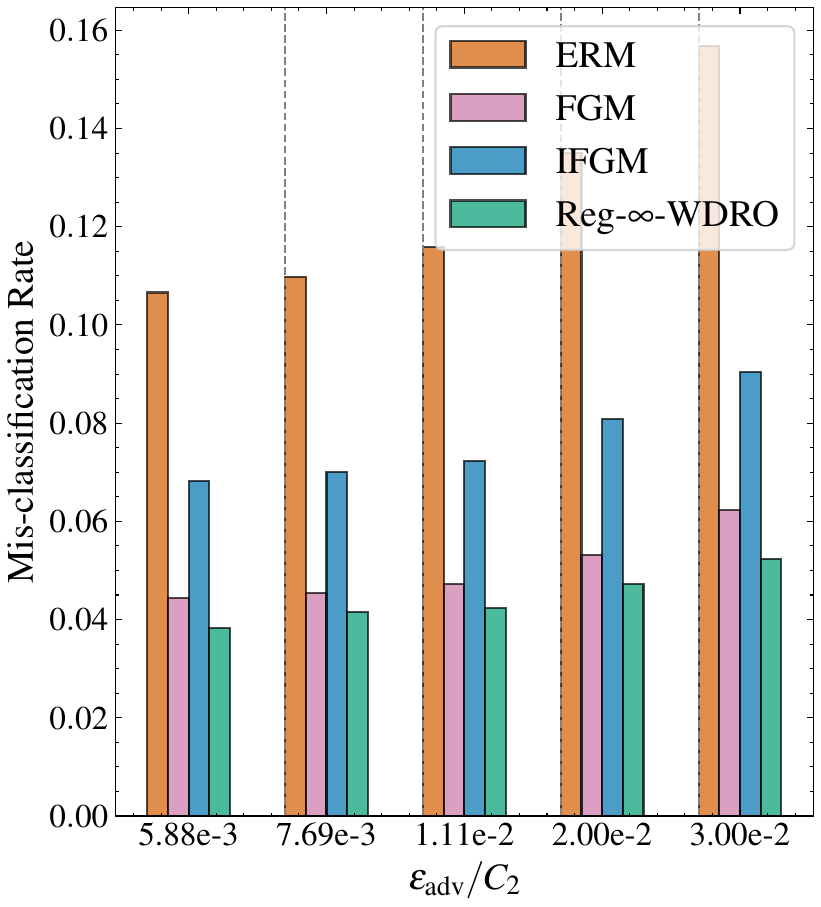}
      \includegraphics[height=0.26\textwidth]{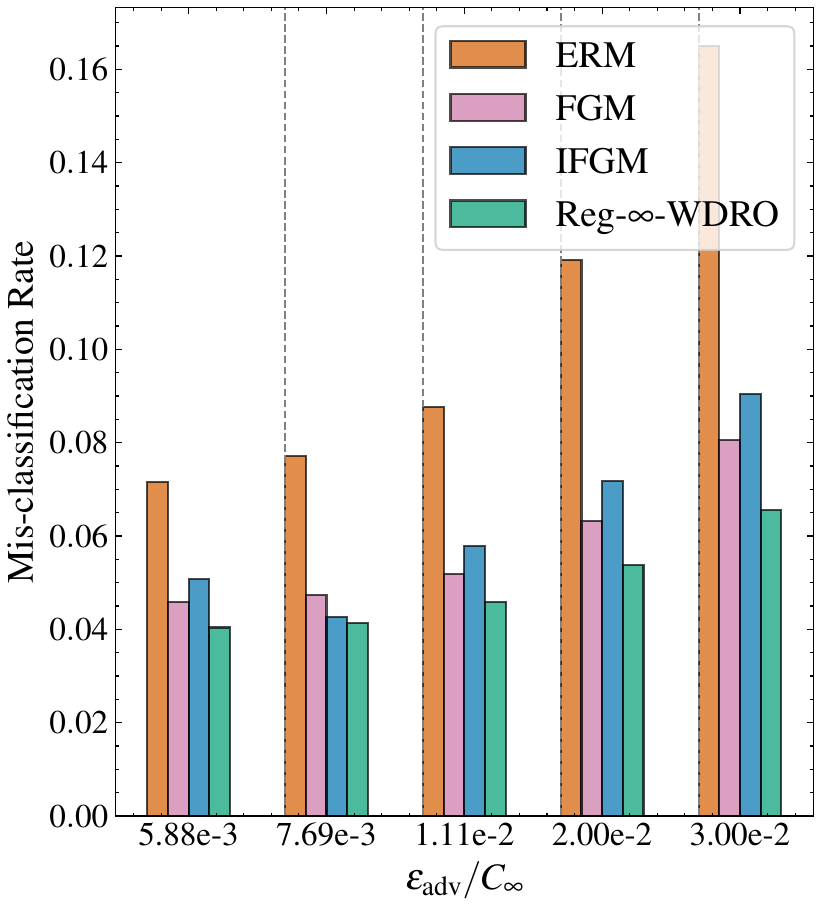}
    \caption{Results of adversarial training in terms of mis-classification rates. 
    From top to bottom, the figures correspond to (a) MNIST; (b) Fashion-MNIST; (c) and Kuzushiji-MNIST datasets.
    From left to right, the figures correspond to (a) $\ell_2$-norm white noise attack; (b) $\ell_\infty$-norm white noise attack; (c)  $\ell_2$-norm PGM attack; and (d) $\ell_{\infty}$-norm PGM attack.}
    \label{fig:performance}
\end{figure}

Figure~\ref{fig:performance} presents the mis-classification results of various methods.
Specifically, three rows correspond to different kinds of datasets~(MNIST, Fashion-MNIST, and Kuzushiji-MNIST), and four columns correspond to different types of perturbations~($\ell_2$/$\ell_{\infty}$ adversarial attack, and $\ell_2/\ell_{\infty}$ white noise attack).
For every single plot, the $x$-axis corresponds to the magnitude of perturbation ($\epsilon_{\text{white}}$ or $\epsilon_{\text{adv}}$) normalized by the average of the norm overall feature vectors (in terms of $2$- or $\infty$-norm, denoted as $C_{2}$ or $C_{\infty}$, respectively).
From these plots, we find that all methods tend to have worse performance as the perturbation level increases, but the regularized adversarial risk model consistently outperforms all baselines.
Especially, it performs well when the perturbation levels are large.
This suggests that our model has superior performance for adversarial training in scenarios with large perturbations.

\subsection{Reinforcement Learning}
Next, we provide a robust algorithm for reinforcement learning~(RL).  
Consider an infinite-horizon discounted finite state MDP represented by a tuple $\langle \mathcal{S}, \mathcal{A}, \mathbb{P}, R, \gamma\rangle$, 
where $\mathcal{S}, \mathcal{A}$ denotes the state and action space, respectively; 
$\mathbb{P}=\{\mathbb{P}(s'\mid s,a)\}_{s,s',a}$ is the set of transition probability metrics; 
$R=\{r(s,a)\}_{s,a}$ is the reward table with $(s,a)$-th entry being the reward for taking the action $a$ at state $s$;
and $\gamma\in(0,1)$ is the discounted factor.
Similar to problem~\eqref{Eq:formulation:wass}, robust reinforcement learning seeks to maximize the worst-case risk function $\sup_{\mathbb{P}\in\mathfrak{R}}\mathbb{E}[\sum_t\gamma^tr(s_t,a_t)]$, with $\mathfrak{R}$ represents the ambiguity set for state-action transitions.
For simplicity, we consider a tabular $Q$-learning setup in this subsection.
The standard $Q$-learning algorithm in RL learns a $Q$-function $Q:~\mathcal{S}\times\mathcal{A}\to\mathbb{R}$ with iterations
\begin{equation}\label{Eq:Q:learning}
Q(s^t,a^t)\leftarrow (1-\alpha_t)Q(s^t,a^t) + \alpha_tr(s^t,a^t) - \gamma\alpha_t\min_a(-Q(s^{t+1},a)),\quad s^{t+1}\sim\mathbb{P}(\cdot\mid s^t,a^t). 
\end{equation}
We modify the last term of the update~\eqref{Eq:Q:learning} with an adversarial state perturbation to take $\infty$-Wasserstein distributional robustness with entropic regularization into account, leading to the new update 
\[
Q(s^t,a^t)\leftarrow (1-\alpha_t)Q(s^t,a^t) + \alpha_tr(s^t,a^t) - \gamma\alpha_t
\min_a\left\{ 
\eta\log\mathbb{E}_{\widehat{s}^{t+1}\sim\beta(s^{t+1};\rho)}e^{-Q(\widehat{s}^{t+1}, a)/\eta}
\right\},
\]

\begin{figure}[!t]
\begin{minipage}{1\textwidth}
\begin{minipage}{0.4\textwidth}
\centering
{\small
\renewcommand{\arraystretch}{1.75}
\begin{tabular}{c|cc}
  Environment & Regular & Robust \\ \hline \hline
  Original MDP & 469.42 $\pm$ 19.03 & \textbf{487.11 $\pm$ 9.09} \\
  \hline 
  Perturbed MDP~(Heavy) & 187.63 $\pm$ 29.40 & \textbf{394.12 $\pm$ 12.01} \\
  Perturbed MDP~(Short) & 355.54 $\pm$ 28.89 &  \textbf{443.17 $\pm$ 9.98} \\
  Perturbed MDP~(Strong $g$) & 271.41 $\pm$ 20.7 &  \textbf{418.42$\pm$ 13.64}\\
  \hline
  \hline
  \end{tabular}
  \captionof{table}{Performance of $Q$-learning algorithms in original MDP and shifted MDP environments.
  Error bars are produced using $10$ independent trials.}
 }
\end{minipage}\qquad\qquad\qquad\quad
\begin{minipage}{0.4\columnwidth}%
\label{Fig:train:MDP}
\centering
\includegraphics[width=0.6\textwidth]{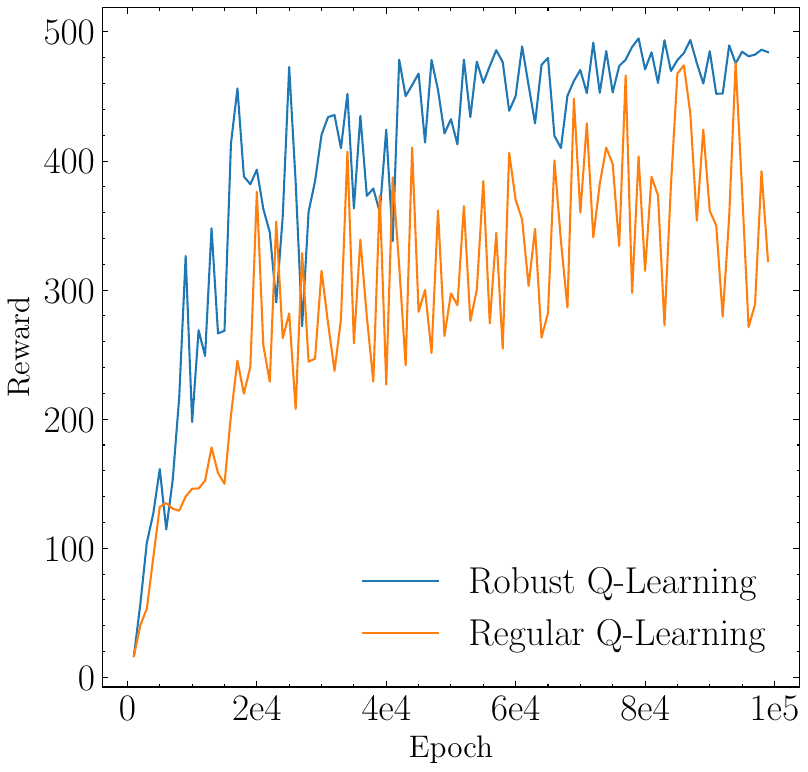}
\captionof{figure}{Episode lengths during training. The environment caps episodes to $400$ steps.}
\end{minipage}
\end{minipage}
\end{figure}

\noindent
where $\beta(s^{t+1};\rho)$ denotes an uniform distribution supported on a $\|\cdot\|_{\infty}$-norm ball of $s^{t+1}$ with radius $\rho$.
Standard fixed point iteration analysis 
\citep{szepesvari1999unified, wang2024reliable, yang2017convex} can be modified to show the convergence of the modified $Q$-learning iteration.
Our proposed algorithm in Section~\ref{Sec:opt} can be naturally applied to proceed the updated $Q$-learning iteration.

We test our algorithm in the cart-pole environment~\citep{brockman2016openai}, where the objective is to balance a pole on a cart by moving the cart to left or right, with state space including the physical parameters such as chart position, chart velocity, angle of pole rotation, and angular of pole velocity.
To generate perturbed MDP environments, we perturb the physical parameters of the system by magnifying the pole's mass by $2$, or shrinking the pole length by $2$, or magnifying the strength of gravity $g$ by $5$.
We name those three perturbed environments as \emph{Heavy}, \emph{Short}, or \emph{Strong $g$} MDP environments, respectively.

Figure~3 demonstrates the training process of regular and robust $Q$-learning algorithms on the original MDP environment. 
Interestingly, the robust $Q$-learning algorithm learns the optimal policy more efficiently than the regular MDP. 
One possible explanation is that taking account into adversarial perturbations increase the exploration ability of the learning algorithm.
Next, we report the performance of trained policies in original and perturbed MDP environments in Table~2, from which we can see that our proposed robust $Q$-learning algorithm consistently outperforms the regular non-robust algorithm.

\subsection{Contextual Learning}
Contextual stochastic optimization~(CSO) seeks the optimal decision to minimize the cost function $\Psi$ involving random parameters $Z$, whose distribution is affected by a vector of relevant covariates denoted as $X$.
Since one has access to covariates $X$ before decision making, we parameterize the optimal decision using $f_{\theta}(\cdot)$ that maps from $X$ to the final decision.
This paradigm, inspired by the seminar work \citep{bertsimas2020predictive}, has achieved phenomenal success in operations research applications.
See the survey \citep{sadana2023survey} that summarizes its recent developments.

Distributionally robust CSO with $\infty$-type casual optimal transport distance has gained great popularity in recent literature~\citep{yang2022decision, arora2022data}.
It seeks the optimal decision parameter $\theta$ to minimize the worst-case risk, where the worst-case means we simultaneously find the casual optimal transport $\gamma$ that maps $\hP$, the empirical distribution from available data $\{(x_i,z_i)\}_i$, to $\bP$ up to certain transportation budget,
Its strong dual reformulation can be reformulated as a special case of \eqref{Eq:formulation:wass}:
\begin{equation}\label{Eq:casual:dual}
\min_{\theta}~\left\{\bE_{\widehat{x}\sim \hP_{\widehat{X}}}\left[ 
\sup_{x'\in\mathbb{B}_{\rho}(\widehat{x})}~\bE_{\widehat{z}\sim\hP_{\widehat{Z}\mid\widehat{X}=\widehat{x}}}[\Psi(f_{\theta}(x'), \widehat{z})]
\right]\right\}.
\end{equation}
Similar to adversarial robust learning, Problem~\eqref{Eq:casual:dual} can be challenging to solve because computing the optimal value of the inner maximization problem is usually NP-hard.
Instead, we replace the inner maximization with OCE risk, leading to the approximation problem
\begin{equation}\label{reg:casual:dual}
\min_{\theta}~\left\{\bE_{\widehat{x}\sim \hP_{\widehat{X}}}\left[ 
\inf_{\mu\in\mathbb{R}}~\Big\{ 
\mu + \mathbb{E}_{x'\sim \nu_{\widehat{x}}}\Big[(\eta\phi)^*(
\bE_{\widehat{z}\sim\hP_{\widehat{Z}\mid\widehat{X}=\widehat{x}}}[\Psi(f_{\theta}(x'), \widehat{z})] - \mu
)\Big]
\Big\}
\right]\right\},
\end{equation}
where $\nu_{\widehat{x}}$ denotes the uniform distribution supported on $\mathbb{B}_{\rho}(\widehat{x})$.
Alternatively, we can express \eqref{reg:casual:dual} as a special case of \eqref{Eq:ent:adv:transport}.
See the detailed discussion in Appendix~\ref{Proof:casual:OT}.

In the following, we test our algorithm in the application of data-driven personalized pricing problem, using the similar setup in \citep[Example~6]{yang2022decision}: 
Let $w\in\mathbb{R}$ denote price, $x\in\mathbb{R}^{10}$ denote side information, and $z\in\mathbb{R}^2$ denote price sensitivity coefficient that describes how price influences customer demand.
The loss $\Psi(w,z)=-wz\trans \begin{pmatrix}
w\\1
\end{pmatrix}$, denoting the negative revenue under price $w$ and coefficient $z$.
Assume $z$ depends on $x$ in a nonlinear way:
\[
z = \begin{pmatrix}
\tanh(3\beta_1\trans x)\\
\exp(-2\beta_2\trans x)
\end{pmatrix} + \mathcal{N}(0, \textbf{I}_2),
\]
where $\beta_1,\beta_2\sim \mathcal{U}([-0.1,0.1]^{10})$ and $x\sim\mathcal{N}(0, \textbf{I}_{10})$.
We solve this problem using the linear decision rule approach, by taking $f_{\theta}(x) = \theta\trans g(x)$, where $g:~\mathbb{R}^{10}\to\mathbb{R}^{100}$ is a random feature model:
\[
g(x) = \left( 
\cos(\omega_i\trans x + b_i)
\right)_{i\in[100]},\quad \omega_i\sim \mathcal{N}(0, \textbf{I}_{10}), b\sim \mathcal{U}([0, 2\pi]).
\]
Throughout the experiment, we take hyper-parameters $\rho=0.45$ and $\eta=0.9$.
When creating training dataset, we generate $M\in\{25, 50, 100, 200\}$ samples of $x$, denoted as $\{x_i\}_{i\in[M]}$, and for each $x_i$, we generate $m\in\{10,30,50,100,200\}$ samples of $z$ from the conditional distribution of $z$ given $x=x_i$.

\begin{figure}[!ht]
    \centering
    \includegraphics[width=1.1\linewidth]{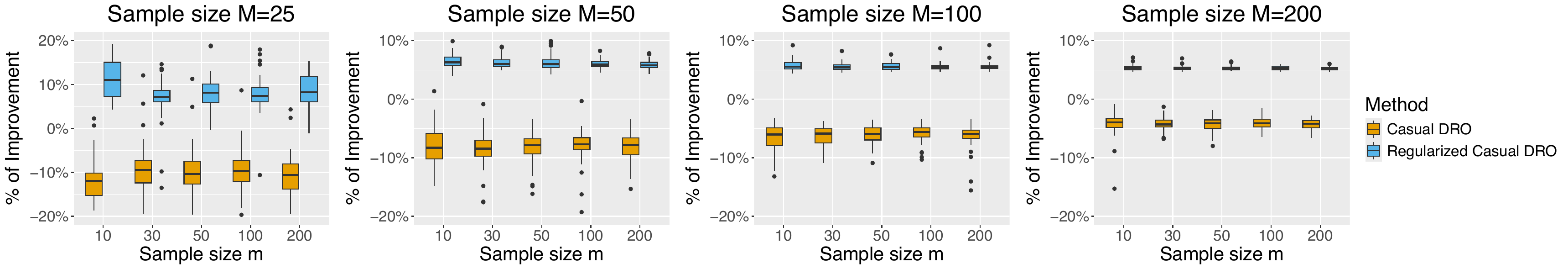}
    \caption{Results of $\infty$-type Casual DRO and its regularized version in terms of percentage of improvements. From left to right, the figures correspond to $M=25,50,100,200$, respectively.}
    \label{fig:OM}
\end{figure}
We quantify the performance of a given decision $\theta$ using the percentage of improvements (compared to ERM) measure:
\[
\mathcal{J}(\theta) = 1 - \frac{\mathcal{R}(\theta) - \mathcal{R}^*}{\mathcal{R}(\theta_{\text{ERM}}) - \mathcal{R}^*},
\]
where $\mathcal{R}^*$ denotes the ground truth optimal revenue provided the distribution of $(x,z)$ is exactly known, $\theta_{\text{ERM}}$ denotes the decision obtained from the ERM, the non-robust training approach, and $\mathcal{R}(\theta)$ denotes the expected revenue of the decision $\theta$ under the ground truth distribution.
The plots in Figure~\ref{fig:OM} report the percentage of improvements obtained either by solving the standard casual CSO problem~\eqref{Eq:casual:dual} using the heuristic FGM method or its (KL-divergence-)regularized formulation~\eqref{reg:casual:dual}.
The error bars are reproduced using 50 independent trials.
For all scenarios, we can see the regularized robust CSO model outperforms the un-regularized one.
Besides, the un-regularized DRO model has negative improvements in general, mainly due to the computational intractability of Problem~\eqref{Eq:casual:dual}.




\section{Conclusion}
In this paper, we proposed a $\phi$-divergence regularized framework for adversarial robust training.
From the computational perspective, this new formulation is easier to solve compared with the original one.
From the statistical perspective, this framework is asymptotically equivalent to certain regularized ERM under different scaling regimes of the regularization and robustness hyper-parameters.
From the generalization perspective, we derived the population regularized adversarial risk is upper bounded by the empirical one up to small residual error.
Numerical experiments indicate that our proposed framework achieves state-of-the-art performance, in the applications of supervised learning, reinforcement learning, and contextual learning.



\theendnotes
%

\bibliographystyle{informs2014} 
{
\bibliography{shortbib}
}

\clearpage

\ECSwitch

\ECHead{Supplementary for \emph{``Regularization for Adversarial Robust Learning''}}

\section{Preliminaries on Projected Stochastic (Sub-)Gradient Descent}
In the following, we present the convergence results on 
 the standard projected stochastic (sub-)gradient descent algorithm with unbiased gradient estimates, which can be useful for the complexity analysis in Section~\ref{Sec:complexity}.

Consider minimization of the objective function $F(\theta)$ over the constrained domain set $\Theta$.
\subsubsection*{Nonsmooth Convex Optimization.}
Let the objective $F(\theta)$ be a convex function in $\theta$.
Assume one can obtain stochastic estimate $G(\theta, \xi)$ such that for any $\theta\in\Theta$, 
\begin{itemize}
    \item $\mathbb{E}[G(\theta, \xi)] \in \partial F(\theta)$, where $\partial F(\theta)$ denotes the subgradient of $F$ at $\theta$;
    \item $\mathbb{E}\left\|G(\theta, \xi)\right\|^2\le M^2$.
\end{itemize}
Starting from an initial guess $\theta_1\in\Theta$, the projected stochastic subgradient descent algorithm generates iterates
\begin{equation}
\theta_{t+1}=\text{Proj}_{\Theta}(\theta_t - \gamma G(\theta_t, \xi_t)),\quad t=1,\ldots,T-1.
\tag{Projected-SGD}
\label{Eq:proj:SGD}
\end{equation}
where $\gamma>0$ is a constant step size, and $\xi_1,\ldots,\xi_{T-1}$ are i.i.d. copies of $\xi$.
We take the average of all iterates $\{\theta_1,\ldots,\theta_T\}$ as the estimated optimal solution, denoted as $\widetilde{\theta}$.
\begin{lemma}[{\citep{nemirovski2009robust}}]\label{Lemma:standard:SGD}
Under the above setting, suppose we take the step size $\gamma=\frac{D_*}{M\sqrt{T}}$, then it holds that
\[
\bE\left[F(\widetilde{\theta}) - \min_{\theta\in\Theta}~F(\theta)\right]\le \frac{D_*M}{\sqrt{T}},
\]
where the constant $D_*=F(\theta_1) - \min\limits_{\theta\in\Theta}~F(\theta)$.
\end{lemma}

\subsubsection*{Smooth Non-Convex Optimization.}
In this part, we do not assume the convexity of $F(\theta)$.
Instead, we assume the objecitve $F(\theta)$ is continuously differentiable and $S$-smooth such that
\[
\|F(\theta) - F(\theta')\|\le S\|\theta - \theta'\|,\qquad \forall \theta,\theta'\in\Theta.
\]
Besides, assume one can obtain stochastic estimate $G(\theta, \xi)$ such that for any $\theta\in\Theta$,
\begin{itemize}
    \item $\mathbb{E}[G(\theta, \xi)] = \nabla F(\theta)$;
    \item $\mathbb{E}\left\|G(\theta, \xi) - \nabla F(\theta)\right\|^2\le \sigma^2$.
\end{itemize}
We generate iteration points using nearly the same procedure as in \eqref{Eq:proj:SGD}, except that we update iteration points using mini-batch gradient estimator:
\begin{equation}
\theta_{t+1}=\text{Proj}_{\Theta}(\theta_t - \gamma V(\theta_t, \xi_t^{1:m})),\quad t=1,\ldots,T-1,\quad 
V(\theta_t, \xi_t^{1:m})=\frac{1}{m}\sum_{i=1}^mG(\theta_t, \xi_t^i),
\tag{Mini-Projected-SGD}
\label{Eq:min:proj:SGD}
\end{equation}
where $\xi_i^t, t=1,\ldots,T-1, i=1,\ldots,m$ are i.i.d. copies of $\xi$.
We take the estimated optimal solution $\widetilde{\theta}$ as the one that is randomly selected from $\{\theta_1,\ldots,\theta_T\}$ with equal probability.\begin{lemma}[Corollary~3 in {\citep{ghadimi2016mini}}]\label{Lemma:SMD:smooth:noncvx}
Under the above setting, suppose we take the step size $\gamma=\frac{1}{2S}$, then it holds that
 \[
\bE\left\|
\frac{1}{\gamma}
\left[
\widetilde{\theta} - \mathrm{Proj}_{\Theta}\big(\widetilde{\theta} - \gamma \nabla F(\widetilde{\theta})\big)\right]
\right\|^2\le \frac{8SD_*}{T} + \frac{6\sigma^2}{m},
\]
where the constant $D_*:=F(\theta_1) - \min_{\theta\in\Theta}~F(\theta).
$
\end{lemma}



\clearpage
\section{Proofs of Technical Results in Section~\ref{Sec:phi:divergence}}\label{Appendix:Sec:phi:divergence}

\proof{Proof of Theorem~\ref{Thm:strong:dual:general}.}
Based on the assumption, it holds that $\diff\gamma(z,z') = \diff\hP(z)\diff\gamma_z(z')$ for some conditional optimal transport mapping $\gamma_z$.
Then, Problem~\eqref{Eq:primal:WDRO:phi:reg} can be reformulated as
\begin{equation}
\sup_{
\{\gamma_z\}_{z\in\mathrm{supp}\,\hP}}
~\left\{
\bE_{z\sim\hP}\bE_{z'\sim\gamma_z}[f(z)] - \eta 
\bE_{z\sim\hP}\bE_{z'\sim\nu_z}\left[ 
\phi\left( 
\frac{\diff\gamma_z(z')}{\diff\nu_z(z')}
\right)
\right]
\right\}.
\end{equation}
Since the optimization over $z\in\mathrm{supp}\,\hP$ is decomposable, it holds that \eqref{Eq:primal:WDRO:phi:reg} equals
\begin{equation}
\bE_{z\sim\hP}\Bigg[ 
\sup_{\gamma_z\in\cP(\cZ)}~\left\{
\bE_{z'\sim\gamma_z}[f(z')] 
-
\eta\bE_{z'\sim\nu_z}
\left[ 
\phi\left( 
\frac{\diff\gamma_z(z')}{\diff\nu_z(z')}
\right)
\right]
\right\}
\Bigg].
\label{Eq:reformula:primal:WDRO:phi:reg}
\end{equation}
The inner supremum problem above is a phi-divergence regularized linear program.
Based on the strong duality result~(see, e.g., Lemma~\ref{Lemma:phi:div}) that reformulates this subproblem, we arrive at the reformulation of Problem~\eqref{Eq:primal:WDRO:phi:reg}.

Next, we show how to construct the worst-case distribution, which suffices to construct an optimal conditional transport mapping $\gamma_z^*$ for $z\in\mathrm{supp}\,\hP$.
By the change-of-measure technique with $\zeta(z') = \frac{\diff\gamma_z(z')}{\diff\nu_z(z')}$, the inner supremum of \eqref{Eq:reformula:primal:WDRO:phi:reg} becomes
\begin{equation}
\sup_{\zeta\in \cZ_+^*}~\Big\{ 
\bE_{z'\sim \nu_z}
\big[ 
f(z')\zeta(z') - \eta\phi(\zeta(z'))
\big]
:~
\bE_{\nu_z}[\zeta]=1
\Big\}.
\label{Eq:opt:zeta}
\end{equation}
We now construct the Lagrangian function associated with \eqref{Eq:opt:zeta} as
\[
\textsf{L}(\zeta, \mu) = \bE_{z'\sim \nu_z}
\big[ 
(f(z')-\mu)\zeta(z') - \eta\phi(\zeta(z'))
\big] + \mu.
\]
Recall from \citep[Proposition~3.3]{bonnans2000stability} that, if there exists $(\zeta_z^*,\mu_z^*)$ such that 
\[
\zeta_z^*\in \cZ_+^*,\quad
\bE_{\nu_z}[\zeta_z^*]=1,\quad
\zeta_z^*\in\argmax_{\zeta\in \cZ_+^*}~\textsf{L}(\zeta, \mu_z^*),
\]
it holds that $\zeta_z^*$ solves \eqref{Eq:opt:zeta}.
The proof is completed by substituting the expression of $\gamma^*$ in terms of $\gamma_z^*$ and then substituting the expression of $\gamma_z^*$ in terms of $\zeta_z^*$.
\QED 
\endproof

\begin{lemma}[{\citep[Section~3.2]{shapiro2017distributionally}}]\label{Lemma:phi:div}
Given a probability reference measure $\gamma\in\cP(\cZ)$ and regularization value $\eta>0$, consider the $\phi$-divergence regularized problem:
\[
\sup_{\gamma\in\cP(\cZ)}~\left\{
\bE_{z\sim\gamma}[f(z)] 
-
\eta\bE_{z\sim\nu}
\left[ 
\phi\left( 
\frac{\diff\gamma(z)}{\diff\nu(z)}
\right)
\right]
\right\}.
\]
There exists an optimal solution to this primal problem, and also, it can be reformulated as the dual problem
\[
\inf_{\mu\in\mathbb{R}}~\bigg\{ 
\mu + \bE_{z\sim\nu}\left[ 
\eta\phi^*\left(\frac{f(z) - \mu}{\eta}\right)
\right]
\bigg\}.
\]
\end{lemma}
\proof{Proof of Proposition~\ref{Pro:consistency}.}
It is easy to verify that
\begin{align*}
\mathrm{Optval}\eqref{Eq:primal:WDRO:phi:reg}
&=
\bE_{z\sim\hP}\Bigg[ 
\sup_{\gamma_z\in\cP(\cZ)}~\left\{
\bE_{z'\sim\gamma_z}[f(z')] 
-
\eta\bE_{z'\sim\nu_z}
\left[ 
\phi\left( 
\frac{\diff\gamma_z(z')}{\diff\nu_z(z')}
\right)
\right]
\right\}
\Bigg]\\
&\le \bE_{z\sim\hP}\Bigg[ 
\sup_{\gamma_z\in\cP(\cZ)}~\Big\{
\bE_{z'\sim\gamma_z}[f(z')] 
\Big\}
\Bigg]=\bE_{z\sim\hP}\left[ 
\max_{z'\in \mathbb{B}_{\rho}(z)}~f(z')
\right].
\end{align*}
Now, it suffices to show the other direction.
For fixed $\eta>0$ and $z$, let $\mu_{z,\eta}^*$ be the minimizer to 
\[
\inf_{\mu}~\bigg\{ 
\mu + \bE_{z'\sim\nu_z}\left[ 
(\eta\phi)^*\big(f(z') - \mu\big)
\right]
\bigg\}.
\]
\begin{enumerate}
    \item 
For the case where $\lim_{t\to\infty}\frac{\phi(t)}{t}<\infty$, by \citep{bayraksan2015data}, the dual formulation~\eqref{Eq:dual:WDRO:phi:reg} implicitly imposes an extra constraint:
\[
\mu_{z,\eta}^*\ge f(z') - \eta\lim_{t\to\infty}\frac{\phi(t)}{t}, \forall z'\in\mathrm{supp}\,\nu_z\implies 
\mu_{z,\eta}^*\ge \esssup_{\nu_z}~f - \eta\lim_{t\to\infty}\frac{\phi(t)}{t}.
\]
It follows that 
\begin{align*}
\eqref{Eq:dual:WDRO:phi:reg}&=
\bE_{z\sim\hP}~\left[ 
\mu_{z,\eta}^* + \bE_{z'\sim\nu_z}\left[ 
\eta\phi^*\left( 
\frac{f(z') - \mu_{z,\eta}^*}{\eta}
\right)
\right]
\right]\\
&\ge
\esssup_{\nu_z}~f - \eta\lim_{t\to\infty}\frac{\phi(t)}{t} + \eta
\bE_{z'\sim\nu_z}\left[ 
\phi^*\left( 
\frac{f(z') - \mu_{z,\eta}^*}{\eta}
\right)
\right]
\end{align*}
By taking $\eta\to0$ both sides, we find
\[
\mathrm{Optval}\eqref{Eq:primal:WDRO:phi:reg}
\ge \esssup_{\nu_z}~f.
\]
\item
For the case where $\lim_{t\to\infty}\frac{\phi(t)}{t}=\infty$, it holds that $\phi^*(s)\in(-\infty,\infty)$ for any finite $s$.
In this case, for fixed $\eta>0$,
\[
\eqref{Eq:dual:WDRO:phi:reg}=
\bE_{z\sim\hP}~\left[ 
\mu_{z,\eta}^* + \bE_{z'\sim\nu_z}\left[ 
\eta\phi^*\left( 
\frac{f(z') - \mu_{z,\eta}^*}{\eta}
\right)
\right]
\right].
\]
For sufficiently small $\eta$, assume on the contrary that $\mu_{z,\eta}^*<\esssup_{\nu_z}~f$, then the event $E_{z,\eta}:=\{z':~f(z')>\mu_{z,\eta}^*\}$ satisfies $\nu_z(E_{z,\eta})>0$.
\begin{itemize}
    \item 
For $z'\notin E_{z,\eta}$, 
\[
\lim_{\eta\to0}~\eta\phi^*\left( 
\frac{f(z') - \mu_{z,\eta}^*}{\eta}
\right)=\lim_{\eta\to0}~\eta\phi^*(0)=0.
\]
   \item
For $z'\in E_{z,\eta}$,
\[
\lim_{\eta\to0}~\eta\phi^*\left( 
\frac{f(z') - \mu_{z,\eta}^*}{\eta}
\right) = 
\lim_{t\to\infty}~
\frac{1}{t}\phi^*\left( 
t\big( 
f(z') - \mu_{z,\eta}^*
\big)
\right)\to 
\infty.
\]
\end{itemize}
Then it follows that 
\[
\begin{aligned}
&\bE_{z'\sim\nu_z}\left[ 
\eta\phi^*\left( 
\frac{f(z') - \mu_{z,\eta}^*}{\eta}
\right)
\right]\\
=&
\bE_{z'\sim\nu_z}\left[ 
\eta\phi^*\left( 
\frac{f(z') - \mu_{z,\eta}^*}{\eta}
\right)\textbf{1}(E_{z,\eta}^c)
\right]
+
\bE_{z'\sim\nu_z}\left[ 
\eta\phi^*\left( 
\frac{f(z') - \mu_{z,\eta}^*}{\eta}
\right)\textbf{1}(E_{z,\eta})
\right]\to\infty.
\end{aligned}
\]
In summary, under the case where $\mu_{z,\eta}^*<\esssup_{\nu_z}~f$, $\eqref{Eq:dual:WDRO:phi:reg}\to\infty$ as $\eta\to0$, which is a contradiction.
Therefore, $\mu_{z,\eta}^*\ge \esssup_{\nu_z}~f$, which follows that 
\[
\eqref{Eq:dual:WDRO:phi:reg}
\ge
\esssup_{\nu_z}~f + \eta
\bE_{z'\sim\nu_z}\left[ 
\phi^*\left( 
\frac{f(z') - \mu_{z,\eta}^*}{\eta}
\right)
\right]
\]
By taking $\eta\to0$ both sides, we obtain $\mathrm{Optval}\eqref{Eq:primal:WDRO:phi:reg}
\ge \esssup_{\nu_z}~f.$
\end{enumerate}
\QED

\clearpage
\section{Proofs of Technical Results in Section~\ref{Sec:opt}}\label{Appendix:Sec:opt}

\proof{Proof of Proposition~\ref{Pro:complexity:alg:Eq:expression:R}.}
We define \textbf{Case~1} as the scenarios where $\phi'(s)\to-\infty$ as $s\to0+$, and \textbf{Case~2} as the scenarios where $\phi'(s)\to K$ as $s\to0+$, with the constant $K>-\infty$ being lower bounded.

For any fixed Lagrangian multiplier $\mu$, $\gamma^*(\mu)\in\mathbb{R}_+^m$ is the optimum solution to $\max_{\gamma\in\mathbb{R}_+^m}~\mathcal{L}(\mu, \gamma)$ if and only if 
\[
f_i-\mu - \eta\phi'(m(\gamma^*(\mu))_i)\le0, \forall i,\quad 
(\gamma^*(\mu))_i\cdot\Big( 
f_i-\mu - \eta\phi'(m(\gamma^*(\mu))_i)
\Big)=0.
\] 
Under \textbf{Case~1}, the above optimality condition simplifies into $f_i-\mu - \eta\phi'(m(\gamma^*(\mu))_i)=0, \forall i$, which implies
\[
(\gamma^*(\mu))_i = \frac{1}{m}(\phi')^{-1}\Big( 
\frac{f_i -\mu}{\eta}
\Big).
\]
Under \textbf{Case~2}, the above optimality condition simplifies into
\[
(\gamma^*(\mu))_i = \left\{ 
\begin{aligned}
0,&\quad\text{if }i\in \mathcal{N}\triangleq \Big\{i\in[m]:~f_i\le \mu + \eta K\Big\},\\
\frac{1}{m}(\phi')^{-1}\Big( 
\frac{f_i - \mu}{\eta}
\Big),&\quad \text{otherwise}.
\end{aligned}
\right.
\]

Therefore, the remaining task of Algorithm~\eqref{alg:Eq:expression:R} is to find the optimal Lagrangian multiplier $\mu$ such that 
\[
h(\mu):=\sum_{i\in[m]}(\gamma^*(\mu))_i-1=\frac{1}{m}\sum_{i\in[m]\setminus\mathcal{N}}(\phi')^{-1}\Big( 
\frac{f_i -\mu}{\eta}
\Big)-1=0.
\]
Due to the strict convexity of $\phi$, $h(\mu)$ is strictly decreasing in $\mu$.
Also, it can be verified that the optimal multiplier belongs to the interval $[\underline{\mu}, \overline{\mu}]$:
\begin{itemize}
    \item 
By the increasing property of $(\phi')^{-1}$, it holds that $h(\underline{\lambda})\ge0$;
    \item
Under \textbf{Case 1}, it holds that $h(\overline{\mu})\le0$.
Under \textbf{Case 2}, it holds that $h(\overline{\mu})\le-1$.
\end{itemize}
Hence, we only need to perform $\cO(\log\frac{1}{\epsilon})$ iterations of bisection search to obtain a near-optimal multiplier with $\epsilon$ precision.
At each iteration of bisection search, the worst-case computational cost is $O(m)$.
To compute the index set $\mathcal{N}$ at each iteration, we need to enumerate all support points $\{f_1,\ldots,f_m\}$, whose computational cost is $\cO(m)$.
In summary, the overall cost is $\cO(m\log\frac{1}{\epsilon})$.
\QED
\endproof

\begin{proposition}[Error Bound on Function Approximation]\label{Proposition:bound:approximate}
Under Assumption~\ref{Assumption:idf:chi2}\ref{Assumption:idf}, it holds that $0\le F(\theta) - F^{\ell}(\theta)\le G_{\text{idf}}\cdot 2^{-\ell}, \forall\theta\in\Theta.$
\end{proposition}

\proof{Proof of Proposition~\ref{Proposition:bound:approximate}.}
It is worth noting that 
\begin{align*}
F(\theta) - F^{\ell}(\theta)&=\bE_{z\sim\hP}
\bE_{\{z_i'\}_{i\in[2^\ell]}\sim \nu_z}~\Big[ 
R(\theta;z) - 
\widehat{R}\big(\theta;z, \{z_i'\}_{i\in[2^\ell]}\big)
\Big],
\end{align*}
where 
\[
\begin{aligned}
R(\theta;z)&=\inf_{\mu}~\Big\{ 
\mu + \bE_{z\sim\nu_z}[(\eta\phi)^*(f_{\theta}(z') - \mu)]
\Big\},\\
\widehat{R}\big(\theta;z, \{z_i'\}_{i\in[2^\ell]}\big)&=\inf_{\mu}~\Big\{ 
\mu + \frac{1}{2^{\ell}}\sum_{i\in[2^{\ell}]}[(\eta\phi)^*(f_{\theta}(z'_i) - \mu)]
\Big\}.
\end{aligned}
\]
By Jensen's inequality, it holds for any fixed $(z,\theta)$ that 
\[
R(\theta;z)\ge \bE_{\{z_i'\}_{i\in[2^\ell]}\sim \nu_z}[\widehat{R}\big(\theta;z, \{z_i'\}_{i\in[2^\ell]}\big)],
\]
and therefore $F(\theta) - F^{\ell}(\theta)\ge 0$.

On the other hand, $R(\theta;z)$ denotes the optimal value of the standard $\phi$-divergence DRO with reference distribution $\nu_z$, and $\widehat{R}\big(\theta;z, \{z_i'\}_{i\in[2^\ell]}\big)$ denotes its sample estimate using $2^{\ell}$ i.i.d. samples generated from $\nu_z$.
By \citep[Proposition~1]{levy2020large}, it holds for any fixed $z$ that 
\[
\bE_{\{z_i'\}_{i\in[2^\ell]}\sim \nu_z}~\Big[ 
R(\theta;z) - 
\widehat{R}\big(\theta;z, \{z_i'\}_{i\in[2^\ell]}\big)
\Big]\le \frac{G_{\text{idf}}}{2^{\ell}}.
\]
The proof is completed.
\QED
\endproof
To prove Proposition~\ref{Pro:bias:var:RTMLMC}, we rely on the following technical lemma that has been revealed in literature.
\begin{lemma}[{Lemma~6 in \citep{levy2020large}}]\label{Lemma:bound:simplex}
Let $\gamma\in\Delta^m$ with $m$ being an even integer.
Let $\mathcal{I}$ be a random subset of $[1:m]$ of size $m/2$.
Then it holds that 
\[
\bE\left[ 
\sum_{i\in\mathcal{I}}\gamma_i-\frac{1}{2}
\right]^2\le \frac{1}{2m}D_{\chi^2}(\gamma, \frac{1}{m}\textsf{1}).
\]
\end{lemma}
\proof{Proof of Proposition~\ref{Pro:bias:var:RTMLMC}.}
\begin{enumerate}
    \item 
By definition, SG and RT-MLMC estimators $V^{\text{SG}}(\theta)$ and $V^{\text{RT-MLMC}}(\theta)$ are the unbiased gradient estimators from some objective function $\widetilde{F}^{\ell}(\theta)$ such that $|\widetilde{F}^{\ell}(\theta) - F^{\ell}(\theta)|\le \epsilon$.
Therefore, the bias can be bounded as
\[
|
\widetilde{F}^{\ell}(\theta) 
-
F(\theta)
|\le 
|\widetilde{F}^{\ell}(\theta) - F^{\ell}(\theta)| + 
|F^{\ell}(\theta) - F(\theta)|\le \epsilon + \frac{G_{\text{idf}}}{2^{\ell}},
\]
where the last inequality follows from Proposition~\ref{Proposition:bound:approximate}.
\item
By definition, 
\begin{align*}
\left\| 
V^{\text{SG}}(\theta)
\right\|^2&=\left\|
\frac{1}{\nout_L}\sum_{i=1}^{\nout_L}g^{L}(\theta, \zeta_i^L)
\right\|^2\le \frac{1}{\nout_L}\sum_{i=1}^{\nout_L}\left\| 
g^{L}(\theta, \zeta_i^L)
\right\|^2.
\end{align*}
Since $\{\zeta_i^L\}_i$ are $\nout_L$ i.i.d. copies of $\zeta^L$, it holds that 
\[
\bE\left\| 
V^{\text{SG}}(\theta)
\right\|^2\le \bE\left\| 
g^{L}(\theta, \zeta^L)
\right\|^2
\]
To bound $\bE\left\| 
g^{L}(\theta, \zeta^L)
\right\|^2$, we define the following notations.
Let $\widehat{\gamma}$ be the optimal solution to $\widehat{R}\big(\theta;z, \{z_i'\}_{i\in[1:2^L]}\big)$ defined in \eqref{Eq:hR:z:z}, and $\widetilde{\gamma}$ be the estimated solution used by the estimator $g^{L}(\theta, \zeta^L)$.
As a consequence, 
\begin{align*}
\|g^{\ell}(\theta,\zeta^{\ell})\|&=\left\|\sum_{i\in[1:2^{\ell}]}\widetilde{\gamma}_i\nabla_{\theta}f_{\theta}(z_i')\right\|
\le L_f\sum_{i\in[1:2^{\ell}]}\widetilde{\gamma}_i\\
&\le L_f\sum_{i\in[1:2^{\ell}]}\Big(\widehat{\gamma}_i + \|\widehat{\gamma} - \widetilde{\gamma}\|_{\infty}\Big)=L_f\cdot\left[ 
1 + \sqrt{\frac{2\epsilon}{\kappa\eta}}
\right].
\end{align*}
Therefore,
\[
\bE\left\| 
g^{L}(\theta, \zeta^L)
\right\|^2\le L_f^2\left[ 
1 + \sqrt{\frac{2\epsilon}{\kappa\eta}}
\right]^2\le 2L_f^2\left[1 + \frac{2\epsilon}{\kappa\eta}\right].
\]
Following the similar argument as in bounding $\bE\left\| 
V^{\text{SG}}(\theta)
\right\|^2$, we find
\begin{align*}
\bE\left\| 
V^{\text{RT-MLMC}}(\theta)
\right\|^2
&\le
\bE_{\widehat{L}_1}\bE_{\zeta^{\widehat{L}_1}}\left\|
\frac{1}{\mathbb{P}(\widehat{L}=\widehat{L}_1)}G^{\widehat{L}_1}(\theta, \zeta^{\widehat{L}_1})
\right\|^2\\
&=\sum_{\ell=0}^L\mathbb{P}(\widehat{L}=\ell)\bE_{\zeta_i^{\ell}}\left\|
\frac{1}{\mathbb{P}(\widehat{L}=\ell)}G^{\ell}(\theta, \zeta^{\ell})
\right\|^2=\sum_{\ell=0}^L\frac{1}{\mathbb{P}(\widehat{L}=\ell)}\cdot \bE_{\zeta^{\ell}}\left\|
G^{\ell}(\theta, \zeta^{\ell})
\right\|^2.
\end{align*}
It suffices to bound $\bE_{\zeta^{\ell}}\left\|
G^{\ell}(\theta, \zeta^{\ell})
\right\|^2$ for fixed level $\ell=0,1,\ldots,L$.
To simplify notation,
\begin{itemize}
    \item 
Let $\gamma, \gamma', \gamma''$ be the estimated optimal solutions corresponding to the objectives $\widetilde{R}\big(\theta;z, \{z_i'\}_{i\in[1:2^\ell]}\big),\widetilde{R}\big(\theta;z, \{z_i'\}_{i\in[1:2^{\ell-1}]}\big)$, and $\widetilde{R}\big(\theta;z, \{z_i'\}_{i\in[2^{\ell-1}+1:2^\ell]}\big)$ defined in \eqref{Eq:tilde:R}, respectively.
\item
Let $\bar{\gamma}, \bar{\gamma}', \bar{\gamma}''$ be the optimal solutions for $\widetilde{R}\big(\theta;z, \{z_i'\}_{i\in[1:2^\ell]}\big), \widetilde{R}\big(\theta;z, \{z_i'\}_{i\in[1:2^{\ell-1}]}\big)$ and $\widetilde{R}\big(\theta;z, \{z_i'\}_{i\in[2^{\ell-1}+1:2^\ell]}\big)$ defined in \eqref{Eq:hR:z:z}, respectively.
\end{itemize}
Then it holds that 
\begin{align*}
\|G^{\ell}(\theta,\zeta^{\ell})\|&=\left\|\sum_{i\in[1:2^{\ell}]}
\Big( 
\gamma_i-\frac{1}{2}\gamma'_i\cdot 1\{i\in[1:2^{\ell-1}]\} - \frac{1}{2}\gamma''_{i-2^{\ell-1}}\cdot 1\{i\in[2^{\ell-1}+1:2^{\ell}]\}
\Big)\nabla_{\theta}f_{\theta}(z_i')\right\|
\\&
\le L_f\sum_{i\in[1:2^{\ell-1}]}|\gamma_i - \gamma'_i/2| + L_f\sum_{i\in[2^{\ell-1}+1:2^{\ell}]}|\gamma_i - \gamma''_{i-2^{\ell-1}}/2|.
\end{align*}
Recall that for each $i\in[1:2^{\ell-1}]$, $\gamma_i=\frac{1}{m}(\phi')^{-1}(\frac{f(z_i)-\mu}{\eta})$ and $\gamma'_i/2=\frac{1}{m}(\phi')^{-1}(\frac{f(z_i)-\mu'}{\eta})$ for constants $\mu,\mu'\in\mathbb{R}$.
Since $\phi$ is strongly convex, $(\phi')^{-1}(\cdot)$ is a strictly increasing function, and $\gamma_i - \gamma'_i/2$ is always of a constant sign for all $i\in[1:2^{\ell-1}]$.
Therefore,
\begin{align*}
\sum_{i\in[1:2^{\ell-1}]}|\gamma_i - \gamma'_i/2|&=\left|
\sum_{i\in[1:2^{\ell-1}]}\gamma_i - \frac{1}{2}\sum_{i\in[1:2^{\ell-1}]}\gamma_i'
\right|\\
&\le 
2^{\ell-1}\|\gamma - \bar{\gamma}\|_{\infty} + 2^{\ell}\|\gamma' - \bar{\gamma}'\|_{\infty}+ \left|
\sum_{i\in[1:2^{\ell-1}]}\bar{\gamma}_i - \frac{1}{2}\sum_{i\in[1:2^{\ell-1}]}\bar{\gamma}'_i
\right|\\
&\le \sqrt{\frac{2\epsilon}{\kappa\eta}} +  \left|
\sum_{i\in[1:2^{\ell-1}]}\bar{\gamma}_i - \frac{1}{2}
\right|,
\end{align*}
where the first inequality is by triangular inequality, the second inequality is by Proposition~\ref{Pro:complexity:alg:Eq:expression:R}\ref{Pro:complexity:alg:Eq:expression:R:III} and the relation $\sum_{i\in[1:2^{\ell-1}]}\bar{\gamma}'_i=1$.
One can follow the similar procedure to bound $\sum_{i\in[2^{\ell-1}+1:2^{\ell}]}|\gamma_i - \gamma''_{i-2^{\ell-1}}/2|$.
As a consequence,
\begin{align*}
\bE_{\zeta^{\ell}}\|G^{\ell}(\theta,\zeta^{\ell})\|^2&\le L_f^2\bE\left[2\sqrt{\frac{2\epsilon}{\kappa\eta}} + \left|
\sum_{i\in[1:2^{\ell-1}]}\bar{\gamma}_i - \frac{1}{2}
\right|
+
\left|
\sum_{i\in[2^{\ell-1}+1:2^{\ell}]}\bar{\gamma}_i - \frac{1}{2}
\right|
\right]^2\\
&\le 3L_f^2\cdot \left(\frac{8\epsilon}{\kappa\eta} + \bE\left|
\sum_{i\in[1:2^{\ell-1}]}\bar{\gamma}_i - \frac{1}{2}
\right|^2
+
\bE\left|
\sum_{i\in[2^{\ell-1}+1:2^{\ell}]}\bar{\gamma}_i - \frac{1}{2}
\right|^2
\right)\\
&\le \frac{24L_f^2}{\kappa\eta}\cdot \epsilon + \frac{3L_f^2D_{\chi^2}(\bar{\gamma}, \frac{1}{2^{\ell}}\textsf{1})}{2^{\ell}}\\
&\le \frac{24L_f^2}{\kappa\eta}\cdot \epsilon + \frac{3L_f^2C}{2^{\ell}}.
\end{align*}
Finally,
\[
\bE\left\| 
V^{\text{RT-MLMC}}(\theta)
\right\|^2\le \frac{96L_f^2}{\kappa\eta}\cdot 2^L\cdot \epsilon +6(L+1)L_f^2C.
\]
\item
Since the random vectors $g^L(\theta, \zeta_i^L)$ for $i=1,\ldots,\nout_L$ are i.i.d., it holds that 
\[
\Var[V^{\text{SG}}(\theta)]=\Var\left[ 
\frac{1}{\nout_L}\sum_{i=1}^{\nout_L}g^{L}(\theta, \zeta_i^L)
\right]=\frac{\Var[g^{L}(\theta, \zeta^L)]}{\nout_L}\le \frac{\bE\|g^{L}(\theta, \zeta^L)\|^2}{\nout_L}\le \frac{2L_f^2\left[1 + (2\epsilon)/(\kappa\eta)\right]}{\nout_L}.
\]
The same argument applies when bounding $\Var[V^{\text{RT-MLMC}}(\theta)]$.
\item
For fixed $i=1,\ldots,\nout_L$, computing $g^L(\theta,\zeta_i^L)$ requires generating $2^L$ samples and then solve the penalized $\phi$-divergence DRO with $2^L$ support points with controlled optimality gap $\epsilon$. 
According to Lemma~\ref{Pro:complexity:alg:Eq:expression:R}, its complexity is $\cO(2^L\log\frac{1}{\epsilon})$.
Hence, generating the SG estimator $V^{\text{SG}}(\theta)$ has cost $\cO(\nout_L\cdot 2^L\log\frac{1}{\epsilon})$.

For fixed $i=1,\ldots,\nout_L$ and $\ell=0,\ldots,L$, computing $G^{\ell}(\theta,\zeta_i^{\ell})$, according to the definition in \eqref{Eq:G:ell}, requires computatonal cost $\cO(2^{\ell+1}\log\frac{1}{\epsilon})$.
Hence, generating the RT-MLMC estimator has expected computational cost
\begin{equation}
\nout_L\cdot \sum_{\ell=0}^L\bP(\widehat{L}=\ell)\cdot \cO(2^{\ell+1}\log\frac{1}{\epsilon})=\cO(\nout_L\cdot L\log\frac{1}{\epsilon}).
\tag*{\QED}
\end{equation}
\end{enumerate}
\endproof

\proof{Proof of Theorem~\ref{Thm:nonsmooth:cvx}.}
We first show the generic result on SGD with biased gradient estimators.
Denote by $\theta_*$ the optimal solution to $\min~F(\theta)$, and $\widetilde{\theta}_*$ is the optimal solution to $\min~\widetilde{F}(\theta)$, where SG and RT-MLMC estimators are unbiased gradient estimators of $\widetilde{F}(\cdot)$.
Based on the triangle inequality, it holds that
\begin{align*}
\bE\big[F(\widetilde{\theta}_{1:T}) - F(\theta_*)\big]&
\le \bE\big[F(\widetilde{\theta}_{1:T}) - \widetilde{F}(\widetilde{\theta}_{1:T})\big]
+
\bE\big[\widetilde{F}(\widetilde{\theta}_{1:T}) - \widetilde{F}(\theta_*)\big]
+
\bE\big[\widetilde{F}(\theta_*) - F(\theta_*)\big]\\
&\le 2\|\widetilde{F} - F\|_{\infty} + \bE\big[\widetilde{F}(\widetilde{\theta}_{1:T}) - \widetilde{F}(\widetilde{\theta}_*)\big],
\end{align*}
where the last inequality is because of the sub-optimality of $\theta_*$ in terms of the objective $\widetilde{F}$.
According to Proposition~\ref{Pro:bias:var:RTMLMC}, it holds that 
\[
\|\widetilde{F} - F\|_{\infty}\le \epsilon + \frac{G_{\text{idf}}}{2^L}.
\]
According to Lemma~\ref{Lemma:standard:SGD}, for SG or RT-MLMC estimator $V(\theta)$ satisfying $\bE\|V(\theta)\|_2^2\le M^2$ and if we take step size $\gamma=\frac{\widetilde{D}_*}{M\sqrt{T}}$, it holds that
\[
\bE\big[\widetilde{F}(\widetilde{\theta}_{1:T}) - \widetilde{F}(\widetilde{\theta}_*)\big]
\le \frac{\widetilde{D}_*M}{\sqrt{T}},
\]
where the constant 
\[
\widetilde{D}_*=\widetilde{F}(\theta_1) - \widetilde{F}(\widetilde{\theta}_*)\le F(\theta_1)-F(\theta_*) + 2\|\widetilde{F} - F\|_{\infty}\le  F(\theta_1)-F(\theta_*) + 2\left[\epsilon + \frac{G_{\text{idf}}}{2^L}\right].
\]
In summary, the error bound for $\widetilde{\theta}_{1:T}$ becomes
\[
\bE\big[F(\widetilde{\theta}_{1:T}) - F(\theta_*)\big]
\le 2\left[\epsilon + \frac{G_{\text{idf}}}{2^L}\right] + \frac{M\Big[F(\theta_1)-F(\theta_*) + 2\left[\epsilon + \frac{G_{\text{idf}}}{2^L}\right]\Big]}{\sqrt{T}}.
\]
\noindent{\bf SG Estimator.}
For SG estimator $V^{\text{SG}}(\theta)$, by Lemma~\ref{Pro:bias:var:RTMLMC}, it holds that $M=2L_f^2[1 + 2\epsilon/(\kappa\eta)]$.
To obtain the desired error bound $\bE\big[F(\widetilde{\theta}_{1:T}) - F(\theta_*)\big]\le\delta$, we specify hyper-parameters such that
\[
2\left[\epsilon + \frac{G_{\text{idf}}}{2^L}\right]\le \frac{\delta}{2},\qquad
\frac{M\Big[F(\theta_1)-F(\theta_*) + 2\left[\epsilon + \frac{G_{\text{idf}}}{2^L}\right]\Big]}{\sqrt{T}}\le\frac{\delta}{2}.
\]
We take $\epsilon=\frac{\delta}{8}$ and $L=\log\frac{8G_{\text{idf}}}{\delta}$ to make the relation on the left-hand-side holds.
Then $M=\cO(1)$.
To make the other relation holds, it suffices to take
\[
T\ge \frac{4M^2\Big[F(\theta_1)-F(\theta_*) + \frac{\delta}{2}\Big]^2}{\delta^2}
=
\cO(1/\delta^2).%
\]
\noindent{\bf RT-MLMC Estimator.}
For SG estimator $V^{\text{RT-MLMC}}(\theta)$, by Lemma~\ref{Pro:bias:var:RTMLMC}, it holds that $M=\frac{96L_f^2}{\kappa\eta}\cdot (2^L\epsilon) +6(L+1)L_f^2C$.
To obtain the desired error bound $\bE\big[F(\widetilde{\theta}_{1:T}) - F(\theta_*)\big]\le\delta$, we specify hyper-parameters such that
\[
2\left[\epsilon + \frac{G_{\text{idf}}}{2^L}\right]\le \frac{\delta}{2},\qquad
\frac{M\Big[F(\theta_1)-F(\theta_*) + 2\left[\epsilon + \frac{G_{\text{idf}}}{2^L}\right]\Big]}{\sqrt{T}}\le\frac{\delta}{2}.
\]
Following the same argument as in the SG estimator part, we take $\epsilon=\frac{\delta}{8}$ and $L=\log\frac{8G_{\text{idf}}}{\delta}$.
Then $M=\cO(\log\frac{1}{\delta})$.
To make the other relation holds, it suffices to take
\[
T\ge \frac{4M^2\Big[F(\theta_1)-F(\theta_*) + \frac{\delta}{2}\Big]^2}{\delta^2}
=
\cO((\log1/\delta)^2/\delta^2).%
\]
\endproof


In the following, we present a technical lemma that is helpful for the proof of Theorem~\ref{Theorem:complexity:BSMD}.
The proof of this technical lemma follows from \citep[Lemma~3.1]{hu2020sample} and  \citep[Proposition~4.1]{hu2021biasvar}.
\begin{lemma}\label{Lemma:tech:sum}
\begin{enumerate}
    \item
Under Assumption~\ref{Assumption:throughout:loss:updated}\ref{Assumption:throughout:loss:bound}, it holds that 
\[
\left| 
F^{\ell}(\theta) - F(\theta)
\right|\le \eta e^{2B/\eta}\cdot 2^{-(\ell+1)},\quad \forall \theta\in\Theta.
\]
    \item\label{Lemma:tech:sum:2}
Under Assumptions~\ref{Assumption:throughout:loss:updated}\ref{Assumption:throughout:loss:bound} and \ref{Assumption:throughout:loss:updated}\ref{Assumption:throughout:loss:lip}, it holds that 
\[
\left\| 
\nabla F^{\ell}(\theta) - \nabla F(\theta)
\right\|_2^2\le L_f^2e^{4B/\eta}\cdot 2^{-\ell},\quad \forall\theta\in\Theta.
\]
    \item\label{Lemma:tech:sum:3}
Under Assumptions~\ref{Assumption:throughout:loss:updated}\ref{Assumption:throughout:loss:bound} and \ref{Assumption:throughout:loss:updated}\ref{Assumption:throughout:loss:lip}, it holds that 
\[
\mathbb{E}\left[
\left\| 
G^{\ell}(\theta,\zeta^{\ell})
\right\|_2^2\right]\le L_f^2e^{4B/\eta}\cdot 2^{-\ell},\quad \forall\theta\in\Theta.
\]
    \item\label{Lemma:tech:sum:4}
Under Assumptions~\ref{Assumption:throughout:loss:updated}\ref{Assumption:throughout:loss:bound}, \ref{Assumption:throughout:loss:updated}\ref{Assumption:throughout:loss:lip} and \ref{Assumption:throughout:loss:updated}\ref{Assumption:throughout:loss:smooth},
it holds that for any $\ell\ge0$, $F^{\ell}(\theta)$ is $\overline{S}$-smooth with
\begin{equation}
\overline{S}:=(S_f^2 + L_f^2/\eta)e^{B/\eta} + L_f^2/\eta e^{2B/\eta}.
\label{Eq:Lemma:tech:sum:4}
\end{equation}
\end{enumerate}    
\end{lemma}
Now we present the formal proof of Theorem~\ref{Theorem:complexity:BSMD}.
\proof{Proof of Theorem~\ref{Theorem:complexity:BSMD}.}
At the beginning, it is without the loss of generality to assume that there exists $\mathfrak{c}, \mathfrak{d}$ such that 
\[
\mathfrak{c}\|\cdot\|_2\le \|\cdot\|_{\omega}\le \mathfrak{d}\|\cdot\|_{2},
\]
where $\|\cdot\|_{\omega}$ is the norm function used in defining the distance generating function for proximal mapping.

\begin{enumerate}
\item
We first specify the maximum level $L$ such that $2L_f^2e^{4B/\eta}\cdot 2^{-L}\le \frac{1}{2}\epsilon^2$, i.e., 
\[
L=\left\lceil\frac{1}{\log2}\left[ 
\log\frac{4L_f^2\cdot e^{4B/\eta}}{\epsilon^2}
\right]\right\rceil.
\]
It suffices to specify hyper-parameters $\nout_L, T, \gamma$ to make
\[
2\bE\left\|
\frac{1}{\gamma}
\left[
\tilde{\theta} - \prox_{\tilde{\theta}}\big(\gamma \nabla F^L(\tilde{\theta})\big)\right]
\right\|^2_{2}\le \frac{1}{2}\epsilon^2.
\]
Before applying Lemma~\ref{Lemma:SMD:smooth:noncvx} to derive upper bound on the left-hand-side term, it is worth noting that
\begin{itemize}
    \item
According to Lemma~\ref{Lemma:tech:sum}\ref{Lemma:tech:sum:4}, the objective $F^L(\theta)$ is $\mathfrak{c}^{-1}\mathfrak{d}\overline{S}$-smooth (with respect to $\|\cdot\|_{\omega}$) with the constant $\overline{S}$ defined in \eqref{Eq:Lemma:tech:sum:4}.
    \item
According to Lemma~\ref{Lemma:tech:sum}\ref{Lemma:tech:sum:3}, the term 
\[
\bE\left\| 
v(\theta_t) - \nabla F^L(\theta_t)
\right\|_{\omega}^2\le \frac{2\mathfrak{d}^2(L+1)L_f^2e^{4B/\eta}}{\nout_L}.
\]
\end{itemize}
Therefore, when taking the step size $\gamma=\kappa/(2\overline{S})$, it holds that 
\[
2\bE\left\|
\frac{1}{\gamma}
\left[
\tilde{\theta} - \prox_{\tilde{\theta}}\big(\gamma \nabla F^L(\tilde{\theta})\big)\right]
\right\|^2_{2}\le 
\frac{16\mathfrak{c}^{-3}\mathfrak{d}\overline{S}\big(F^L(\theta_1) - \min_{\theta}~F^L(\theta)\big)}{\kappa^2T} + \frac{24\mathfrak{c}^{-2}\mathfrak{d}^2\cdot(L+1)L_f^2e^{4B/\eta}}{\kappa^2\nout_L},
\]
With the configuration of the following hyper-parameters, one can guarantee the RT-MLMC scheme finds $\epsilon$-stationary point:
\begin{multline*}
\nout_L=
\left\lceil 
\frac{96\mathfrak{d}^2(L+1)L_f^2e^{4B/\eta}}{\kappa^2\mathfrak{c}^2\epsilon^2}
\right\rceil
,\qquad L=\left\lceil\frac{1}{\log2}\left[ 
\log\frac{4L_f^2\cdot e^{4B/\eta}}{\epsilon^2}
\right]\right\rceil, \\
T=\left\lceil
\frac{64\mathfrak{d}\overline{S}\big(F^L(\theta_1) - \min_{\theta}~F^L(\theta)\big)}{\kappa^2\mathfrak{c}^3\epsilon^2}
\right\rceil,\qquad 
\gamma=\kappa\mathfrak{c}/(2\mathfrak{d}\overline{S}).
\end{multline*}
\item
The proof in this part is a simple corollary from \citep[Corollary 4.1]{hu2021biasvar}.
With the configuration of the following hyper-parameters, one can guarantee the RT-MLMC scheme finds $\epsilon$-stationary point:
\begin{multline*}
\nout_L=
1
,\qquad L=\left\lceil\frac{1}{\log2}\left[ 
\log\frac{4L_f^2\cdot e^{4B/\eta}}{\epsilon^2}
\right]\right\rceil,\\
T=\left\lceil
\frac{128\big(F^L(\theta_1) - \min_{\theta}~F^L(\theta)\big)\overline{S}M_2}{\epsilon^4}
\right\rceil,\qquad 
\gamma=\sqrt{\frac{2\big(F^L(\theta_1) - \min_{\theta}~F^L(\theta)\big)}{\overline{S}TM_2}},
\end{multline*}
where the constant $M_2:=2(L+1)L_f^2e^{4B/\eta}$.

\end{enumerate}

\QED
\endproof

\clearpage
\section{Proofs of Technical Results in Section~\ref{Sec:reg}}\label{Appendix:Sec:reg}

We first show the following technical result, from which one can easily derive the main result in Section~\ref{Sec:reg}.
\begin{proposition}\label{Proposition:linearization}
Under Assumption~\ref{Assumption:f:smooth}, it holds that
$\mathcal{E}_{\hP}(f;\rho,\eta) = \widetilde{\mathcal{E}}_{\hP}(f;\rho,\eta) + O(\rho^2),$
where $O(\cdot)$ hides the multiplicative constant dependent on $\mathbb{E}_{z\sim\hP}[S(z)]$.
\end{proposition}

\proof{Proof of Proposition~\ref{Proposition:linearization}.}
By definition, 
\begin{align*}
&\mathcal{E}_{\hP}(f;\rho,\eta)=
\bE_{z\sim\hP}\Bigg[ 
\sup_{\gamma_z\in\cP(\cZ)}~\left\{
\bE_{z'\sim\gamma_z}[f(z')
-f(z)
] 
-
\eta\bE_{z'\sim\nu_z}
\left[ 
\phi\left( 
\frac{\diff\gamma_z(z')}{\diff\nu_z(z')}
\right)
\right]
\right\}
\Bigg].
\end{align*}
For any $z'\in\mathrm{supp}\,\gamma_z\subseteq \mathbb{B}_{\rho}(z)$, it holds that 
\begin{align*}
&|f(z') - f(z) - \nabla f(z)\trans(z-z')|
=|\nabla f(\tilde{z})\trans(z-z') - \nabla f(z)\trans(z-z')|\\
=&|(\nabla f(\tilde{z}) - \nabla f(z))\trans (z-z')|\le 
\|\nabla f(\tilde{z}) - \nabla f(z)\|_*\|z -z'\|\\
=&\|\tilde{z} - z\|\cdot \|z - z'\|\cdot S(z)\le S(x)\|z - z'\|^2\le S(z)\rho^2,
\end{align*}
where the second equality is by the mean value theorem and take $\tilde{z}$ to be some point on the line segment between $z$ and $z'$,
and the second inequality is based on the fact that $z'\in \mathbb{B}_{\rho}(z)$.

Based on the relation above, it holds that 
\[
\Big| 
\mathcal{E}_{\hP}(f;\rho,\eta)
-
\widetilde{\mathcal{E}}_{\hP}(f;\rho,\eta)
\Big|\le \mathbb{E}_{z\sim\hP}[S(z)]\rho^2,
\]
where
\begin{align*}
&\widetilde{\mathcal{E}}_{\hP}(f;\rho,\eta)\\
=&\bE_{z\sim\hP}\Bigg[ 
\sup_{\gamma_z\in\cP(\cZ)}~\left\{
\bE_{z'\sim\gamma_z}[\nabla f(z)\trans(z-z')
] 
-
\eta\bE_{z'\sim\nu_z}
\left[ 
\phi\left( 
\frac{\diff\gamma_z(z')}{\diff\nu_z(z')}
\right)
\right]
\right\}
\Bigg]\\
=&\mathbb{E}_{z\sim\hP}\left[ 
\inf_{\mu\in\mathbb{R}}\left\{ 
\mu + \mathbb{E}_{z'\sim\nu_z}\left[ 
\eta\phi^*\Big( 
\frac{\nabla f(z)\trans(z-z') - \mu}{\eta}
\Big)
\right]
\right\}
\right]\\
=&\mathbb{E}_{z\sim\hP}\left[ 
\inf_{\mu\in\mathbb{R}}\left\{ 
\mu + \mathbb{E}_{b\sim \beta}\left[ 
\eta\phi^*\Big( 
\frac{\rho\nabla f(z)\trans b - \mu}{\eta}
\Big)
\right]
\right\}
\right].
\end{align*}
By the change of variable technique that replaces $\mu$ by $\rho\mu$, 
\[
\widetilde{\mathcal{E}}_{\hP}(f;\rho,\eta) = \rho\cdot \mathbb{E}_{z\sim\hP}\left[ 
\inf_{\mu\in\mathbb{R}}\left\{ 
\mu + \frac{1}{\rho/\eta}\mathbb{E}_{b\sim \beta}\left[ 
\phi^*\Big( 
\frac{\rho}{\eta}\cdot(\nabla f(z)\trans b - \mu)
\Big)
\right]
\right\}
\right].
\]
The proof is completed.
\QED
\endproof

\proof{Proof of Theorem~\ref{Theorem:ref:eff}\ref{Proposition:3:smooth}}
For all $z\in\mathrm{supp}\,\hP$, by strong duality theory of $\phi$-divergence DRO~\citep{shapiro2017distributionally},
\[
\inf_{\mu\in\mathbb{R}}\left\{ 
\mu + \frac{1}{\rho/\eta}\mathbb{E}_{b\sim \beta}\left[ 
\phi^*\Big( 
\frac{\rho}{\eta}\cdot(\nabla f(z)\trans b - \mu)
\Big)
\right]
\right\}
=
\sup_{\beta'\in\mathcal{P}(\mathbb{B}_1(0))}~\Big\{ 
\mathbb{E}_{b\sim\beta'}[\nabla f(z)\trans b] - \frac{1}{\rho/\eta}\mathbb{D}_{\phi}(\beta',\beta)
\Big\}
\]
Since the optimization problem on the right-hand-side~(RHS) satisfies Slater's condition, the problem on the left-hand-side~(LHS) must contain a non-empty and bounded set of optimal solutions~\citep[Theorem~2.165]{bonnans2013perturbation}.
Subsequently, it holds from \citep[Theorem~5.4]{shapiro2021lectures} that, as $\rho/\eta\to C$,
\[
\inf_{\mu\in\mathbb{R}}\left\{ 
\mu + \frac{1}{\rho/\eta}\mathbb{E}_{b\sim \beta}\left[ 
\phi^*\Big( 
\frac{\rho}{\eta}\cdot(\nabla f(z)\trans b - \mu)
\Big)
\right]
\right\}
\to 
\inf_{\mu\in\mathbb{R}}\left\{ 
\mu + \frac{1}{C}\mathbb{E}_{b\sim \beta}\left[ 
\phi^*\Big( 
C\cdot(\nabla f(z)\trans b - \mu)
\Big)
\right]
\right\}.
\]
Therefore, 
\[
\widetilde{\mathcal{E}}_{\hP}(f;\rho,\eta) = \rho\cdot \mathbb{E}_{z\sim\hP}\left[ 
\inf_{\mu\in\mathbb{R}}\left\{ 
\mu + \frac{1}{C}\mathbb{E}_{b\sim \beta}\left[ 
\phi^*\Big( 
C\cdot(\nabla f(z)\trans b - \mu)
\Big)
\right]
\right\}
\right] + o(\rho) = \mathcal{R}_1(f; \rho,\eta) + o(\rho).
\]
By the relation above and Proposition~\ref{Proposition:linearization}, we obtain the desired result.
\QED
\endproof

\proof{Proof of Theorem~\ref{Theorem:ref:eff}\ref{Prop:rho:Reg}.}
According to Proposition~\ref{Proposition:linearization}, it suffices to build the error bound between $\widetilde{\mathcal{E}}_{\hP}(f;\rho,\eta) $ and $\mathcal{R}_2(f; \rho,\eta)$.
As $\rho/\eta\to\infty$, by repeating the proof argument as in Proposition~\ref{Pro:consistency}, one can show that
\[
\begin{aligned}
&\mathbb{E}_{z\sim\hP}\left[ 
\inf_{\mu\in\mathbb{R}}\left\{ 
\mu + \frac{1}{\rho/\eta}\mathbb{E}_{b\sim \beta}\left[ 
\phi^*\Big( 
\frac{\rho}{\eta}\cdot(\nabla f(z)\trans b - \mu)
\Big)
\right]
\right\}
\right]\\
=&\mathbb{E}_{z\sim\hP}\left[ 
\max_{b\in\mathbb{B}_1(0)}[\nabla f(z)\trans b]
\right] + o(1)
=\mathbb{E}_{z\sim\hP}\left[ 
\|\nabla f(z)\|_*
\right] + o(1).
\end{aligned}
\]
As such, 
\[
\Big| 
\widetilde{\mathcal{E}}_{\hP}(f;\rho,\eta)
-
\mathcal{R}_2(f; \rho,\eta)
\Big| = o(\rho).
\]
This completes the proof.
\QED
\endproof

\proof{Proof of Theorem~\ref{Theorem:ref:eff}\ref{Proposition:2:smooth}.}
Recall that 
\[
\begin{aligned}
\widetilde{\mathcal{E}}_{\hP}(f;\rho,\eta) &=\rho\cdot \mathbb{E}_{z\sim\hP}\left[ 
\inf_{\mu\in\mathbb{R}}\left\{ 
\mu + \frac{1}{\rho/\eta}\mathbb{E}_{b\sim \beta}\left[ 
\phi^*\Big( 
\frac{\rho}{\eta}\cdot(\nabla f(z)\trans b - \mu)
\Big)
\right]
\right\}
\right]\\
&=\rho\cdot \mathbb{E}_{z\sim\hP}\left[ 
\sup_{\beta'\in\mathcal{P}(\mathbb{B}_1(0))}~\Big\{ 
\mathbb{E}_{b\sim\beta'}[\nabla f(z)\trans b] - \frac{1}{\rho/\eta}\mathbb{D}_{\phi}(\beta',\beta)
\Big\}
\right]
\end{aligned}
\]
and it suffices to analyze the approximation of $\sup_{\beta'\in\mathcal{P}(\mathbb{B}_1(0))}~\Big\{ 
\mathbb{E}_{b\sim\beta'}[\nabla f(z)\trans b] - \frac{1}{\rho/\eta}\mathbb{D}_{\phi}(\beta',\beta)
\Big\}$ for $z\in\mathrm{supp}\,\hP$ when $\rho/\eta\to0$.

Note that we can re-write 
\[
\sup_{\beta'\in\mathcal{P}(\mathbb{B}_1(0))}~\Big\{ 
\mathbb{E}_{b\sim\beta'}[\nabla f(z)\trans b] - \frac{1}{\rho/\eta}\mathbb{D}_{\phi}(\beta',\beta)
\Big\} = \sup_{\Upsilon\ge0:~\mathbb{E}_{\beta}[\Upsilon]=1}~\Big\{ 
\mathbb{E}_{b\sim\beta}[\nabla f(z)\trans b\cdot \Upsilon(b)] -  \frac{1}{\rho/\eta}\cdot \mathbb{E}_{b\sim\beta}[\phi(\Upsilon(b))]
\Big\},
\]
where $\Upsilon$ is a non-negative random variable satisfying $\mathbb{E}_{\beta}[\Upsilon]=1$.
We use the change-of-variable technique to define $\overline{\Delta} = (\rho/\eta)^{-1}\cdot (\Upsilon-1)$, then by the relation $\mathbb{E}_{b\sim\beta}[\nabla f(z)\trans b]=0$, the optimization above can be equivalently reformulated as
\[
\frac{\rho}{\eta}\cdot
\sup_{\overline{\Delta}\ge-(\rho/\eta)^{-1}:~\mathbb{E}_{\beta}[\overline{\Delta}]=0}~\Big\{ 
\mathbb{E}_{b\sim\beta}[\nabla f(z)\trans b\cdot\overline{\Delta}(b)] -  (\rho/\eta)^{-2}\cdot \mathbb{E}_{b\sim\beta}[\phi(1 +\rho/\eta\cdot \overline{\Delta}(b))]
\Big\}.
\]
We take a feasible solution $\overline{\Delta}(b) = a\cdot \nabla f(z)\trans b$ with some constant $a>0$ provided that $a\cdot \|\nabla f(z)\|_*\le (\rho/\eta)^{-1}$.
Then, it holds that
\begin{align}
&\sup_{\overline{\Delta}\ge-(\rho/\eta)^{-1}:~\mathbb{E}_{\beta}[\overline{\Delta}]=0}~\Big\{ 
\mathbb{E}_{b\sim\beta}[\nabla f(z)\trans b\cdot\overline{\Delta}(b)] -  (\rho/\eta)^{-2}\cdot \mathbb{E}_{b\sim\beta}[\phi(1 +\rho/\eta\cdot \overline{\Delta}(b))]
\Big\}\nonumber\\
\ge&\sup_{a\ge0:~a\cdot \|\nabla f(z)\|_*\le (\rho/\eta)^{-1}}~\Big\{ 
a\cdot \mathbb{V}\text{ar}_{b\sim\beta}[\nabla f(z)\trans b] -  (\rho/\eta)^{-2}\cdot \mathbb{E}_{b\sim\beta}[\phi(1 +a\rho/\eta\cdot \nabla f(z)\trans b)]
\Big\}\label{Eq:lower:23}
\end{align}
Since $\phi(t)$ is two times continuously differentiable at $t=1$, as $\rho/\eta\to 0$, the following convergence holds uniformly for any bounded $a\cdot \nabla f(z)\trans b$:
\[
(\rho/\eta)^{-2}\cdot\phi(1 +a\rho/\eta\cdot \nabla f(z)\trans b)\to a^2(\nabla f(z)\trans b)^2\phi''(1)/2.
\]
Consequently, for any $\epsilon>0$, there exists $\delta_0>0$ such that as long as $\rho/\eta<\delta_0$, \eqref{Eq:lower:23} can be lower bounded as the following:
\[
\begin{aligned}
\eqref{Eq:lower:23}\ge&\sup_{a\ge0:~a\cdot \|\nabla f(z)\|_*\le (\rho/\eta)^{-1}}~\Big\{ 
a\cdot \mathbb{V}\text{ar}_{b\sim\beta}[\nabla f(z)\trans b] -
(1+\epsilon)a^2\cdot \mathbb{V}\text{ar}_{b\sim\beta}[\nabla f(z)\trans b]\cdot \phi''(1)/2
\Big\}\\
=&\frac{\mathbb{V}\text{ar}_{b\sim\beta}[\nabla f(z)\trans b]}{2(1+\epsilon)\phi''(1)}.
\end{aligned}
\]
Since $\epsilon$ can be arbitrarily small, it holds that 
\[
\widetilde{\mathcal{E}}_{\hP}(f;\rho,\eta)\ge \frac{\rho^2}{2\eta\cdot \phi''(1)}\cdot \bE_{z\sim \hP}\Big[ 
\Var_{b\sim\beta}[\nabla f(z)\trans b]
\Big] + o(\rho).
\]
For the upper bound, by strong duality result,
\begin{align*}
&\sup_{\overline{\Delta}\ge-(\rho/\eta)^{-1}:~\mathbb{E}_{\beta}[\overline{\Delta}]=0}~\Big\{ 
\mathbb{E}_{b\sim\beta}[\nabla f(z)\trans b\cdot\overline{\Delta}(b)] -  (\rho/\eta)^{-2}\cdot \mathbb{E}_{b\sim\beta}[\phi(1 +\rho/\eta\cdot \overline{\Delta}(b))]
\Big\}\\
=&\min_{\overline{\mu}}~\Big\{ 
\sup_{\overline{\Delta}\ge-(\rho/\eta)^{-1}}~
\mathbb{E}_{b\sim\beta}[(\nabla f(z)\trans b + \overline{\mu})\cdot\overline{\Delta}(b)] -  (\rho/\eta)^{-2}\cdot \mathbb{E}_{b\sim\beta}[\phi(1 +\rho/\eta\cdot \overline{\Delta}(b))]
\Big\}\\
=&\min_{\overline{\mu}}~\Big\{ 
\mathbb{E}_{b\sim\beta}\Big[
\sup_{\overline{\Delta}\ge-(\rho/\eta)^{-1}}
(\nabla f(z)\trans b + \overline{\mu})\cdot\overline{\Delta} -  (\rho/\eta)^{-2}\cdot \phi(1 +\rho/\eta\cdot \overline{\Delta})\Big]
\Big\}\\
\le &\mathbb{E}_{b\sim\beta}\Big[
\sup_{\overline{\Delta}\ge-(\rho/\eta)^{-1}}
\nabla f(z)\trans b\cdot\overline{\Delta} -  (\rho/\eta)^{-2}\cdot \phi(1 +\rho/\eta\cdot \overline{\Delta})
\Big]
\end{align*}
Since $\phi$ is convex with $\phi''(1)>0$, it holds that the family of continuous functions
\[
s_{\delta}(y):=\sup_{\overline{\Delta}\ge-(\rho/\eta)^{-1}}
y\cdot\overline{\Delta} -  (\rho/\eta)^{-2}\cdot \phi(1 +\rho/\eta\cdot \overline{\Delta})
\]
converges uniformly on compact sets to
\[
s_0(y):=\sup_{\overline{\Delta}}
y\cdot\overline{\Delta} -  \frac{\overline{\Delta}^2\phi''(1)}{2} = \frac{y^2}{2\phi''(1)}.
\]
Consequently, 
\[
\widetilde{\mathcal{E}}_{\hP}(f;\rho,\eta)\le \frac{\rho^2}{\eta}\mathbb{E}_{z\sim\hP}\mathbb{E}_{b\sim\beta}\Big[
\sup_{\overline{\Delta}\ge-(\rho/\eta)^{-1}}
\nabla f(z)\trans b\cdot\overline{\Delta} -  (\rho/\eta)^{-2}\cdot \phi(1 +\rho/\eta\cdot \overline{\Delta})
\Big].
\]
When we take $\rho/\eta\to0$ both sides, the RHS becomes
\[
\frac{\rho^2}{\eta}\mathbb{E}_{z\sim\hP}\mathbb{E}_{b\sim\beta}\Big[ 
\frac{(\nabla f(z)\trans b)^2}{2\phi''(1)}
\Big] + o(\rho) = \frac{\rho^2}{2\eta\cdot \phi''(1)}\cdot \bE_{z\sim \hP}\Big[ 
\Var_{b\sim\beta}[\nabla f(z)\trans b]
\Big] + o(\rho).
\]
Combining lower and upper bounds gives our desired result.

\QED
\endproof

\clearpage
\section{Proof of Technical Result in Section~\ref{Sec:generation}}

\proof{Proof of Lemma~\ref{Lemma:covering:Gadv}.}
Let $\mathcal{C}(\mathcal{G})=\Big\{
(c_i^j(\cdot)):~i\in[n]
\Big\}$
be a $(\epsilon, \|\cdot\|_{\infty})$-cover of the set $ \mathcal{G}_{\mid S}$.
Define the operator
\[
\widetilde{c}_i^j = \inf_{\mu\in\mathbb{R}}~\bigg\{ 
\mu + \bE_{b\sim\beta}\left[ 
(\eta\phi)^*\Big(c_i^j(b) - \mu\Big)
\right]
\bigg\}.
\]
We now claim that $\mathcal{C}(\mathcal{G}_{\text{adv}})=\Big\{
\widetilde{c}_i^j:~i\in[n]
\Big\}$ is a $(\epsilon, |\cdot|)$-cover of the set $\mathcal{G}_{\text{adv}\mid S}$.
Indeed, for any $\theta\in\Theta$, there exists an index (by definition) $j(\theta)$ such that 
\[
\max_{i\in[n]}\max_{b\in\mathbb{B}_1(0)}~\Big|\ell(g_{\theta}(x_i+b),y_i) - c_i^{j(\theta)}(x_i,y_i)\Big| =  \max_{i\in[n]}
\|\ell(g_{\theta}(x_i + \cdot),y_i) - c_i^{j(\theta)}(\cdot)\|_{\infty}\le \epsilon.
\]
Define the functional $T:~\mathbb{R}^{\mathbb{B}_1(0)}\to\mathbb{R}$ as 
\begin{align*}
T(g) &= \inf_{\mu}\Big\{ 
\mu + \bE_{b\sim\beta}\left[ 
(\eta\phi)^*\Big(g(b) - \mu\Big)
\right]
\Big\}\\ 
&=\sup_{\bP\in\cP(\cZ)}\left\{ 
\bE_{\bP}[g] - \eta\bE_{b\sim\bP}\left[\phi\left(\frac{\diff\bP(b)}{\diff\beta(b)}\right)\right]
\right\}.
\end{align*}
Since $\phi$ is strictly convex, its directional derivative is well-defined, which is denoted as 
\[
\nabla T(g)V = \bE_{b\sim\bP_g^*}\Big[V(b)\Big],
\]
where 
\[
\bP_g^*=\argmax_{\bP\in\cP(\cZ)}\left\{ 
\bE_{\bP}[g] - \eta\bE_{b\sim\bP}\left[\phi\left(\frac{\diff\bP(b)}{\diff\beta(b)}\right)\right]
\right\}.
\]
For fixed $i$, with slight abuse of notation, let $\tau_i(\cdot;t)=t\ell(g_{\theta}(x_i + \cdot),y_i) + (1-t)c_i^{j(\theta)}(\cdot)$. 
Therefore, we have that for $g\in\mathcal{G}_{\text{adv}}$,
\begin{align*}
\max_{i\in[n]}\Big|
g(x_i,y_i) - \widetilde{c}_i^{j(\theta)}
\Big|&=\max_{i\in[n]}\Big| 
T(\ell(g_{\theta}(x_i + \cdot),y_i)) - T(c_i^{j(\theta)}(\cdot))
\Big|\\
&\le 
\max_{i\in[n]}\sup_{t\in[0,1]}~\Big| \nabla T(\tau_i(\cdot;t))(\ell(g_{\theta}(x_i + \cdot),y_i) - c_i^{j(\theta)}(\cdot))\Big|\\
&=
\max_{i\in[n]}
\sup_{t\in[0,1]}~\left|
\bE_{\varepsilon\sim\bP^*_{\tau_i(\cdot;t)}}\Big[\ell(g_{\theta}(x_i + \varepsilon),y_i) - c_i^{j(\theta)}(\varepsilon)\Big]
\right|\\
&\le \max_{i\in[n]}
\|\ell(g_{\theta}(x_i + \cdot),y_i) - c_i^{j(\theta)}(\cdot)\|_{\infty}\le \epsilon.
\end{align*}
The proof is completed.
\QED
\endproof

\clearpage
\section{Proof of Strong Duality for \texorpdfstring{$\infty$}{}-Type Casual Optimal Transport DRO}\label{Proof:casual:OT}
Let us consider the $\infty$-type Casual Optimal transport DRO problem
\begin{align}\label{Eq:casual}
\min_{\theta}\Bigg\{\sup_{\mathbb{P}, \gamma}&~\mathbb{E}_{(x,z)\sim\mathbb{P}}[\Psi(f_{\theta}(x), z)]:~
\begin{array}{l}
 ((\widehat{X}, \widehat{Z}), (X,Z))\sim\gamma\implies X\perp\widehat{Z}\mid\widehat{X},\\
 \esssup_{\gamma}~\|(\widehat{X}, \widehat{Z})- (X,Z)\|\le \rho, \Proj_{1\#}\gamma=\hP, \Proj_{2\#}\gamma=\bP
\end{array}
\Bigg\},
\end{align}
where the norm $\|(\widehat{X}, \widehat{Z})- (X,Z)\|\triangleq \|\widehat{X} - X\| + \infty\cdot \textbf{1}\{\widehat{Z}\ne Z\}$, meaning we take into account only the distribution shift of the covariate and omit the random vector distribution shift.
Here the transport mapping $\gamma$ is said to be \emph{casual} since $((\widehat{X}, \widehat{Z}), (X,Z))\sim\gamma$ implies $X\perp\widehat{Z}\mid\widehat{X}$, and it additionally satisfies the $\infty$-type optimal transport constraint with transportation budget $\rho$.
In the following, we derive the strong dual reformulation of \eqref{Eq:casual}.
We expand its objective as
\begin{align*}
&\bE_{(x,z)\sim\bP}[\Psi(f_{\theta}(x), z)]
=\bE_{((\widehat{x}, \widehat{z}), (x,z))\sim\gamma}[\Psi(f_{\theta}(x), z)]\\
=&\bE_{\widehat{x}\sim \gamma(\widehat{x})}
\bE_{x\sim \gamma(x\mid\widehat{x})}
\bE_{\widehat{z}\sim \gamma(\widehat{z}\mid x, \widehat{x})}
\bE_{z\sim \gamma(z\mid \widehat{z}, x, \widehat{x})}[\Psi(f_{\theta}(x), z)]\\
=&\bE_{\widehat{x}\sim \gamma(\widehat{x})}\bE_{x\sim \gamma(x\mid\widehat{x})}
\bE_{\widehat{z}\sim \gamma(\widehat{z}\mid \widehat{x})}
\bE_{z\sim \delta_{\widehat{z}}}[\Psi(f_{\theta}(x), z)]\\
=&\bE_{\widehat{x}\sim \gamma(\widehat{x})}\bE_{x\sim \gamma(x\mid\widehat{x})}
\bE_{\widehat{z}\sim \gamma(\widehat{z}\mid \widehat{x})}
[\Psi(f_{\theta}(x), \widehat{z})]
\end{align*}
where the first and second equality is by the law of total probability, the third equality is because $\gamma$ satisfies \emph{casual} property and we impose infinity transportation cost for moving $\widehat{z}$ to other locations.
Since $\gamma(\widehat{x}) = \widehat{\mathbb{P}}_{\widehat{x}}$ and $\gamma(\widehat{z}\mid \widehat{x})=\hP_{\widehat{z}\mid \widehat{x}}$, we are able to reformulate \eqref{Eq:casual} as
\begin{align}
&\min_{\theta}\Bigg\{\sup_{\mathbb{P}, \gamma}\bE_{\widehat{x}\sim \widehat{\mathbb{P}}_{\widehat{x}}}\bE_{x\sim \gamma(x\mid\widehat{x})}
\bE_{\widehat{z}\sim \hP_{\widehat{z}\mid \widehat{x}}}
[\Psi(f_{\theta}(x), \widehat{z})]:~
\begin{array}{l}
 \esssup_{\gamma}~\|\widehat{X}- X\|\le \rho,
\\
 \Proj_{1\#}\gamma=\hP, \Proj_{2\#}\gamma=\bP
\end{array}
\Bigg\}\\
=&\min_{\theta}~\left\{\bE_{\widehat{x}\sim \hP_{\widehat{X}}}\left[ 
\sup_{x'\in\mathbb{B}_{\rho}(\widehat{x})}~\bE_{\widehat{z}\sim\hP_{\widehat{Z}\mid\widehat{X}=\widehat{x}}}[\Psi(f_{\theta}(x'), \widehat{z})]
\right]\right\},
\end{align}
where the last equality is by the $\infty$-Wasserstein DRO strong duality result, adopted from \citep[Lemma~EC.2]{gao2022wasserstein}, with loss $\bE_{\widehat{z}\sim\hP_{\widehat{Z}\mid\widehat{X}=\widehat{x}}}[\Psi(f_{\theta}(x'), \widehat{z})]$.

Following similar procedure, one can express \eqref{reg:casual:dual} using its primal reformulation that involves $\phi$-divergence regularization:
\begin{equation}
\tag{\ref{reg:casual:dual}-Primal}
\min_{\theta}\Bigg\{\sup_{\mathbb{P}, \gamma}\mathbb{E}_{(x,z)\sim\mathbb{P}}[\Psi(f(x), z)] - \eta \mathbb{D}_{\phi}(\gamma, \gamma_0):~
\begin{array}{l}
 ((\widehat{X}, \widehat{Z}), (X,Z))\sim\gamma\implies X\perp\widehat{Z}\mid\widehat{X},\\
 \esssup_{\gamma}~\|(\widehat{X}, \widehat{Z})- (X,Z)\|\le \rho, \Proj_{1\#}\gamma=\hP, \Proj_{2\#}\gamma=\bP
\end{array}
\Bigg\},
\end{equation}
where the reference measure $\gamma_0$ satisfies the \emph{bicasual} property, $\gamma_0(x\mid\widehat{x})\equiv \nu_{\widehat{x}}(x), \forall x$ and $\gamma_0(z\mid \widetilde{z}, \widehat{x}, x)\equiv \gamma_0(z\mid\widetilde{z})\equiv\delta_{\widetilde{z}}(z), \forall z$. 
Namely, its joint distribution is decomposed as 
\begin{align*}
\gamma_0((\widehat{x}, \widehat{z}), (x,z))&=\gamma_0(\widehat{x})\gamma_0(x\mid\widehat{x})\gamma_0(\widehat{z}\mid \widehat{x},x)\gamma_0(z\mid \widehat{z}, \widehat{x}, x)\\
&=\gamma_0(\widehat{x})\gamma_0(x\mid\widehat{x})\gamma_0(\widehat{z}\mid \widehat{x})\gamma_0(z\mid \widehat{z})=\gamma_0(\widehat{x})\nu_{\widehat{x}}(x)\gamma_0(\widehat{z}\mid \widehat{x})\delta_{\widehat{z}}(z).
\end{align*}

\section{Implementation Details for Loss in Section~\ref{Sec:visualization:worst:case}}\label{Appendix:imp:detail}
For the loss $f(\cdot)$ displayed in Section~\ref{Sec:visualization:worst:case},
we take the loss $f(z) = (g(z) - 0)^2$, where $g:~\mathbb{R}\to\mathbb{R}$ is a feed-forward neural network function.
The structure of $g$ is as follows.
We first take a basis expansion to form $z'=(z, \sqrt{|z|}, z^2, \sin(z), \cos(z))\in\mathbb{R}^5$.
Then take \[
g(z)=W_4\cdot\mathrm{Sigmoid}(W_3\cdot \mathrm{Sp}(W_2\cdot\mathrm{Sp}(W_1z'))),
\]
where $\mathrm{Sig}(\cdot)$ and $\mathrm{Sp}(\cdot)$ are the sigmoid and softplus activation functions, respectively.
Weight matrics $W_1\in\mathbb{R}^{512\times 5}, W_2\in\mathbb{R}^{512\times 512}, W_2\in\mathbb{R}^{10\times512}, W_4\in\mathbb{R}^{10\times 1}$, and the entries of $W_2,W_3,W_4$ follow i.i.d. from $\mathcal{N}(0,1)$ whereas that of $W_1$ follow $\mathcal{N}(0, 0.25)$. 
This example can be viewed as adversarial robust supervised learning for using a neural network to fit a constant function at the origin.

\end{document}